\documentclass{article}

\usepackage{microtype}
\usepackage{graphicx}
\usepackage{booktabs} 

\usepackage{hyperref}

\usepackage[accepted]{icml2024}

\usepackage{amsmath}
\usepackage{amssymb}
\usepackage{mathtools}
\usepackage{amsthm}

\usepackage[capitalize,noabbrev]{cleveref}

\theoremstyle{plain}

\theoremstyle{definition}

\theoremstyle{remark}

\usepackage[textsize=tiny]{todonotes}

\usepackage{epsfig}
\usepackage{multicol, multirow, makecell}
\usepackage{subcaption}
\usepackage[linesnumbered, ruled, algo2e]{algorithm2e}
\let\oldnl\nl
\newcommand{\nonl}{\renewcommand{\nl}{\let\nl\oldnl}}
\usepackage{float}
\usepackage{bbding}

\icmltitlerunning{KernelWarehouse: Rethinking the Design of Dynamic Convolution}

\begin{document}

\twocolumn[
\icmltitle{KernelWarehouse: Rethinking the Design of Dynamic Convolution}



\icmlsetsymbol{equal}{*}

\begin{icmlauthorlist}
\icmlauthor{Chao Li}{ilc}
\icmlauthor{Anbang Yao}{ilc}
\end{icmlauthorlist}

\icmlaffiliation{ilc}{Intel Labs China}

\icmlcorrespondingauthor{Anbang Yao}{anbang.yao@intel.com}

\icmlkeywords{Machine Learning, ICML}

\vskip 0.3in
]



\printAffiliationsAndNotice{}  

\begin{abstract}
Dynamic convolution learns a linear mixture of $n$ static kernels weighted with their input-dependent attentions, demonstrating superior performance than normal convolution. However, it increases the number of convolutional parameters by $n$ times, and thus is not parameter efficient. This leads to no research progress that can allow researchers to explore the setting $n>100$ (an order of magnitude larger than the typical setting $n<10$) for pushing forward the performance boundary of dynamic convolution while enjoying parameter efficiency. To fill this gap, in this paper, we propose ~\textit{KernelWarehouse}, a more general form of dynamic convolution, which redefines the basic concepts of ``kernels", ``assembling kernels" and ``attention function" through the lens of exploiting convolutional parameter dependencies within the same layer and across neighboring layers of a ConvNet. We testify the effectiveness of KernelWarehouse on ImageNet and MS-COCO datasets using various ConvNet architectures. Intriguingly, KernelWarehouse is also applicable to Vision Transformers, and it can even reduce the model size of a backbone while improving the model accuracy. For instance, KernelWarehouse ($n=4$) achieves $5.61\%|3.90\%|4.38\%$ absolute top-1 accuracy gain on the ResNet18$|$MobileNetV2$|$DeiT-Tiny backbone, and KernelWarehouse ($n=1/4$) with 65.10\% model size reduction still achieves 2.29\% gain on the ResNet18 backbone. The code and models are available at \url{https://github.com/OSVAI/KernelWarehouse}.
\end{abstract}

\section{Introduction}

Convolution is the key operation in 
convolutional neural networks (ConvNets). In a convolutional layer, normal convolution $\mathbf{y} = \mathbf{W}*\mathbf{x}$ computes the output $\mathbf{y}$ by applying the same convolutional kernel $\mathbf{W}$ defined as a set of convolutional filters to every input sample $\mathbf{x}$.~\textit{For brevity, we refer to ``convolutional kernel" as ``kernel" and omit the bias term throughout this paper}. Although the efficacy of normal convolution is extensively validated with various types of ConvNet architectures~\citep{CNN_AlexNet,CNN_ResNet,CNN_MobileNets,CNN_ConvNeXt} on many 
computer vision tasks, recent progress in the efficient ConvNet architecture design shows that dynamic convolution, known as CondConv~\citep{DynamicConv_CondConv} and DY-Conv~\citep{DynamicConv_DyConv}, achieves large performance gains.

The basic idea of dynamic convolution is to replace the single kernel in normal convolution by a linear mixture of $n$ same dimensioned kernels, $\mathbf{W}=\alpha_{1}\mathbf{W}_1+...+\alpha_{n}\mathbf{W}_n$, where $\alpha_{1},...,\alpha_{n}$ are scalar attentions generated by an input-dependent attention module. Benefiting from the additive property of $\mathbf{W}_1,...,\mathbf{W}_n$ and compact attention module designs, dynamic convolution improves the feature learning ability with little extra multiply-add cost against normal convolution. However, it increases the number of convolutional parameters by $n$ times, which leads to a huge rise in model size because the convolutional layers of a modern ConvNet occupy the vast majority of parameters. 
There exist few research works to alleviate this problem.
DCD
~\citep{DynamicConv_ResDyConv} learns a base kernel and a sparse residual to approximate dynamic convolution via matrix decomposition. This approximation abandons the basic mixture learning paradigm
, and thus cannot retain the representation power of dynamic convolution when $n$ becomes large. 
ODConv~\citep{DynamicConv_ODConv} presents an improved attention module to dynamically weight static kernels along different dimensions instead of one single dimension, which can get competitive performance with a reduced number of kernels. However, under the same setting of $n$, ODConv has more parameters than vanilla dynamic convolution. Recently, ~\citet{DynamicConv_SDConv} directly used popular weight pruning strategy to compress DY-Conv via multiple pruning-and-retraining phases.

In a nutshell, existing dynamic convolution methods based on the linear mixture learning paradigm 
are not parameter-efficient. Restricted by this, the kernel number is typically set to $n=8$~\citep{DynamicConv_CondConv} or $n=4$~\citep{DynamicConv_DyConv,DynamicConv_ODConv}. However, 
a plain fact is that the improved capacity of a ConvNet constructed with dynamic convolution comes from increasing the kernel number $n$ per convolutional layer facilitated by the attention mechanism. This causes a fundamental conflict between the desired model size and capacity. In this work, we rethink the design of dynamic convolution, with the goal of reconciling such a conflict, enabling us to explore the performance boundary of dynamic convolution with the significantly larger kernel number setting $n>100$ (an order of magnitude larger than the typical setting $n<10$) while enjoying parameter efficiency. Note that, for existing dynamic convolution methods, $n>100$ means that the model size will be about $>100$ times larger than the backbone built with normal convolution. 

\begin{figure*}[ht]
\vskip 0.05 in
\begin{center}
\centerline{\includegraphics[width=0.8\linewidth]{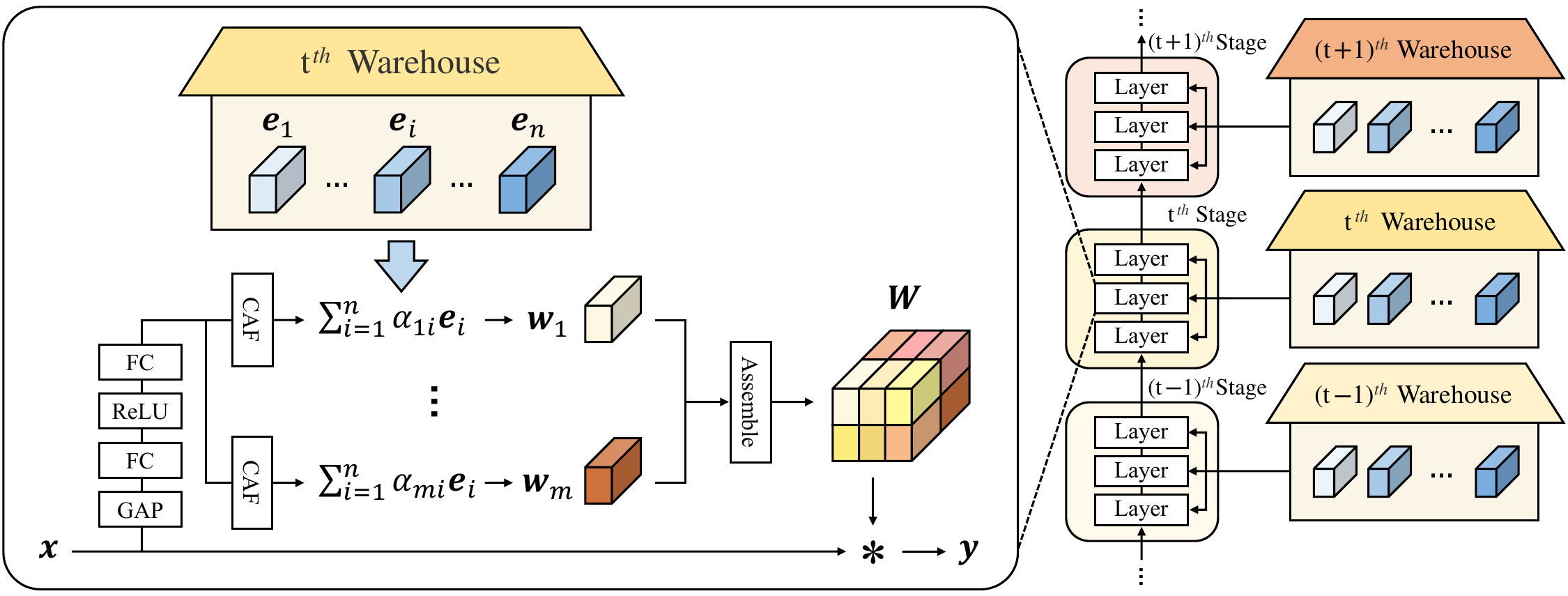}}
\vskip -0.05 in
\caption{A schematic overview of~\textit{KernelWarehouse} to a ConvNet. As a more general form of dynamic convolution, KernelWarehouse consists of three interdependent components, namely kernel partition, warehouse construction-with-sharing and contrasting-driven attention function (CAF), which redefine the basic concepts of ``kernels", ``assembling kernels" and ``attention function" in the perspective of exploiting convolutional parameter dependencies within the same layer and across neighboring layers of a ConvNet,  enabling to use significantly large kernel number settings (e.g., $n>100$) while enjoying improved model accuracy and parameter efficiency. Please see the Method section for the detailed formulation.}
\label{fig:architecture}
\end{center}
\vskip -0.2 in
\end{figure*}

To this goal, we present a more general form of dynamic convolution called~\textit{KernelWarehouse} (Figure~\ref{fig:architecture} shows a schematic overview).
Our work is inspired by two observations about existing dynamic convolution methods: (1) They treat all parameters in a regular convolutional layer as a static kernel, increase the kernel number from 1 to $n$, and use their attention modules to assemble $n$ static kernels into a linearly mixed kernel. Though straightforward and effective, they pay no attention to parameter dependencies within the static kernel at a convolutional layer; (2) They allocate different sets of $n$ static kernels for individual convolutional layers of a ConvNet, 
ignoring parameter dependencies across neighboring convolutional layers. In a sharp contrast with existing methods, the core philosophy of KernelWarehouse is to~\textit{exploit convolutional parameter dependencies within the same layer and across neighboring layers of a ConvNet},
reformulating dynamic convolution towards achieving a substantially better trade-off between parameter efficiency and representation power. 

KernelWarehouse consists of three components, namely kernel partition, warehouse construction-with-sharing and contrasting-driven attention function, which are closely interdependent. Kernel partition exploits parameter dependencies within the same convolutional layer, by which we redefine ``\textit{kernels}" for the linear mixture in terms of a much smaller local scale instead of a holistic scale. Warehouse construction-with-sharing exploits parameter dependencies across neighboring convolutional layers, by which we redefine ``\textit{assembling kernels}" across all same-stage convolutional layers instead of within a single layer, and generate a large warehouse consisting of $n$ local kernels (e.g., $n=108$) shared for cross-layer linear mixtures. Contrasting-driven attention function is customized to address the attention optimization problem under the cross-layer linear mixture learning paradigm with the challenging settings of $n>100$, 
by which we redefine ``\textit{attention function}". Given different convolutional parameter budgets (see Section.~\ref{sec:3.3} for the definition of convolutional parameter budget), these concept shifts provide a high degree of flexibility for KernelWarehouse, allowing to well balance parameter efficiency and representation power with sufficiently large values of $n$.

As a drop-in replacement of normal convolution, KernelWarehouse can be easily used to various types of ConvNet architectures. We validate the effectiveness of KernelWarehouse through extensive experiments on ImageNet and MS-COCO datasets. On one hand, we show that KernelWarehouse achieves superior performance compared to existing dynamic convolution methods (e.g., the ResNet18$|$ResNet50$|$MobileNetV2$|$ConvNeXt-Tiny model with KernelWarehouse trained on ImageNet dataset reaches 76.05\%$|$81.05\%$|$75.92\%$|$82.55\% top-1 accuracy, setting new performance records for dynamic convolution research). On the other hand, we show that all three components of KernelWarehouse are essential to the performance boost in terms of model accuracy and parameter efficiency, and KernelWarehouse can even reduce the model size of a ConvNet while improving the model accuracy (e.g., our ResNet18 model with 65.10\% parameter reduction to the baseline still achieves 2.29\% absolute top-1 accuracy gain), and it is also applicable to Vision Transformers (e.g., our DeiT-Tiny model reaches 76.51\% top-1 accuracy, bringing 4.38\% absolute top-1 accuracy gain to the baseline).

\section{Related Work}

\textbf{ConvNet Architectures.} In the past decade, many notable ConvNet architectures such as AlexNet~\citep{CNN_AlexNet}, VGGNet~\citep{CNN_VGGNet}, GoogLeNet~\citep{CNN_GoogLeNet}, ResNet~\citep{CNN_ResNet}, DenseNet~\citep{CNN_DenseNet}, ResNeXt~\citep{CNN_ResNeXt} and RegNet~\citep{CNN_RegNet} have been presented. 
Around the same time, some lightweight architectures like MobileNet~\citep{CNN_MobileNets,CNN_MobileNetV2,CNN_MobileNetV3}, ShuffleNet~\citep{CNN_ShuffleNet} and EfficientNet~\citep{CNN_EfficientNet} have been designed for resource-constrained applications. Recently,~\citet{CNN_ConvNeXt} presented ConvNeXt whose performance can match newly emerging vision transformers~\citep{SelfAttention_ViT,SelfAttention_Swin}. 
Our method could be used to improve their performance, as we show in the experiments.

\textbf{Feature Recalibration.} An effective way to enhance the capacity of a ConvNet is feature recalibration. It relies on attention mechanisms to adaptively refine the feature maps learnt by a convolutional block. Popular feature recalibration modules such as RAN~\citep{Attention_ResidualAttention}, SE~\citep{Attention_SE}, BAM~\citep{Attention_BAM}, CBAM~\citep{Attention_CBAM}, GE~\citep{Attention_GE}, SRM~\citep{Attention_SRM} and ECA~\citep{Attention_ECA} focus on different design aspects: using channel attention, or spatial attention, or hybrid attention to emphasize important features and suppress unnecessary ones. Unlike these methods that retain the static kernel, dynamic convolution replaces the single static kernel of a convolutional layer by a linear mixture of $n$ static kernels weighted with the attention mechanism.

\textbf{Dynamic Weight Networks.} Many research efforts have been made on developing effective methods to generate the weights for a neural network.~\citet{WeightGeneration_STNetworks} proposed a Spatial Transformer module which uses a localisation network that predicts the feature transformation parameters conditioned on the learnt feature itself. Dynamic Filter Network~\citep{WeightGeneration_DynamicFilterNetworks} and Kernel Prediction Networks~\citep{WeightGeneration_KPN, WeightGeneration_KPN_BurstDenoising} introduce two filter generation frameworks which share the same idea: using a deep neural network to generate sample-adaptive filters conditioned on the input. Based on this idea, DynamoNet~\citep{WeightGeneration_DynamoNet} uses dynamically generated motion filters to boost video-based human action recognition. CARAFE~\citep{WeightGeneration_CARAFE} and Involution~\citep{WeightGeneration_Involution} further extend this idea by designing efficient generation modules to predict the weights for extracting informative features. By connecting this idea with SE, WeightNet~\citep{WeightGeneration_WeightNet}, CGC~\citep{DynamicConv_CGC} and WE~\citep{DynamicConv_WeightExcitation} design different attention modules to adjust the weights in convolutional layers of a ConvNet. Hypernetwork~\citep{WeightGeneration_Hypernetworks} uses a small network to generate the weights for a larger recurrent network instead of a ConvNet. MetaNet~\citep{WeightGeneration_MetaNetworks} introduces a meta learning model consisting of a base learner and a meta learner, allowing the learnt network for rapid generalization across different tasks. Unlike them, this work focuses on advancing dynamic convolution research. 

\section{Method}



\subsection{Motivation and Components of KernelWarehouse}

For a convolutional layer, let $\mathbf{x} \in \mathbb{R}^{h \times w \times c}$ and $\mathbf{y} \in \mathbb{R}^{h \times w \times f}$ be the input having $c$ feature channels and the output having $f$ feature channels respectively, where $h \times w$ denotes the channel size. Normal convolution $\mathbf{y} = \mathbf{W}*\mathbf{x}$ uses a static kernel $\mathbf{W} \in \mathbb{R}^{k \times k \times c \times f}$ consisting of $f$ convolutional filters with the spatial size $k\times k$. Dynamic convolution~\citep{DynamicConv_CondConv,DynamicConv_DyConv} replaces $\mathbf{W}$ in normal convolution by a linear mixture of $n$ same dimensioned static kernels $\mathbf{W}_1,...,\mathbf{W}_n$ weighted with $\alpha_{1},...,\alpha_{n}$ generated by an attention module $\phi(x)$
, which is defined as

\begin{equation}
\label{eq:00}
\begin{aligned}
 & \mathbf{W}=\alpha_{1} \mathbf{W}_1+...+\alpha_{n} \mathbf{W}_n.
\end{aligned}
\end{equation}

As we discussed earlier, the kernel number $n$ is typically set to $n<10$, restricted by the parameter-inefficient shortcoming. The main motivation of this work is to reformulate this linear mixture learning paradigm, 
enabling us to explore significantly larger settings, e.g., $n>100$ (an order of magnitude larger than the typical setting $n<10$), for pushing forward the performance boundary of dynamic convolution while enjoying parameter efficiency. To that end, our proposed KernelWarehouse has three key components: kernel partition, warehouse construction-with-sharing, and contrasting-driven attention function.




\subsection{Kernel Partition}

\begin{figure*}[ht]
\vskip 0.05 in
\begin{center}
\centerline{\includegraphics[width=0.725\linewidth]{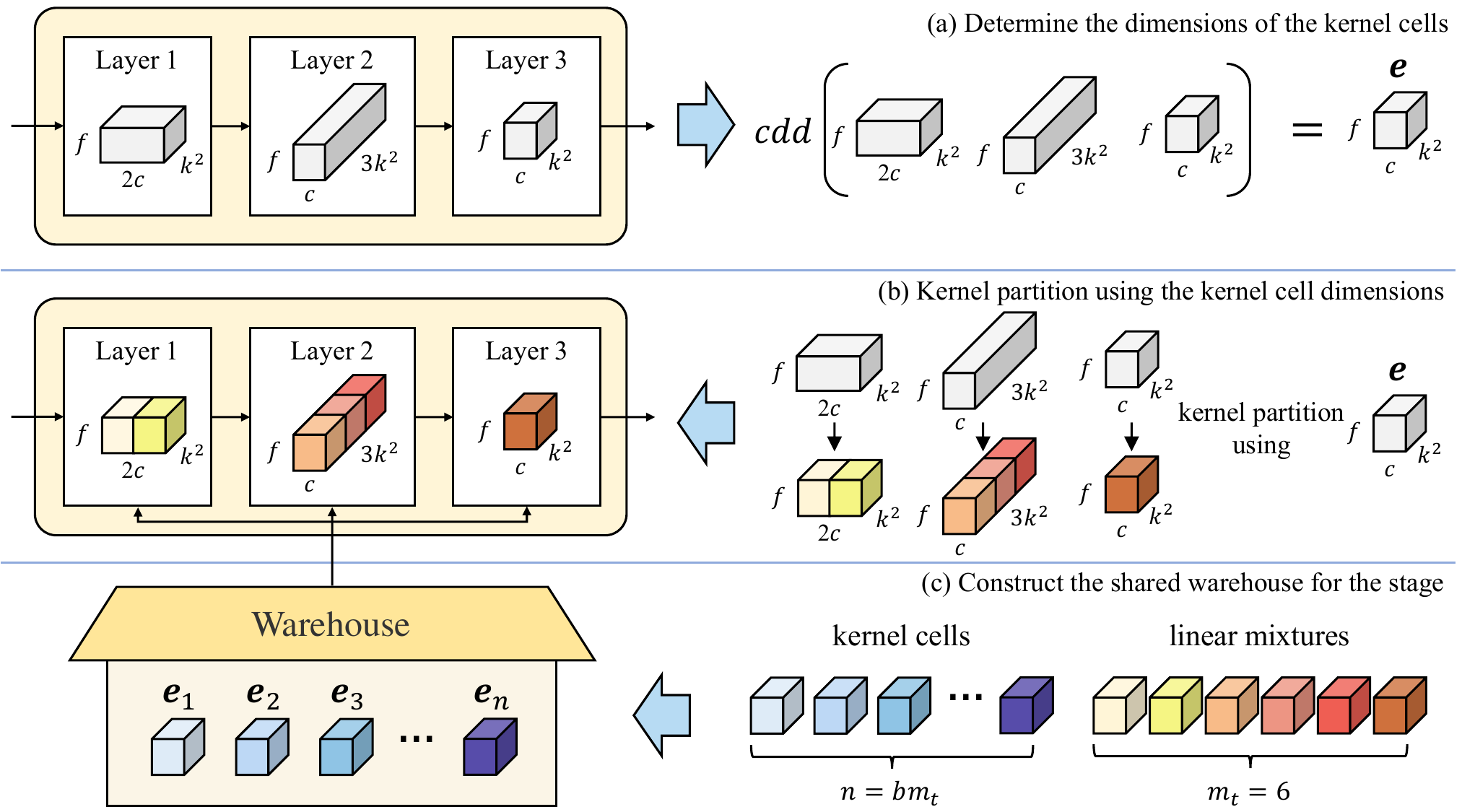}}
\vskip -0.05 in
\caption{An illustration of~\textit{kernel partition} and~\textit{warehouse construction-with-sharing} across three same-stage convolutional layers of a ConvNet. $cdd$ denotes common kernel dimension divisors, and~\textbf{$b$} is the desired convolutional parameter budget.}
\label{fig:kernel_partition}
\end{center}
\vskip -0.2 in
\end{figure*}

The main idea of kernel partition is to reduce kernel dimension via simply exploiting the parameter dependencies within the same convolutional layer. Specifically, for a regular convolutional layer, we sequentially divide the static kernel $\mathbf{W}$ along spatial and channel dimensions into $m$ disjoint parts $\mathbf{w}_1$,...,$\mathbf{w}_m$ called ``\textit{kernel cells}" that have the same dimensions.~\textit{For brevity, here we omit to define kernel cell dimensions}. Kernel partition can be defined as

\begin{equation}
\label{eq:01}
\begin{aligned}
 & \mathbf{W} = \mathbf{w}_1\cup...\cup \mathbf{w}_m, \ \\ & \mathrm{and}\ \forall\ i,j\in\{1,...,m\}, i \ne j, \ \mathbf{w}_i \cap \mathbf{w}_j = \mathbf{\emptyset}.
\end{aligned}
\end{equation}

After kernel partition, we treat kernel cells $\mathbf{w}_1$,...,$\mathbf{w}_m$ as ``\textit{local kernels}", and define a ``\textit{warehouse}" containing $n$ kernel cells $\mathbf{E}=\{\mathbf{e}_1,...,\mathbf{e}_n\}$, where $\mathbf{e}_1,...,\mathbf{e}_n$ have the same dimensions as $\mathbf{w}_1$,...,$\mathbf{w}_m$. Then, 
we use the warehouse $\mathbf{E}=\{\mathbf{e}_1,...,\mathbf{e}_n\}$ to represent each of $m$ kernel cells $\mathbf{w}_1$,...,$\mathbf{w}_m$ as a linear mixture
\begin{equation}
\label{eq:02}
\mathbf{w}_i =\alpha_{i1} \mathbf{e}_1+...+\alpha_{in} \mathbf{e}_n, \ \mathrm{and}\ i\in\{1,...,m\},
\end{equation}
where $\alpha_{i1},...,\alpha_{in}$ are the input-dependent scalar attentions generated by an attention module $\phi(x)$. Finally, the static kernel $\mathbf{W}$ in a regular convolutional layer is replaced by 
its corresponding $m$ linear mixtures.

Thanks to kernel partition, the dimensions of the kernel cell $\mathbf{w_i}$ could be much smaller than the dimensions of the static kernel $\mathbf{W}$. For example, when $m=16$, the number of convolutional parameters in the kernel cell $\mathbf{w_i}$ is 1/16 to that of the static kernel $\mathbf{W}$. Under a desired convolutional parameter budget $b$ (we will define it in Section.~\ref{sec:3.3}), this allows a warehouse can easily have a significantly larger value of $n$ (e.g., $n=64$),
in comparison to existing dynamic convolution methods that define the linear mixture in terms of $n$ (e.g., $n=4$) ``\textit{holistic kernels}". 

\subsection{Warehouse Construction-with-Sharing} 
\label{sec:3.3}

The main idea of warehouse construction-with-sharing is to 
further improve the warehouse-based linear mixture learning formulation through simply exploiting parameter dependencies across neighboring convolutional layers. Specifically, for $l$ same-stage convolutional layers of a ConvNet, we construct a shared warehouse $\mathbf{E}=\{\mathbf{e}_1,...,\mathbf{e}_n\}$ by using the same kernel cell dimensions for kernel partition. This allows the shared warehouse not only can have a larger
value of n (e.g., $n=188$) compared to the layer-specific warehouse (e.g., $n=36$), but also can have improved representation power (see Table~\ref{table:ablation_sharingscale}). Thanks to the modular design mechanism of modern ConvNets, 
we simply use common dimension divisors for all $l$ same-stage static kernels as the uniform kernel cell dimensions to perform kernel partition, which naturally determines the kernel cell number $m$ for each same-stage convolutional layer, as well as $n$ for the shared warehouse when a desired convolutional parameter budget $b$ is given. Figure~\ref{fig:kernel_partition} illustrates the processes of kernel partition and warehouse construction-with-sharing.

\textbf{Convolutional Parameter Budget.} For vanilla dynamic convolution~\citep{DynamicConv_CondConv,DynamicConv_DyConv}, the convolutional parameter budget $b$ relative to normal convolution is always equal to the kernel number. That is, $b==n$, and $n>=1$. When setting a large value of $n$, e.g., $n=188$, existing dynamic convolution methods get $b=188$, leading to about $188$ times larger model size for a ConvNet backbone. For KernelWarehouse, such drawbacks are resolved. Let $m_{t}$ be the total number of kernel cells in $l$ same-stage convolutional layers ($m_{t} = m$, when $l = 1$) of a ConvNet. 
Then, the convolutional parameter budget of KernelWarehouse relative to normal convolution can be defined as~\textbf{$b=n/m_{t}$}. 
In implementation, we use the same value of $b$ to all convolutional layers of a ConvNet, so that 
KernelWarehouse can easily scale up or scale down the model size of a ConvNet by changing $b$. Compared to normal convolution: (1) When $b<1$, KernelWarehouse tends to get the reduced model size; (2) When $b=1$, KernelWarehouse tends to get the similar model size; (3) When $b>1$, KernelWarehouse tends to get the increased model size. 
 
\textbf{Parameter Efficiency and Representation Power.} Intriguingly, given a desired parameter budget $b$, a proper and large value of $n$ can be obtained by simply changing $m_t$ (controlled by kernel partition and warehouse construction-with-sharing), providing a representation power guarantee for KernelWarehouse. Thanks to this flexibility, KernelWarehouse can strike a favorable trade-off between parameter efficiency and representation power, under different convolutional parameter budgets, as well validated by trained model exemplifications KW ($1/2\times, 3/4\times, 1\times, 4\times$) in Table~\ref{table:classification_resnet_100epoch}, Table~\ref{table:classification_cnn_300epoch} and Table~\ref{table:classification_mobilenetv2}.
\subsection{Contrasting-driven Attention Function} 

With the above formulation, the optimization of KernelWarehouse differs with existing dynamic convolution methods in three aspects: (1) The linear mixture is used to a dense local kernel cell scale instead of a holistic kernel scale; (2) The number of kernel cells in a warehouse is significantly larger ($n>100$ vs. $n<10$); (3) A warehouse is not only shared to represent each of $m$ kernel cells for a specific convolutional layer of a ConvNet, but also is shared to represent every kernel cell for the other $l-1$ same-stage convolutional layers. However, for KernelWarehouse with these optimization properties, we empirically find that popular attention functions 
lose their effectiveness (see Table~\ref{table:ablation_function}). 
We present contrasting-driven attention function (CAF), a customized design, to solve the optimization of KernelWarehouse. For $i^{th}$ kernel cell in the static kernel $\mathbf{W}$, let $z_{i1},...,z_{in}$ be the feature logits generated by the second fully-connected layer of a compact SE-typed attention module $\phi(x)$ (\textit{its structure is clarified in Appendix}), then CAF is defined as
%
\begin{equation}
\label{eq:03}
\alpha_{ij} = \tau\beta_{ij} + (1-\tau) \frac{z_{ij}}{\sum^{n}_{p=1}{|z_{ip}|}}, \ \mathrm{and}\ j\in\{1,...,n\},
\end{equation}
%
where $\tau$ is a temperature linearly reducing from $1$ to $0$ in the early training stage; $\beta_{ij}$ is a binary value (0 or 1) for initializing attentions; $\frac{z_{ij}}{\sum^{n}_{p=1}{|z_{ip}|}}$ is a normalization function.

Our CAF relies on two smart design principles: (1) The first term ensures that the initial valid kernel cells ($\beta_{ij}=1$) in a shared warehouse are uniformly allocated to represent different linear mixtures at all $l$ same-stage convolutional layers of a ConvNet when starting the training; (2) The second term enables the existence of both negative and positive attentions, unlike popular attention functions that always generate positive attentions. This encourages the optimization process to learn contrasting and diverse attention distributions among all linear mixtures at $l$ same-stage convolutional layers sharing the same warehouse (as illustrated in Figure~\ref{figure:attention_resnet18_heatmap}), guaranteeing to improving model performance. 

At CAF initialization, the setting of $\beta_{ij}$ at $l$ same-stage convolutional layers should assure the shared warehouse can assign: (1) At least one specified kernel cell ($\beta_{ij}=1$) to every linear mixture, given 
$b\geq1$; (2) At most one specific kernel cell ($\beta_{ij}=1$) to every linear mixture, given $b<1$.
We adopt a simple strategy to assign one of the total $n$ kernel cells in a shared warehouse to each of $m_{t}$ linear mixtures at $l$ same-stage convolutional layers without repetition.
When $n<m_{t}$,
we let the remaining linear mixtures always have $\beta_{ij}=0$ once $n$ kernel cells are used up.~\textit{In the Appendix}, we provide visualization examples to illustrate this strategy. 

\subsection{Discussion}

Note that the split-and-merge strategy with multi-branch group convolution has been widely used in many ConvNet architectures~\citep{CNN_GoogLeNet,CNN_ResNeXt,CNN_MobileNetV2,Attention_SKNets,GroupConv_MixConv,GroupConv_LegoFilters,GroupConv_MixConv,GroupConv_PSConv,CNN_ConvNeXt}. Although KernelWarehouse also uses the parameter splitting idea in kernel partition, its focus and motivation we have clarified above are clearly different from them. Besides, KernelWarehouse could be also used to improve their performance as they use normal convolution. We validate this on MobileNetV2 and ConvNeXt (see Table~\ref{table:classification_cnn_300epoch} and Table~\ref{table:classification_mobilenetv2}).

According to its formulation, KernelWarehouse will degenerate into vanilla dynamic convolution~\citep{DynamicConv_CondConv,DynamicConv_DyConv} when uniformly setting $m=1$ in kernel partition (i.e., all kernel cells in each warehouse have the same dimensions as the static kernel $\mathbf{W}$ in normal convolution) and $l=1$ in warehouse sharing (i.e., each warehouse is used to its specific convolutional layer). Therefore, KernelWarehouse is a more general form of dynamic convolution. 

In formulation, the three key components of KernelWarehouse are closely interdependent, and their joint regularizing effect leads to significantly improved performance in terms of both model accuracy and parameter efficiency, as validated by multiple ablations in the Experiments section.

\section{Experiments}
In this section, we conduct comprehensive experiments on ImageNet dataset~\citep{Dataset_ImageNet} and MS-COCO dataset~\citep{Dataset_MS_COCO} to evaluate the effectiveness of our proposed KernelWarehouse (``\textbf{KW}" for short, in Tables). 
\subsection{Image Classification on ImageNet Dataset}
Our basic experiments are conducted on ImageNet dataset. 

\begin{table}[ht]
\vskip 0.05 in
\caption{Results comparison on ImageNet with the ResNet18 backbone using the traditional training strategy.~\textit{Best results are bolded.}}
\label{table:classification_resnet_100epoch}
\vskip -0.1 in
\begin{center}
\begin{small}
\resizebox{0.95\linewidth}{!}{
\begin{tabular}{l|c|c|c}
\hline
Models & Params & Top-1 Acc (\%) & Top-5 Acc (\%)\\
\hline
ResNet18 & 11.69M & 70.25 & 89.38 \\
\hline
+ SE & 11.78M & 70.98 ($\uparrow$0.73) & 90.03 ($\uparrow$0.65) \\
+ CBAM & 11.78M & 71.01 ($\uparrow$0.76) & 89.85 ($\uparrow$0.47) \\
+ ECA & 11.69M & 70.60 ($\uparrow$0.35) & 89.68 ($\uparrow$0.30) \\
+ CGC & 11.69M & 71.60 ($\uparrow$1.35) & 90.35 ($\uparrow$0.97) \\
+ WeightNet & 11.93M & 71.56 ($\uparrow$1.31) & 90.38 ($\uparrow$1.00) \\
+ DCD & 14.70M & 72.33 ($\uparrow$2.08) & 90.65 ($\uparrow$1.27) \\
+ CondConv ($8\times$) & 81.35M & 71.99 ($\uparrow$1.74) & 90.27 ($\uparrow$0.89) \\
+ DY-Conv ($4\times$) & 45.47M & 72.76 ($\uparrow$2.51) & 90.79 ($\uparrow$1.41) \\
+ ODConv ($4\times$) & 44.90M & 73.97 ($\uparrow$3.72) & 91.35 ($\uparrow$1.97) \\
\hline
+ KW ($1/2\times$) & 7.43M & 72.81 ($\uparrow$2.56) & 90.66 ($\uparrow$1.28) \\
+ KW ($1\times$) & 11.93M & 73.67 ($\uparrow$3.42) & 91.17 ($\uparrow$1.79) \\
+ KW ($2\times$) & 23.24M & 74.03 ($\uparrow$3.78) & 91.37 ($\uparrow$1.99) \\
+ KW ($4\times$) & 45.86M & \textbf{74.16} ($\uparrow$\textbf{3.91}) & \textbf{91.42} ($\uparrow$\textbf{2.04}) \\
\hline
\end{tabular}
}
\end{small}
\end{center}
\vskip -0.1 in
\end{table}

\textbf{ConvNet Backbones.} We select five ConvNet backbones from MobileNetV2~\citep{CNN_MobileNetV2}, ResNet~\citep{CNN_ResNet} and ConvNeXt~\citep{CNN_ConvNeXt} families for experiments, including both lightweight and larger architectures. 

\textbf{Experimental Setup.}
In the experiments, we make diverse comparisons of our method with related methods to demonstrate its effectiveness. Firstly, on the ResNet18 backbone, we compare our method with various state-of-the-art attention based methods, including: (1) SE~\citep{Attention_SE}, CBAM~\citep{Attention_CBAM} and ECA~\citep{Attention_ECA}, which focus on feature recalibration; (2) CGC~\citep{DynamicConv_CGC} and WeightNet~\citep{WeightGeneration_WeightNet}, which focus on adjusting convolutional weights; (3) CondConv~\citep{DynamicConv_CondConv}, DY-Conv~\citep{DynamicConv_DyConv}, DCD~\citep{DynamicConv_ResDyConv} and ODConv~\citep{DynamicConv_ODConv}, which focus on dynamic convolution.
Secondly, we select DY-Conv~\citep{DynamicConv_DyConv} and ODConv~\citep{DynamicConv_ODConv} as our key reference methods, since they are top-performing dynamic convolution methods which are most closely related to our method. We compare our KernelWarehouse with them on all the other ConvNet backbones except ConvNeXt-Tiny (since there is no publicly available implementation of them on ConvNeXt).
To make fair comparisons, all the methods are implemented using the public codes with the same settings for training and testing.
In the experiments, we use $b\times$ to denote the convolutional parameter budget of each dynamic convolution method relative to normal convolution, the values of $n$ and $m$ in KernelWarehouse and the experimental details for each ConvNet backbone are provided~\textit{in the Appendix}.

\begin{table}[ht]
\vskip 0.05 in
\caption{Results comparison on ImageNet with the ResNet18, ResNet50 and ConvNeXt-Tiny backbones using the advanced training strategy~\citep{CNN_ConvNeXt}.} 
\label{table:classification_cnn_300epoch}
\vskip -0.1 in
\begin{center}
\begin{small}
\resizebox{0.95\linewidth}{!}{
\begin{tabular}{l|c|c|c}
\hline
Models & Params & Top-1 Acc (\%) & Top-5 Acc (\%)\\
\hline
ResNet18 & 11.69M & 70.44 & 89.72 \\
+ DY-Conv ($4\times$) & 45.47M & 73.82 ($\uparrow$3.38) & 91.48 ($\uparrow$1.76) \\
+ ODConv ($4\times$) & 44.90M & 74.45 ($\uparrow$4.01) & 91.67 ($\uparrow$1.95)\\
+ KW ($1/4\times$) & 4.08M & 72.73 ($\uparrow$2.29) & 90.83 ($\uparrow$1.11) \\
+ KW ($1/2\times$) & 7.43M & 73.33 ($\uparrow$2.89) & 91.42 ($\uparrow$1.70) \\
+ KW ($1\times$) & 11.93M & 74.77 ($\uparrow$4.33) & 92.13 ($\uparrow$2.41) \\
+ KW ($2\times$) & 23.24M & 75.19 ($\uparrow$4.75) & 92.18 ($\uparrow$2.46) \\
+ KW ($4\times$) & 45.86M & \textbf{76.05} ($\uparrow$\textbf{5.61}) & \textbf{92.68} ($\uparrow$\textbf{2.96}) \\
\hline
ResNet50 & 25.56M & 78.44 & 94.24 \\
+ DY-Conv ($4\times$) & 100.88M & 79.00 ($\uparrow$0.56) & 94.27 ($\uparrow$0.03) \\
+ ODConv ($4\times$) & 90.67M & 80.62 ($\uparrow$2.18) & 95.16 ($\uparrow$0.92)\\
+ KW ($1/2\times$) & 17.64M & 79.30 ($\uparrow$0.86) & 94.71 ($\uparrow$0.47) \\
+ KW ($1\times$) & 28.05M & 80.38 ($\uparrow$1.94) & 95.19 ($\uparrow$0.95) \\
+ KW ($4\times$) & 102.02M & \textbf{81.05} ($\uparrow$\textbf{2.61}) & \textbf{95.21} ($\uparrow$\textbf{0.97}) \\
\hline
ConvNeXt-Tiny & 28.59M & 82.07 & 95.86 \\
+ KW ($3/4\times$) & 24.53M & 82.23 ($\uparrow$0.16) & 95.88 ($\uparrow$0.02) \\
+ KW ($1\times$) & 32.99M & \textbf{82.55} ($\uparrow$\textbf{0.48}) & \textbf{96.08} ($\uparrow$\textbf{0.22}) \\
\hline
\end{tabular}
}
\end{small}
\end{center}
\vskip -0.1 in
\end{table}

\textbf{Results Comparison with Traditional Training Strategy.} 
We first use the traditional training strategy adopted by lots of previous studies, training the ResNet18 backbone for 100 epochs. The results are shown in Table~\ref{table:classification_resnet_100epoch}.
%
It can be observed that dynamic convolution methods (CondConv, DY-Conv and ODConv), which introduce obviously more extra parameters, tend to have better performance compared with the other reference methods (SE, CBAM, ECA, CGC, WeightNet and DCD). Note that our KW ($1/2\times$), which has 36.45\% parameters less than the baseline, can even outperform all the above attention based methods (except ODConv ($4\times$), but including CondConv ($8\times$) and DY-Conv ($4\times$) which increase the model size to 6.96$|$3.89 times). Comparatively, our KW ($4\times$) achieves the best results. 
%
%
%
%
\begin{table*}[ht!]
\vskip 0.05 in
\caption{Results comparison on MS-COCO using the pre-trained backbone models.~\textit{Best results are bolded.}}
\vskip -0.1 in
\label{table:detection}
\begin{center}
\begin{small}
\resizebox{0.9\linewidth}{!}
{
\begin{tabular}{c|l|c|c|c|c|c|c|c|c|c|c|c|c}
\hline
\multirow{2}*{\makecell[c]{Detectors}} & \multirow{2}*{\makecell[c]{Backbone Models}} & \multicolumn{6}{c|}{Object Detection} &  \multicolumn{6}{c}{Instance Segmentation} \\
\cline{3-14}
& & $AP $ & $AP_{50} $ & $AP_{75} $ & $AP_{S} $ & $AP_{M} $ & $AP_{L} $ & $AP $ & $AP_{50} $ & $AP_{75} $ & $AP_{S} $ & $AP_{M} $ & $AP_{L} $ \\
\hline
\multirow{12}*{\makecell[c]{Mask R-CNN}} & ResNet50 & 39.6 & 61.6 & 43.3 & 24.4 & 43.7 & 50.0 & 36.4 & 58.7 & 38.6 & 20.4 & 40.4 & 48.4 \\
& + DY-Conv ($4\times$) & 39.6 & 62.1 & 43.1 & 24.7 & 43.3 & 50.5 & 36.6 & 59.1 & 38.6 & 20.9 & 40.2 & 49.1 \\
& + ODConv ($4\times$) & 42.1 & 65.1 & 46.1 & \textbf{27.2} & 46.1 & 53.9 & 38.6 & 61.6 & 41.4 & \textbf{23.1} & 42.3 & 52.0 \\
& + KW ($1\times$) & 41.8 & 64.5 & 45.9 & 26.6 & 45.5 & 53.0 & 38.4 & 61.4 & 41.2 & 22.2 & 42.0 & 51.6 \\
& + KW ($4\times$) & \textbf{42.4} & \textbf{65.4} & \textbf{46.3} & \textbf{27.2} & \textbf{46.2} & \textbf{54.6} & \textbf{38.9} & \textbf{62.0} & \textbf{41.5} & 22.7 & \textbf{42.6} & \textbf{53.1} \\
\cline{2-14}
& MobileNetV2 ($1.0\times$) & 33.8 & 55.2 & 35.8 & 19.7 & 36.5 & 44.4 & 31.7 & 52.4 & 33.3 & 16.4 & 34.4 & 43.7 \\
& + DY-Conv ($4\times$) & 37.0 & 58.6 & 40.3 & 21.9 & 40.1 & 47.9 & 34.1 & 55.7 & 36.1 & 18.6 & 37.1 & 46.3 \\
& + ODConv ($4\times$) & 37.2 & 59.4 & 39.9 & 22.6 & 40.0 & 48.0 & 34.5 & 56.4 & 36.3 & 19.3 & 37.3 & 46.8 \\
& + KW ($1\times$) & 36.4 & 58.3 & 39.2 & 22.0 & 39.6 & 47.0 & 33.7 & 55.1 & 35.7 & 18.9 & 36.7 & 45.6 \\
& + KW ($4\times$) & \textbf{38.0} & \textbf{60.0} & \textbf{40.8} & \textbf{23.1} & \textbf{40.7} & \textbf{50.0} & \textbf{34.9} & \textbf{56.6} & \textbf{37.0} & \textbf{19.4} & \textbf{37.9} & \textbf{47.8} \\
\cline{2-14}
& ConvNeXt-Tiny & 43.4 & 65.8 & 47.7 & 27.6 & 46.8 & 55.9 & 39.7 & 62.6 & 42.4 & 23.1 & 43.1 & 53.7 \\
& + KW ($3/4\times$) & 44.1 & 66.8 & 48.4 & 29.7 & 47.4 & 56.7 & 40.2 & 63.6 & 43.0 & \textbf{24.8} & 43.6 & 54.3 \\
& + KW ($1\times$) & \textbf{44.8} & \textbf{67.7} & \textbf{48.9} & \textbf{29.8} & \textbf{48.3} & \textbf{57.3} & \textbf{40.6} & \textbf{64.4} & \textbf{43.4} & 24.7 & \textbf{44.1} & \textbf{54.8} \\
\hline
\end{tabular}
}
\end{small}
\end{center}
\vskip -0.1 in
\end{table*}

%

\textbf{Results Comparison with Advanced Training Strategy.} %
To better explore the potential of our method, we further adopt the advanced training strategy recently proposed in ConvNeXt~\citep{CNN_ConvNeXt}, with a longer training schedule of 300 epochs and aggressive augmentations, for comparisons on the ResNet18, ResNet50 and ConvNeXt-Tiny backbones. From the results summarized in Table~\ref{table:classification_cnn_300epoch}, we can observe:
(1) KW ($4\times$) gets the best results on the ResNet18 backbone, bringing a significant top-1 gain of 5.61\%. Even with 36.45\%$|$65.10\% parameter reduction, KW ($1/2\times$)$|$KW($1/4\times$) brings 2.89\%$|$2.29\% top-1 accuracy gain to the baseline model;
(2) On the larger ResNet50 backbone, vanilla dynamic convolution DY-Conv ($4\times$) brings small top-1 gain to the baseline model, but KW ($1/2\times, 1\times, 4\times$) gets promising top-1 gains. Even with 30.99\% parameter reduction against the baseline model, KW ($1/2\times$) attains a top-1 gain of 0.86\%. KW ($4\times$) outperforms DY-Conv ($4\times$) and ODConv ($4\times$) by 2.05\% and 0.43\% top-1 gain, respecrtively.
Besides, we also apply our method to the ConvNeXt-Tiny backbone to investigate its performance on the state-of-the-art ConvNet architecture. Results show that our method generalizes well on ConvNeXt-Tiny. Surprisingly, KW($3/4\times$) with 14.20\% parameter reduction to the baseline model gets 0.16\% top-1 gain.

\begin{table}[ht]
\vskip 0.0 in
\caption{Results comparison on ImageNet with the MobileNetV2 backbones trained for 150 epochs.} 
\label{table:classification_mobilenetv2}
\vskip -0.1 in
\begin{center}
\begin{small}
\resizebox{0.95\linewidth}{!}{
\begin{tabular}[b]{l|c|c|c}
\hline
Models & Params & Top-1 Acc (\%) & Top-5 Acc (\%) \\
\hline
MobileNetV2 ($0.5\times$) & 1.97M & 64.30 & 85.21 \\
+ DY-Conv ($4\times$) & 4.57M & 69.05 ($\uparrow$4.75) & 88.37 ($\uparrow$3.16) \\
+ ODConv ($4\times$) & 4.44M & 70.01 ($\uparrow$5.71) & 89.01 ($\uparrow$3.80)  \\
+ KW ($1/2\times$) & 1.47M & 65.19 ($\uparrow$0.89) & 85.98 ($\uparrow$0.77) \\
+ KW ($1\times$) & 2.85M & 68.29 ($\uparrow$3.99) & 87.93 ($\uparrow$2.72)  \\
+ KW ($4\times$) & 4.65M & \textbf{70.26} ($\uparrow$\textbf{5.96}) & \textbf{89.19} ($\uparrow$\textbf{3.98}) \\
\hline
MobileNetV2 ($1.0\times$) & 3.50M & 72.02 & 90.43  \\
+ DY-Conv ($4\times$) & 12.40M & 74.94 ($\uparrow$2.92) & 91.83 ($\uparrow$1.40)  \\
+ ODConv ($4\times$) & 11.52M & 75.42 ($\uparrow$3.40) & 92.18 ($\uparrow$1.75)   \\
+ KW ($1/2\times$) & 2.65M & 72.59 ($\uparrow$0.57) & 90.71 ($\uparrow$0.28)  \\
+ KW ($1\times$) & 5.17M & 74.68 ($\uparrow$2.66) & 91.90 ($\uparrow$1.47)  \\
+ KW ($4\times$) & 11.38M & \textbf{75.92} ($\uparrow$\textbf{3.90}) & \textbf{92.22} ($\uparrow$\textbf{1.79})  \\
\hline
\end{tabular}
}
\end{small}
\end{center}
\vskip -0.1 in
\end{table}

\textbf{Results Comparison on MobileNets.} We further apply our method to MobileNetV2 ($1.0\times$, $0.5\times$) to validate its effectiveness on lightweight ConvNet architectures.
Since the lightweight MobileNetV2 backbones have lower model capacity compared to ResNet and ConvNeXt backbones, we don't use aggressive augmentations.
The results are shown in Table~\ref{table:classification_mobilenetv2}. We can see that
%
our method can strike a favorable trade-off between parameter efficiency and representation power for lightweight ConvNets as well as larger ones. Even on the lightweight MobileNetV2 ($1.0\times, 0.5\times$) with 3.50M$|$1.97M parameters, KW ($1/2\times$) can reduce the model size by 24.29\%$|$25.38\% while bringing top-1 gain of 0.57\%$|$0.89\%. Similar to the results on the ResNet18 and ResNet50 backbones, KW ($4\times$) also obtains the best results on both MobileNetV2 ($1.0\times$) and MobileNetV2 ($0.5\times$).


\subsection{Detection and Segmentation on MS-COCO Dataset}
Next, to evaluate the generalization ability of the classification backbone models trained by our method to downstream object detection and instance segmentation tasks, we conduct comparative experiments on MS-COCO dataset.

\textbf{Experimental Setup.} We adopt Mask R-CNN~\citep{DL_Instance_MaskRCNN} as the detection framework, ResNet50 and MobileNetV2 ($1.0\times$) built with different dynamic convolution methods as the backbones which are pre-trained on ImageNet dataset. Then, all the models are trained with standard $1\times$ schedule 
on MS-COCO dataset.
For fair comparisons, we adopt the same settings including data processing pipeline and hyperparameters for all the models. Experimental details are described~\textit{in the Appendix}.

\begin{table}[ht]
\vskip 0. in
\caption{Effect of kernel partition.}
\vskip -0.1 in
\label{table:ablation_partition}
\begin{center}
\begin{small}
\resizebox{0.95\linewidth}{!}{
\begin{tabular}{l|c|c|c|c}
\hline
Models & Kernel Partition & Params & Top-1 Acc (\%) & Top-5 Acc (\%)\\
\hline
ResNet18 & - & 11.69M & 70.44 & 89.72 \\
\hline
\multirow{2}{*}{+ KW ($1\times$)} & \Checkmark & 11.93M & \textbf{74.77} ($\uparrow$\textbf{4.33}) & \textbf{92.13} ($\uparrow$\textbf{2.41}) \\
 & \XSolidBrush \ & 11.78M & 70.49 ($\uparrow$0.05) & 89.84 ($\uparrow$0.12) \\
\hline
\end{tabular}
}
\end{small}
\end{center}
\vskip -0.1 in
\end{table}

\textbf{Results Comparison.} Table~\ref{table:detection} summarizes all results. 
For Mask R-CNN with the ResNet50 backbone, we observe a similar trend to the main experiments on ImageNet dataset: KW ($4\times$) outperforms DY-Conv ($4\times$) and ODConv ($4\times$) on both object detection and instance segmentation tasks. Our KW ($1\times$) brings an AP improvement of 2.2\%$|$2.0\% on object detection and instance segmentation tasks, which is even on par with ODConv ($4\times$).
On the MobileNetV2 ($1.0\times$) backbone, our method yields consistent high accuracy improvements to the baseline, and KW ($4\times$) achieves the best results.
With the ConvNeXt-Tiny backbone, the performance gains of KW ($1\times$) and KW ($3/4\times$) to the baseline model become more pronounced on MS-COCO dataset compared to those on ImageNet dataset, showing good transfer learning ability of our method. 

\subsection{Ablation Studies}
For a better understanding of our method, we further conduct a lot of ablative experiments on ImageNet dataset, 
using the advanced training strategy proposed in ConvNeXt~\citep{CNN_ConvNeXt}.

\textbf{Effect of Kernel Partition.} 
Thanks to the kernel partition component, our method can apply dense kernel assembling with a large number of kernel cells. In Table~\ref{table:ablation_partition}, we provide the ablative experiments on the ResNet18 backbone with KW ($1\times$) to study the efficacy of kernel partition. We can see, when removing kernel partition, the top-1 gain of our method to the baseline model sharply decreases from 4.33\% to 0.05\%, demonstrating its great importance to our method.

\textbf{Effect of Warehouse Sharing Range in terms of Layer.}
\begin{table}[ht]
\begin{minipage}[ht]{1.0\linewidth}
\vskip 0.05 in
\caption{Effect of warehouse sharing range in terms of layer.}
\vskip -0.1 in
\label{table:ablation_sharingscale}
\begin{center}
\begin{small}
\resizebox{0.99\linewidth}{!}{
\begin{tabular}{l|c|c|c|c}
\hline
Models & Warehouse Sharing Range & Params & Top-1 Acc (\%) & Top-5 Acc (\%)\\
\hline
ResNet18 & - & 11.69M & 70.44 & 89.72 \\
\hline
\multirow{3}{*}{+ KW ($1\times$)} & Within each stage & 11.93M & \textbf{74.77} ($\uparrow$\textbf{4.33}) & \textbf{92.13} ($\uparrow$\textbf{2.41}) \\
 & Within each layer & 11.81M & 74.34 ($\uparrow$3.90) & 91.82 ($\uparrow$2.10) \\
 & Without sharing & 11.78M & 72.49 ($\uparrow$2.05) & 90.81 ($\uparrow$1.09) \\
\hline
\end{tabular}
}
\end{small}
\end{center}
\end{minipage}
\vskip 0.2 in
\begin{minipage}[ht]{1.0\linewidth}
\vskip 0.0 in
\caption{Effect of warehouse sharing range in terms of static kernel dimensions.}
\vskip -0.1 in
\label{table:ablation_sharingshape}
\begin{center}
\begin{small}
\resizebox{0.99\linewidth}{!}{
\begin{tabular}{l|c|c|c|c}
\hline
Models & Warehouse sharing range & Params & Top-1 Acc (\%) & Top-5 Acc (\%)\\
\hline
ResNet50 & - & 25.56M & 78.44 & 94.24 \\
\hline
\multirow{2}{*}{+ KW ($1\times$)} & Different dimensioned kernels & 28.05M & \textbf{80.38} ($\uparrow$\textbf{1.94}) & \textbf{95.27} ($\uparrow$\textbf{1.03}) \\
& Same dimensioned kernels & 26.95M & 79.80 ($\uparrow$1.36) & 95.01 ($\uparrow$0.77) \\
\hline
\end{tabular}
}
\end{small}
\end{center}
\end{minipage}
\vskip 0.2 in
\begin{minipage}[ht]{1.0\linewidth}
\vskip 0.0 in
\caption{Effect of different attention functions.}
\vskip -0.1 in
\label{table:ablation_function}
\begin{center}
\begin{small}
\resizebox{0.99\linewidth}{!}{
\begin{tabular}{l|c|c|c|c}
\hline
Models & Attention Functions & Params & Top-1 Acc (\%) & Top-5 Acc (\%)\\
\hline
ResNet18 & - & 11.69M & 70.44 & 89.72 \\
\hline
\multirow{4}{*}{+ KW ($1\times$)}
& $z_{ij}/\sum^{n}_{p=1}{|z_{ip}|}$ (ours) & 11.93M & \textbf{74.77} ($\uparrow$\textbf{4.33}) & \textbf{92.13} ($\uparrow$\textbf{2.41}) \\
& Softmax & 11.93M & 72.67 ($\uparrow$2.23) & 90.82 ($\uparrow$1.10) \\
& Sigmoid & 11.93M & 72.09 ($\uparrow$1.65) & 90.70 ($\uparrow$0.98) \\
& $max(z_{ij},0)/\sum^{n}_{p=1}{|z_{ip}|}$ & 11.93M & 72.74 ($\uparrow$2.30) & 90.86 ($\uparrow$1.14) \\
\hline
\end{tabular}
}
\end{small}
\end{center}
\end{minipage}
\vskip -0.1 in
\end{table}
To validate the effectiveness of the warehouse construction-with-sharing component,
we first study warehouse sharing range in terms of layer by performing ablative experiments on the ResNet18 backbone with KW ($1\times$).
From the results shown in Table~\ref{table:ablation_sharingscale}, 
we can see that when sharing warehouse in a wider range, our method brings larger performance improvement to the baseline model. This clearly indicates that explicitly enhancing convolutional parameter dependencies either within the same or across the neighboring layers can strengthen the capacity of a ConvNet.


\textbf{Effect of Warehouse Sharing Range in terms of Static Kernel Dimensions.}
For modern ConvNet backbones, a convolutional block in the same-stage mostly contains several static kernels having different dimensions ($k\times k\times c\times f$).
Next, we perform ablative experiments on the ResNet50 backbone to study the effect of warehouse sharing range in terms of static kernel dimensions in the same-stage convolutional layers of a ConvNet.
Results are summarized in Table~\ref{table:ablation_sharingshape}, showing that warehouse sharing across convolutional layers having different dimensioned kernels performs better than only sharing warehouse across convolutional layers having the same dimensioned kernels.
Combining the results in Table~\ref{table:ablation_sharingscale} and Table~\ref{table:ablation_sharingshape}, we can conclude that enhancing the warehouse sharing between more kernel cells tends to achieve better performance for our method.
%

%
\begin{table}[ht]
\vskip 0.05 in
\begin{minipage}[ht]{1.0\linewidth}
\caption{Effect of attentions initialization strategy.}
\vskip -0.1 in
\label{table:ablation_temperature}
\begin{center}
\begin{small}
\resizebox{0.99\linewidth}{!}{
\begin{tabular}{l|c|c|c|c}
\hline
Models & Attentions Initialization Strategy & Params & Top-1 Acc (\%) & Top-5 Acc (\%)\\
\hline
ResNet18 & - & 11.69M & 70.44 & 89.72 \\
\hline
\multirow{2}{*}{+ KW ($1\times$)} & \Checkmark & 11.93M & \textbf{74.77} ($\uparrow$\textbf{4.33}) & \textbf{92.13} ($\uparrow$\textbf{2.41}) \\
 & \XSolidBrush & 11.93M & 73.39 ($\uparrow$2.95) & 91.24 ($\uparrow$1.52) \\
\hline
\end{tabular}
}
\end{small}
\end{center}
\end{minipage}
\vskip 0.2 in
\vskip 0.0 in
\begin{minipage}[ht]{1.0\linewidth}
\caption{Applying CAF to other dynamic convolution methods.}
\vskip -0.1 in
\label{table:attention_function}
\begin{center}
\begin{small}
\resizebox{0.99\linewidth}{!}{
\begin{tabular}{l|c|c|c|c}
\hline
Models & Params & Attention Function & Top-1 Acc (\%) & Top-5 Acc (\%) \\
\hline
ResNet18 & 11.69M & - & 70.44 & 89.72 \\
\hline
\multirow{2}*{\makecell[l]{+ DY-Conv ($4\times$)}} & \multirow{2}*{\makecell[c]{45.47M}} & Softmax & \textbf{73.82} ($\uparrow$\textbf{3.38}) & \textbf{91.48} ($\uparrow$\textbf{1.76}) \\
& & Our CAF & 73.74 ($\uparrow$3.30) & 91.45 ($\uparrow$1.73) \\
\hline
\multirow{2}*{\makecell[l]{+ ODConv ($4\times$)}} & \multirow{2}*{\makecell[c]{44.90M}} & Softmax & \textbf{74.45} ($\uparrow$\textbf{4.01}) & \textbf{91.67} ($\uparrow$\textbf{1.95}) \\
& & Our CAF & 74.27 ($\uparrow$3.83) & 91.62 ($\uparrow$1.90) \\
\hline
\multirow{2}*{\makecell[l]{+ KW ($1\times$)}} & \multirow{2}*{\makecell[c]{11.93M}} & Softmax & 72.67 ($\uparrow$2.23) & 90.82 ($\uparrow$1.10) \\
& & Our CAF & \textbf{74.77} ($\uparrow$\textbf{4.33}) & \textbf{92.13} ($\uparrow$\textbf{2.41}) \\
\hline
\multirow{2}*{\makecell[l]{+ KW ($4\times$)}} & \multirow{2}*{\makecell[c]{45.86M}} & Softmax & 74.31 ($\uparrow$3.87) & 91.75 ($\uparrow$2.03) \\
& & Our CAF & \textbf{76.05} ($\uparrow$\textbf{5.61}) & \textbf{92.68} ($\uparrow$\textbf{2.96}) \\
\hline
\end{tabular}
}
\end{small}
\end{center}
\end{minipage}
\vskip 0.2 in
\begin{minipage}[ht]{1.0\linewidth}
\vskip 0.0 in
\caption{Applying KW to Vision Transformer backbones.}
\label{table:transformer}
\vskip -0.1 in
\begin{center}
\begin{small}
\resizebox{0.8\linewidth}{!}{
\begin{tabular}{l|c|c|c}
\hline
Models & Params & Top-1 Acc (\%) & Top-5 Acc (\%) \\
\hline
DeiT-Small & 22.06M & 79.78 & 94.99 \\
+ KW ($3/4\times$) & 19.23M & 79.94 ($\uparrow$0.16) & 95.05 ($\uparrow$0.06) \\
+ KW ($1\times$) & 24.36M & \textbf{80.63} ($\uparrow$\textbf{0.85}) & \textbf{95.24} ($\uparrow$\textbf{0.25}) \\
\hline
DeiT-Tiny & 5.72M & 72.13 & 91.32 \\
+ KW ($1\times$) & 6.39M & 73.56 ($\uparrow$1.43) & 91.82 ($\uparrow$0.50) \\
+ KW ($4\times$) & 20.44M & \textbf{76.51} ($\uparrow$\textbf{4.38}) & \textbf{93.05} ($\uparrow$\textbf{1.73}) \\
\hline
\end{tabular}
}
\end{small}
\end{center}
\end{minipage}
\vskip -0.1 in
\end{table}

\textbf{Effect of Different Attention Functions.} 
Recall that our method relies on the proposed contrasting-driven attention function (CAF). 
To explore its role, we also conduct ablative experiments on the ResNet18 backbone to compare the performance of KernelWarehouse using different attention functions. According to the results in Table~\ref{table:ablation_function}, the top-1 gain of our CAF $z_{ij}/\sum^{n}_{p=1}{|z_{ip}|}$ against popular attention function Softmax$|$Sigmoid reaches 2.10\%$|$2.68\%, and our CAF also outperforms another counterpart $max(z_{ij},0)/\sum^{n}_{p=1}{|z_{ip}|}$ by 2.03\% top-1 gain. These experiments well validate the efficacy and importance of two design principles (clarified in the Method section) of our CAF.

\textbf{Effect of Attentions Initialization Strategy.} 
To help the optimization of KernelWarehouse in the early training stage, our CAF uses binary $\beta_{ij}$ with temperature $\gamma$ to initialize the scalar attentions. Next, we perform experiments on the ResNet18 backbone to study its role. As shown in Table~\ref{table:ablation_temperature}, this attentions initialization strategy is essential to our method in learning contrasting and diverse attention relationships between linear mixtures and kernel cells, bringing 1.38\% top-1 gain to the baseline model with KW ($1\times$).

\textbf{Applying CAF to Other Dynamic Convolution Methods.}
In Table~\ref{table:attention_function}, we provide experimental results for applying the proposed attention function CAF to existing top-performing dynamic convolution methods DY-Conv and ODConv. We can see that CAF gets slight model accuracy drop compared to the original Softmax function. This is because CAF is customized to fit three unique optimization properties of KernelWarehouse, as we discussed in the Method section. 

\begin{figure*}[ht]
\vskip 0.0 in 
    \begin{subfigure}[b]{0.24\linewidth}
        \begin{center}
            \centerline{\includegraphics[width=\textwidth]{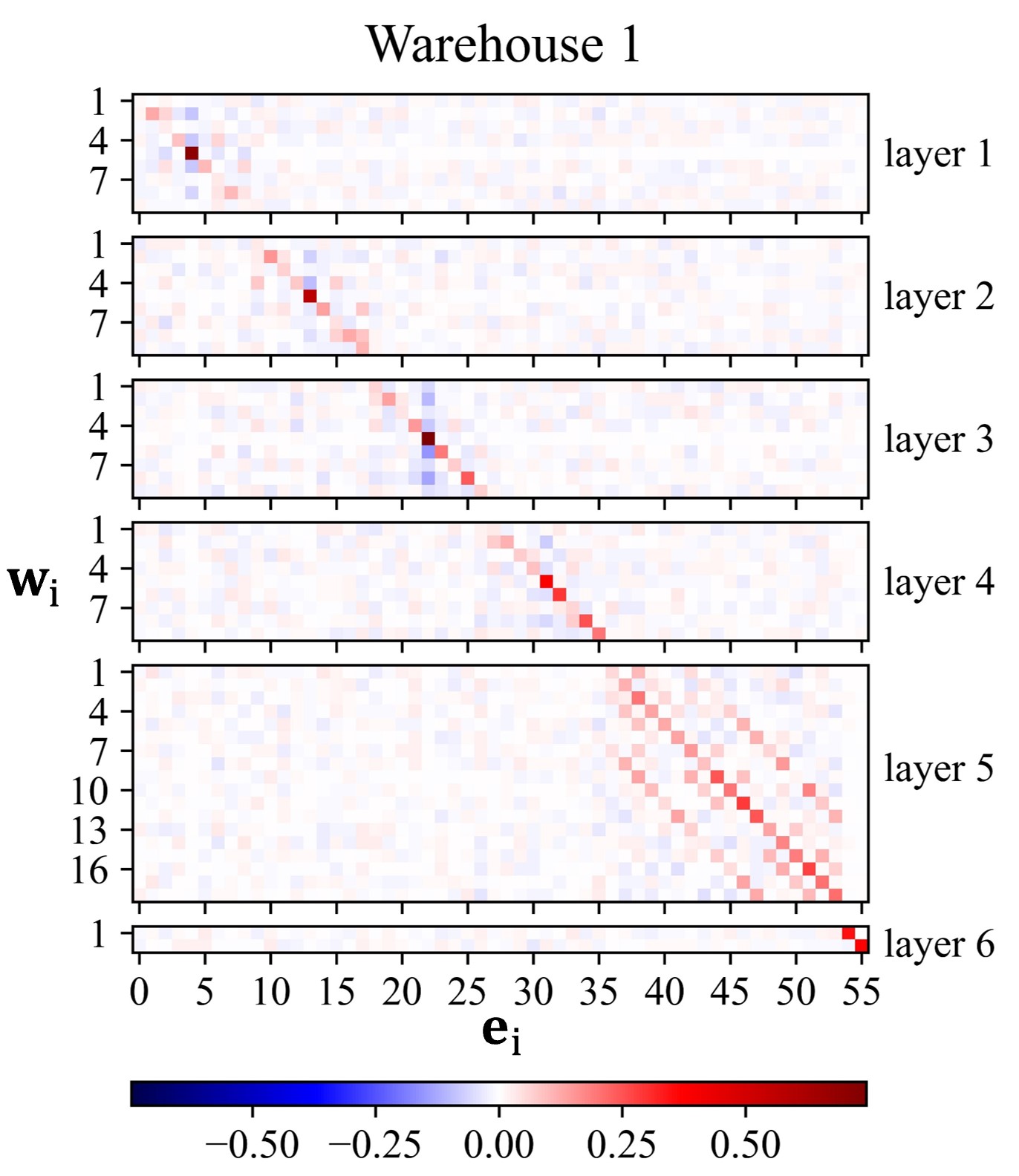}}
        \end{center}
    \end{subfigure}
    \hfill
    \begin{subfigure}[b]{0.24\linewidth}
        \begin{center}
            \centerline{\includegraphics[width=\textwidth]{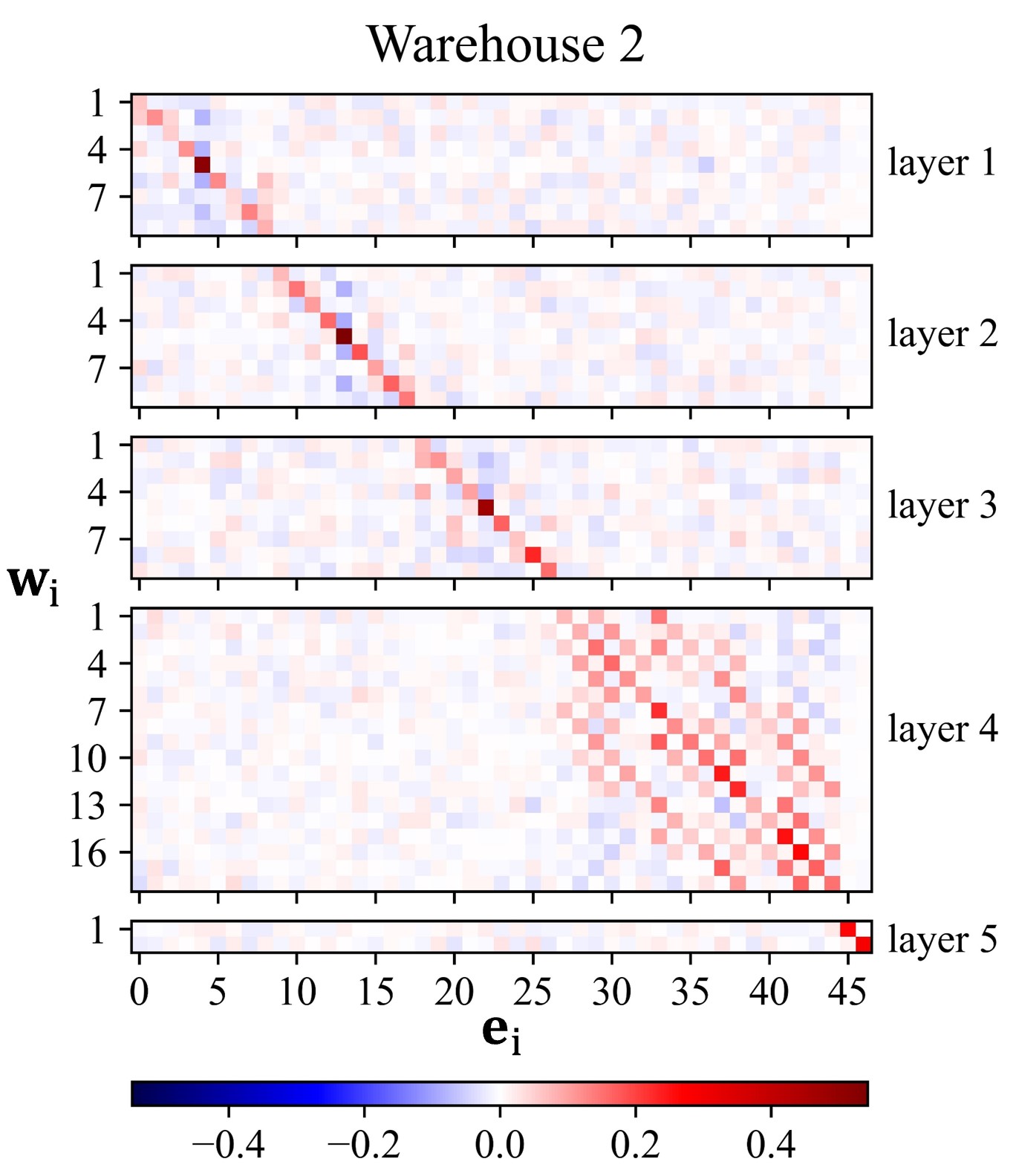}}
        \end{center}
    \end{subfigure}
     \hfill
    \begin{subfigure}[b]{0.24\linewidth}
        \begin{center}
            \centerline{\includegraphics[width=\textwidth]{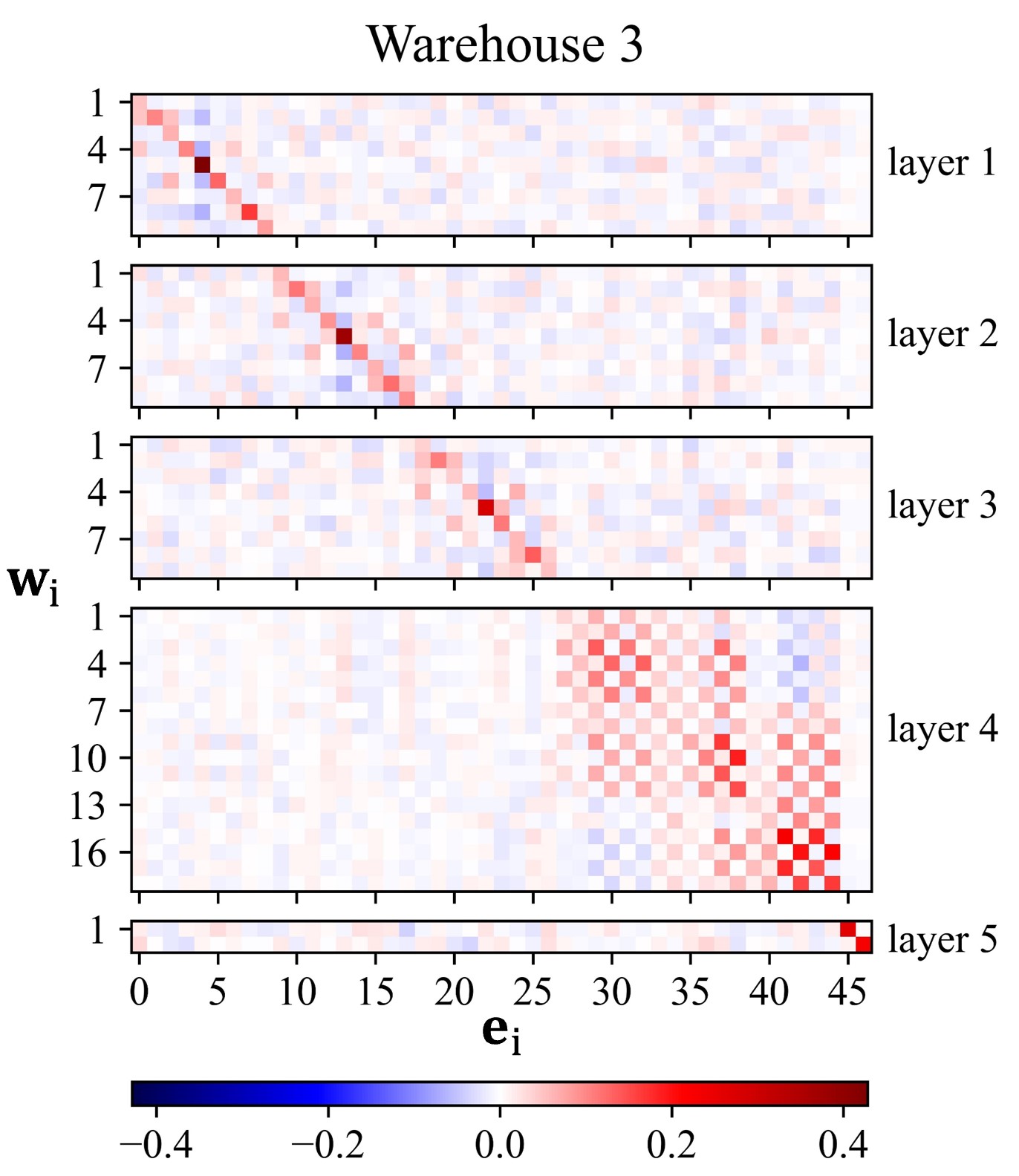}}
        \end{center}
    \end{subfigure}
    \hfill
    \begin{subfigure}[b]{0.24\linewidth}
        \begin{center}
        \centerline{\includegraphics[width=\textwidth]{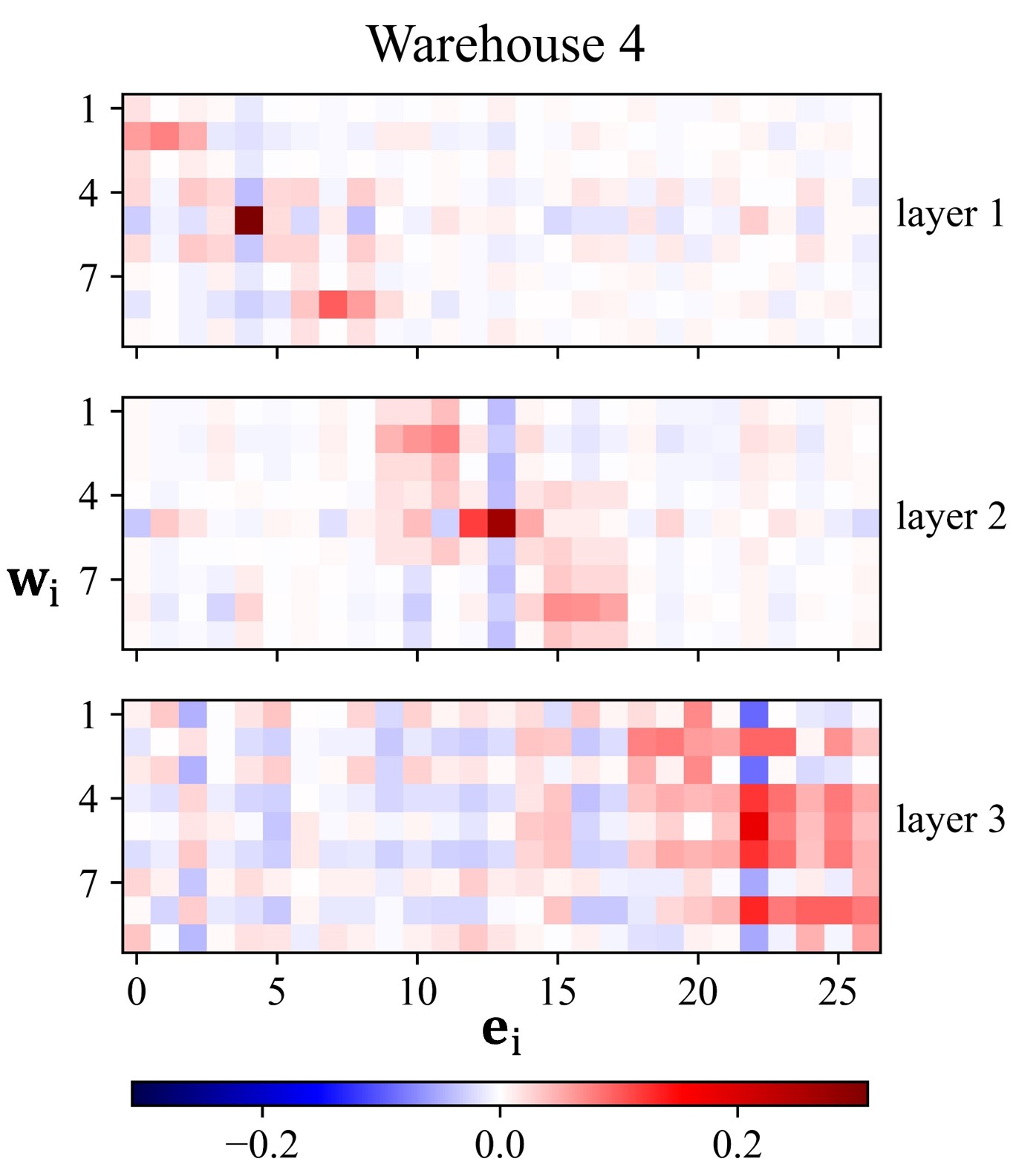}}
        \end{center}
    \end{subfigure}
    \vskip -0.2 in
    \caption{Visualization of statistical mean values of learnt attention $\alpha_{ij}$ in each warehouse. The results are obtained from the pre-trained ResNet18 model with KW ($1\times$) using the whole ImageNet validation set. Best viewed with zoom-in.
    }
    \label{figure:attention_resnet18_heatmap}
\end{figure*}

\textbf{Applying KW to Vision Transformer Backbones.}
One interesting question is whether our method can be generalized to Vision Transformer backbones. We study it by applying our method to two popular DeiT~\citep{SelfAttention_DeiT} backbones. 
\begin{table}[ht]
\vskip -0.0 in
\caption{Runtime model speed (frames per second (FPS)) comparison. All models are tested on an NVIDIA TITAN X GPU (with batch size 100 and input image size 224$\times$224).}
\vskip -0.1 in
\label{table:ablation_speed}
\begin{center}
\begin{small}
\resizebox{0.95\linewidth}{!}{
\begin{tabular}{l|c|c|c}
\hline
Models & Params & Top-1 Acc (\%) & Runtime Speed (FPS) \\
\hline
ResNet50 & 25.56M & 78.44 & 647.0 \\
+ DY-Conv ($4\times$) & 100.88M & 79.00 ($\uparrow$0.56) & 322.7 \\
+ ODConv ($4\times$) & 90.67M & 80.62 ($\uparrow$2.18) &142.3 \\
+ KW ($1/2\times$) & 17.64M & 79.30 ($\uparrow$0.86) & 227.8 \\
+ KW ($1\times$) & 28.05M & 80.38 ($\uparrow$1.94) & 265.4 \\
+ KW ($4\times$) & 102.02M & 81.05 ($\uparrow$2.61) & 191.1 \\
\hline
MobileNetV2 ($1.0\times$) & 3.50M & 72.02 & 1410.8 \\
+ DY-Conv ($4\times$) & 12.40M & 74.94 ($\uparrow$2.92) & 862.4 \\
+ ODConv ($4\times$) & 11.52M & 75.42 ($\uparrow$3.40) & 536.5 \\
+ KW ($1/2\times$) & 2.65M & 72.59 ($\uparrow$0.57) & 926.0 \\
+ KW ($1\times$) & 5.17M & 74.68 ($\uparrow$2.66) & 798.7 \\
+ KW ($4\times$) & 11.38M & 75.92 ($\uparrow$3.90) & 786.9 \\
\hline
\end{tabular}
}
\hfill
\end{small}
\end{center}
\vskip -0.15 in
\end{table}In the experiments, each of partitioned cells of weight matrices for ``value and MLP” layers of each DeiT backbone is represented as a linear mixture of kernel warehouse shared across multiple multi-head self-attention blocks and MLP blocks, except the ``query” and ``key” matrices which are used to compute self-attention. From the results shown in Table~\ref{table:transformer}, it can be seen that: (1) With a small parameter budget, e.g., $b=3/4$, KW gets slightly improved model accuracy while reducing model size of DeiT-Small; (2) With a larger parameter budget, e.g., $b=4$, KW can significantly improve model accuracy, bringing 4.38\% top-1 accuracy gain to DeiT-Tiny; (3) These performance trends are similar to our results on ConvNet backbones (see Table~\ref{table:classification_cnn_300epoch} and Table~\ref{table:classification_mobilenetv2}), demonstrating the appealing generalization ability of our method to different neural network architectures.

\textbf{Visualization of Learnt Attentions.}
In Figure~\ref{figure:attention_resnet18_heatmap}, we provide visualization results to analyze the statistical mean values of learnt attention $\alpha_{ij}$, for better understanding our method. The results are obtained from the pre-trained ResNet18 model with KW ($1\times$). We can observe:
(1) Each linear mixture can learn its own distribution of scalar attentions for different kernel cells;
(2) In each warehouse, the maximum value of $\alpha_{ij}$ in each row mostly appears in the diagonal line throughout the whole warehouse. This indicates that our attentions initialization strategy can help our method to build rich one-to-one relationships between linear mixtures and kernel cells; 
(3) The attentions $\alpha_{ij}$ with higher absolute values for linear mixtures in the same layer have more overlaps than linear mixtures across different layers. This indicates that parameter dependencies within the same kernel are stronger than those across neighboring layers, which can be learned by our method.

%

\textbf{Runtime Model Speed and Discussion.} The above experimental results demonstrate that KernelWarehouse strikes a favorable trade-off between parameter efficiency and representation power. 
Table~\ref{table:ablation_speed} further compares the runtime model speeds of 
different dynamic convolution methods. We can see, under the same convolutional parameter budget $b$, the runtime model speed of KernelWarehouse: (1) is slower than DY-Conv (vanilla dynamic convolution) on larger backbones like ResNet50; (2) is at a similar level to DY-Conv on lightweight backbones like MobileNetV2; (3) is always faster than ODConv (existing best-performing dynamic convolution method). The model speed gap of KernelWarehouse to DY-Conv is primarily due to the dense attentive mixture and assembling operations at the same-stage convolutional layers having a shared warehouse. Thanks to the parallel property of these operations, there are some potential optimization strategies to alleviate this gap. Given a ConvNet model pre-trained with KernelWarehouse, a customized optimization strategy is to adaptively allocate available Tensor Cores and CUDA Cores for a better trade-off of memory-intensive and compute-intensive operations in KernelWarehouse at different convolutional layers. Another potential optimization strategy is to reduce the number of kernel cells in a shared warehouse or reduce the warehouse sharing range, under the condition that the desired model size and accuracy can be still reached in real applications. 

\textit{More experiments and visualizations are in the Appendix.}  


\section{Conclusion}

In this paper, we rethink the design of dynamic convolution and present KernelWarehouse. As a more general form of dynamic convolution, KernelWarehouse can improve the performance of modern ConvNets while enjoying parameter efficiency. Experiments on ImageNet and MS-COCO datasets show its great potential. We hope our work would inspire future research in dynamic convolution.


\section*{Acknowledgements}

This work was done when Chao Li was an intern at Intel Labs China, supervised by Anbang Yao who led the project and the paper writing. We thank Intel Data Center \& AI group’s great support of their DGX-2 and DGX-A100 servers for training large models in this project.

\section*{Impact Statement}

This paper presents KernelWarehouse, a more general form of dynamic convolution, which advances dynamic convolution research towards substantially better parameter efficiency and representation power. 
As far as we know, our work does not have any ethical impacts and potential societal consequences.

\nocite{langley00}

\bibliography{example_paper}
\bibliographystyle{icml2024}

\newpage
\appendix
\onecolumn
\section{\textbf{Appendix}}

\subsection{Datasets and Implementation Details}

\subsubsection{Image Classification on ImageNet}

Recall that we use ResNet~\citep{CNN_ResNet}, MobileNetV2~\citep{CNN_MobileNetV2} and ConvNeXt~\citep{CNN_ConvNeXt} families for the main experiments on ImageNet dataset~\citep{Dataset_ImageNet}, which consists of over 1.2 million training images and 50,000 validation images with 1,000 object categories.
We use an input image resolution of 224$\times$224 for both training and evaluation. All the input images are standardized with mean and standard deviation per channel.
For evaluation, we report top-1 and top-5 recognition rates of a single 224$\times$224 center crop on the ImageNet validation set.
All the experiments are performed on the servers having 8 GPUs. Specifically, the models of ResNet18, MobileNetV2 ($1.0\times$), MobileNetV2 ($0.5\times$) are trained on the servers with 8 NVIDIA Titan X GPUs. The models of ResNet50, ConvNeXt-Tiny are trained on the servers with 8 NVIDIA Tesla V100-SXM3 or A100 GPUs. The training setups for different models are as follows.

\textbf{Training setup for ResNet models with the traditional training strategy.}
All the models are trained by the stochastic gradient descent (SGD) optimizer for 100 epochs, with a batch size of 256, a momentum of 0.9 and a weight decay of 0.0001. The initial learning rate is set to 0.1 and decayed by a factor of 10 for every 30 epoch. Horizontal flipping and random resized cropping are used for data augmentation. For KernelWarehouse, the temperature $\tau$ linearly reduces from 1 to 0 in the first 10 epochs.

\textbf{Training setup for ResNet and ConvNeXt models with the advanced training strategy.}
Following the settings of ConvNeXt~\citep{CNN_ConvNeXt}, all the models are trained by the AdamW optimizer with $\beta_{1}=0.9, \beta_{2}=0.999$ for 300 epochs, with a batch size of 4096, a momentum of 0.9 and a weight decay of 0.05. The initial learning rate is set to 0.004 and annealed down to zero following a cosine schedule. Randaugment~\citep{augmentation_randaugment}, mixup~\citep{augmentation_mixup}, cutmix~\citep{augmentation_cutmix}, random erasing~\citep{augmentation_randerazing} and label smoothing~\citep{augmentation_labelsmoothing} are used for augmentation. For KernelWarehouse, the temperature $\tau$ linearly reduces from 1 to 0 in the first 20 epochs.

\textbf{Training setup for MobileNetV2 models.}
All the models are trained by the SGD optimizer for 150 epochs, with a batch size of 256, a momentum of 0.9 and a weight decay of 0.00004. The initial learning rate is set to 0.1 and annealed down to zero following a cosine schedule. Horizontal flipping and random resized cropping are used for data augmentation.
For KernelWarehouse, the temperature $\tau$ linearly reduces from 1 to 0 in the first 10 epochs.

\subsubsection{Object Detection and Instance Segmentation on MS-COCO}

Recall that we conduct comparative experiments for object detection and instance segmentation on the MS-COCO 2017 dataset~\citep{Dataset_MS_COCO}, which contains 118,000 training images and 5,000 validation images with 80 object categories.
We adopt Mask R-CNN as the detection framework, ResNet50 and MobileNetV2 (1.0$\times$) built with different dynamic convolution methods as the backbones which are pre-trained on ImageNet dataset.
All the models are trained with a batch size of 16 and standard 1$\times$ schedule on the MS-COCO dataset using multi-scale training. The learning rate is decreased by a factor of 10 at the 8$^{th}$ and the 11$^{th}$ epoch of total 12 epochs.
For a fair comparison, we adopt the same settings including data processing pipeline and hyperparameters for all the models. All the experiments are performed on the servers with 8 NVIDIA Tesla V100 GPUs.
The attentions initialization strategy is not used for KernelWarehouse during fine-tuning to avoid disrupting the learnt relationships of the pre-trained models between kernel cells and linear mixtures.
For evaluation, we report both bounding box Average Precision (AP) and mask AP on the MS-COCO 2017 validation set, including $AP_{50}$, $AP_{75}$ (AP at different IoU thresholds) and $AP_{S}$, $AP_{M}$, $AP_{L}$ (AP at different scales).

\subsection{Visualization Examples of Attentions Initialization Strategy}

\begin{figure}[ht]
    \vskip 0.05 in
    \begin{center}
        \centerline{\includegraphics[width=0.75\linewidth]{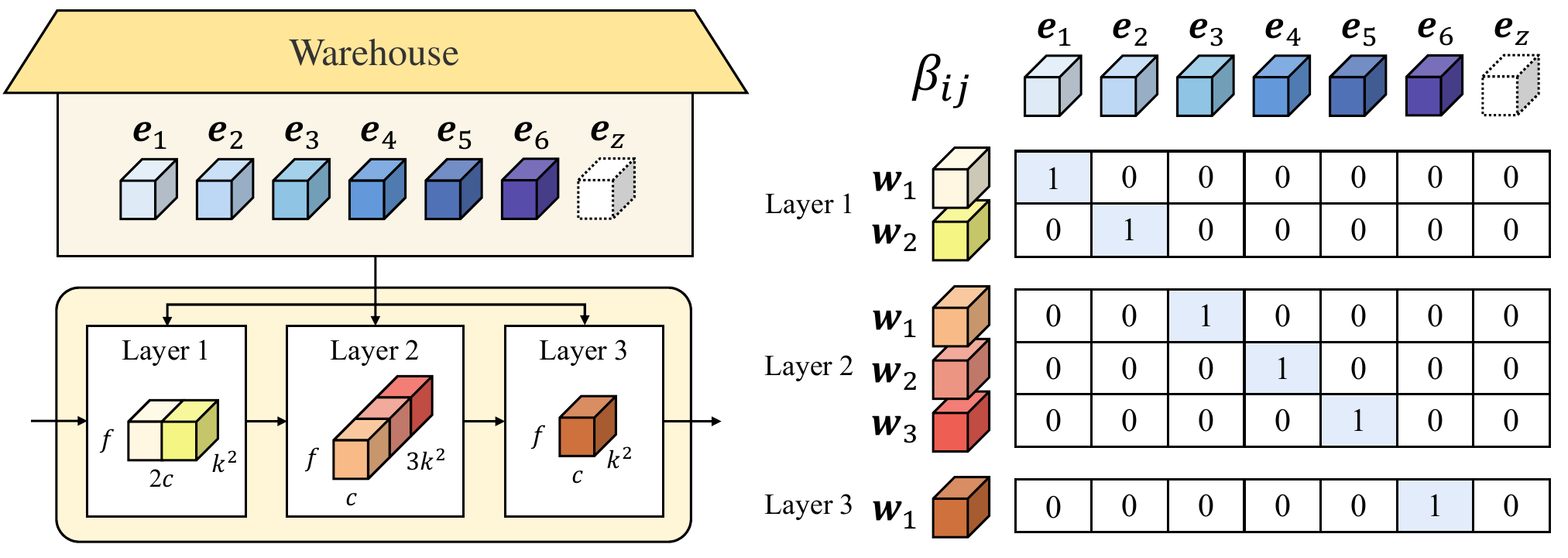}}
    \end{center}
    \vskip -0.05 in
    \caption{A visualization example of attentions initialization strategy for KW ($1\times$), where both $n$ and $m_{t}$ equal to 6.
    It helps the ConvNet to build one-to-one relationships between kernel cells and linear mixtures in the early training stage according to our setting of $\beta_{ij}$.
    $\mathbf{e}_{z}$ is a kernel cell that doesn't really exist and it keeps as a zero matrix constantly. In the beginning of the training process when temperature $\tau$ is 1, a ConvNet built with KW ($1\times$) can be roughly seen as a ConvNet with standard convolutions.
    }
    \label{fig:attention_initialization_1x}
    \vskip -0.1 in
\end{figure}

\begin{figure}[ht]
    \begin{minipage}[t]{0.49\linewidth}
        \begin{center}
            \centerline{\includegraphics[width=\textwidth]{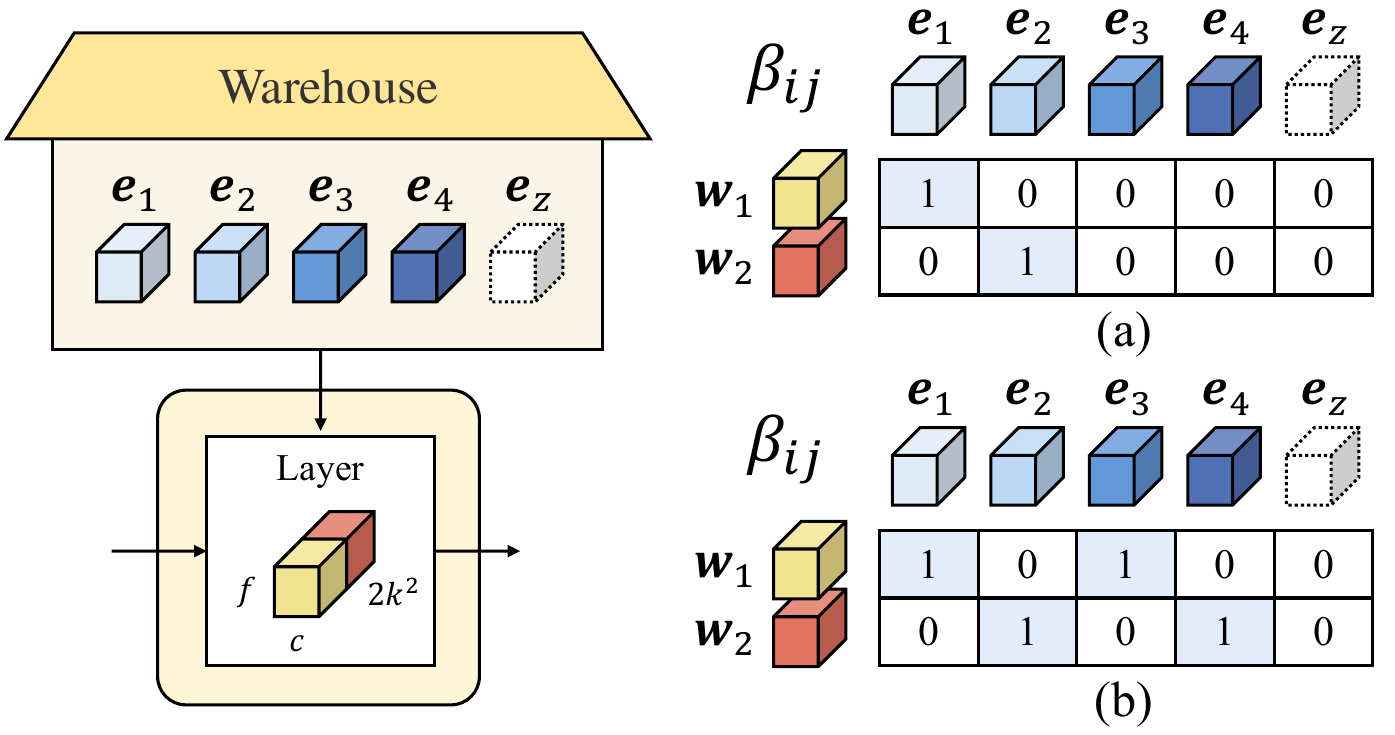}}
        \end{center}
        \vskip -0.1 in
        \caption{Visualization examples of attentions initialization strategies for KW ($2\times$), where $n=4$ and $m_{t}=2$. (a) our proposed strategy builds one-to-one relationships between kernel cells and linear mixtures; (b) an alternative strategy which builds two-to-one relationships between kernel cells and linear mixtures.}
        \label{fig:attention_initialization_2x}
    \end{minipage}
    \hfill
    \begin{minipage}[t]{0.49\linewidth}
        \begin{center}
            \centerline{\includegraphics[width=\textwidth]{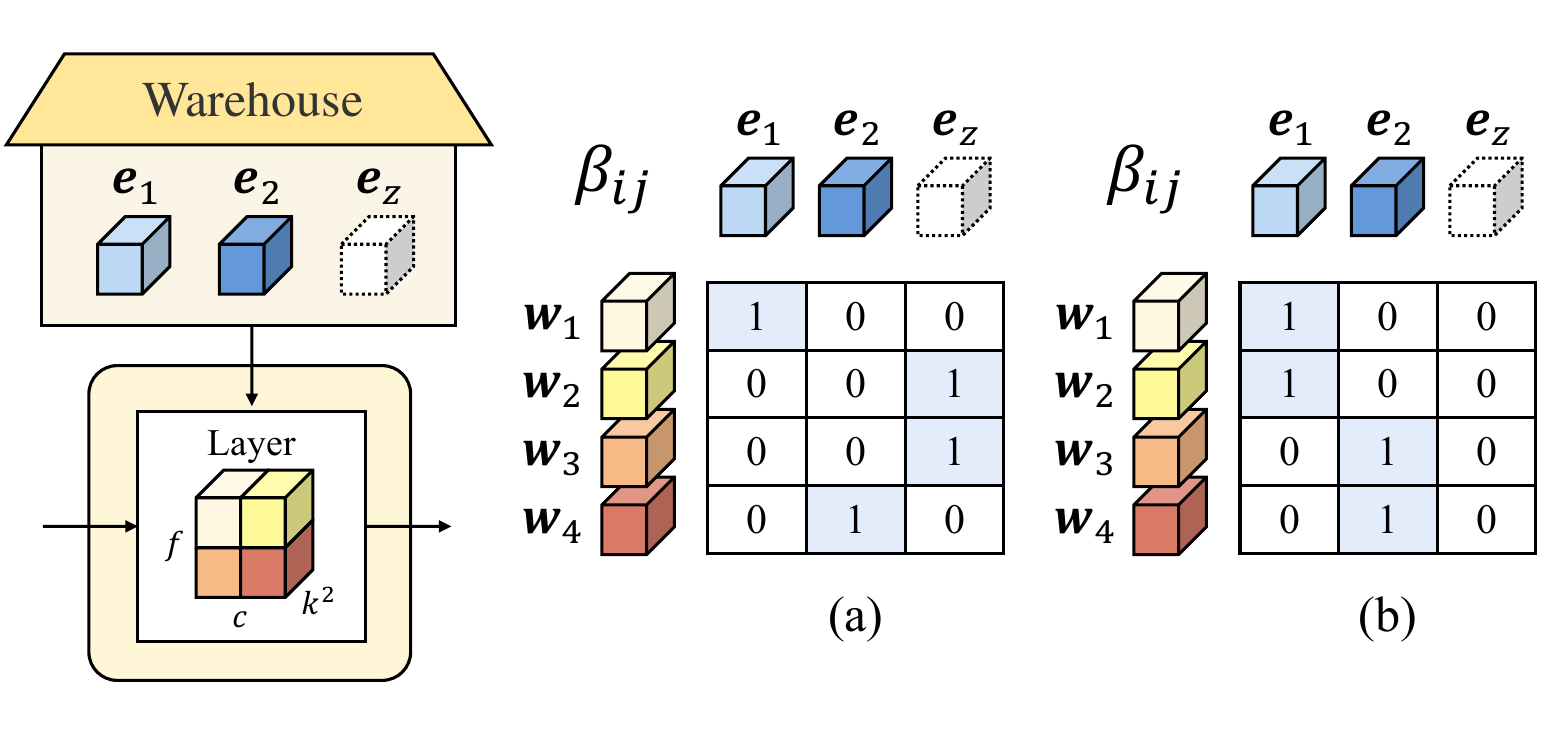}}
        \end{center}
        \vskip -0.1 in
        \caption{Visualization examples of attentions initialization strategies for KW ($1/2\times$), where $n=2$ and $m_{t}=4$. (a) our proposed strategy builds one-to-one relationships between kernel cells and linear mixtures; (b) an alternative strategy which builds one-to-two relationships between kernel cells and linear mixtures.}
        \label{fig:attention_initialization_1d2x}
    \end{minipage}
    \vskip -0.1in
\end{figure}

\begin{figure*}[ht!]
{
\setlength{\algomargin}{1.5em}
\vskip 0.05 in
\begin{center}
\scalebox{0.80}{
\begin{minipage}{0.8\linewidth}
\begin{algorithm2e}[H]
    \label{algorithm}
    \caption{Implementation of KernelWarehouse}
    \SetAlgoLined
    \SetKwInput{Part}{Part}
    \SetKwInOut{Require}{Require}
    \SetKwInOut{Return}{Return}
    \setcounter{AlgoLine}{0}
    {\centering \nonl \textbf{Part-A: Kernel Partition and Warehouse Construction-with-Sharing} \\}
    \Require{network $\mathbf{M}$ consisting of $S$ convolutional stages, parameter budget $b$}
    \For{$s\leftarrow 1$ \KwTo $S$}{
    $\{\mathbf{W}_{i} \in \mathbb{R}^{k_{i} \times k_{i} \times c_{i} \times f_{i}}\}^{l}_{i=1}$= static kernels in the $s$-th stage of $\mathbf{M}$ \\
    $k_{e}, c_{e}, f_{e} \leftarrow $\textbf{cdd}$(\{k_{i}\}^{l}_{i=1}), $\textbf{cdd}$(\{c_{i}\}^{l}_{i=1}), $\textbf{cdd}$(\{f_{i}\}^{l}_{i=1})$ \\
    $m_{t} \leftarrow 0$ \\
        \For{$i\leftarrow 1$ \KwTo $l$}{
        $\{\mathbf{w}_{j} \in \mathbb{R}^{k_{e}\times k_{e} \times c_{e} \times f_{e}}\}^{m}_{j=1} \leftarrow $\textbf{kernel\_partition}$(\mathbf{W}_{i}, k_{e}, c_{e}, f_{e})$\\
        $\mathbf{W}_{i} \leftarrow \mathbf{w}_{1} \cup \dots \cup \mathbf{w}_{m}$, and $\forall i,j \in \{1,\dots,m\}, i \neq j, w_{i} \cap w_{j}=\varnothing$ \\
        $m \leftarrow k_{i}k_{i}c_{i}f_{i} / k_{e}k_{e}c_{e}f_{e}$ \\
        $m_{t} \leftarrow m + m_{t}$\\
        }
        $n \leftarrow bm_{t}$ \\
        $\mathbf{E}_{s} \leftarrow \{\mathbf{e}_{i} \in \mathbb{R}^{k_{e} \times k_{e} \times c_{e} \times f_{e}} \}_{i=1}^{n}$ \\
     }
     $\mathbf{E} \leftarrow \{\mathbf{E}_{1},\dots ,\mathbf{E}_{S}\}$ \\
\Return{network $\mathbf{M}$ with partitioned kernels, set $\mathbf{E}$ consisting of $S$ warehouses}
\nonl \hrulefill \\
    {\centering \nonl \textbf{Part-B: Kernel Assembling for Single Same-Stage Convolutional Layer} \\}
    \setcounter{AlgoLine}{0}
    \Require{input $\mathbf{x}$, attention module $\phi$, warehouse $\mathbf{E}=\{\mathbf{e}_{i}\}_{i=1}^{n}$, linear mixtures $\{\mathbf{w}_{i}\}_{i=1}^{m}$}
    $\boldsymbol{\alpha} \leftarrow \phi(\mathbf{x})$ \\
    \For{$i\leftarrow 1$ \KwTo $m$}{
        $\mathbf{w_{i}} \leftarrow \alpha_{i1}\mathbf{e}_{1}+ \dots +\alpha_{in}\mathbf{e}_{n}$ \\
    }
    $\mathbf{W} \leftarrow \mathbf{w}_{1} \cup \dots \cup \mathbf{w}_{m}$, and $\forall i,j \in \{1,\dots,m\}, i \neq j, w_{i} \cap w_{j}=\varnothing$ \\
\Return{assembled kernel $\mathbf{W}$}
\end{algorithm2e}
\end{minipage}
}
\end{center}
}
\vskip -0.15 in
\end{figure*}
\begin{table}[ht]
\vskip 0.0 in
\caption{Ablation of KernelWarehouse with different attentions initialization strategies.}
\vskip -0.05 in
\label{table:ablation_initialization}
\begin{center}
\begin{small}
\resizebox{.75\linewidth}{!}{
\begin{tabular}{l|l|c|c|c}
\hline
Models & Attentions Initialization Strategies & Params & Top-1 Acc (\%) & Top-5 Acc (\%)\\
\hline
ResNet18 & - & 11.69M & 70.44 & 89.72 \\
\hline
\multirow{3}{*}{+ KW ($1\times$)} & 1 kernel cell to 1 linear mixture & 11.93M & \textbf{74.77} ($\uparrow$\textbf{4.33}) & \textbf{92.13} ($\uparrow$\textbf{2.41}) \\
 & all the kernel cells to 1 linear mixture & 11.93M & 73.36 ($\uparrow$2.92) & 91.41 ($\uparrow$1.69) \\
 & without attentions initialization & 11.93M & 73.39 ($\uparrow$2.95) & 91.24 ($\uparrow$1.52) \\
\hline
\multirow{2}{*}{+ KW ($4\times$)} &  1 kernel cell to 1 linear mixture & 45.86M & \textbf{76.05} ($\uparrow$\textbf{5.61}) & \textbf{92.68} ($\uparrow$\textbf{2.96}) \\
 &  4 kernel cells to 1 linear mixture & 45.86M & 76.03 ($\uparrow$5.59) & 92.53 ($\uparrow$2.81) \\
 \hline
 \multirow{2}{*}{+ KW ($1/2\times$)} &  1 kernel cell to 1 linear mixture & 7.43M & \textbf{73.33} ($\uparrow$\textbf{2.89}) & \textbf{91.42} ($\uparrow$\textbf{1.70}) \\
 &  1 kernel cell to 2 linear mixtures & 7.43M & 72.89 ($\uparrow$2.45) & 91.34 ($\uparrow$1.62) \\
 \hline
\end{tabular}
}
\end{small}
\end{center}
\vskip -0.1 in
\end{table}

Recall that we adopt an attentions initialization strategy for KernelWarehouse using $\tau$ and $\beta_{ij}$. It forces the scalar attentions to be one-hot in the early training stage for building one-to-one relationships between kernel cells and linear mixtures. To give a better understanding of this strategy, we provide visualization examples for KW ($1\times$), KW ($2\times$) and KW ($1/2\times$), respectively. We also provide a set of ablative experiments to compare our proposed strategy with other alternatives.

\textbf{Attentions Initialization for KW ($1\times$).}
A visualization example of attentions initialization strategy for KW ($1\times$) is shown in Figure~\ref{fig:attention_initialization_1x}.
In this example, a warehouse $\mathbf{E}=\{\mathbf{e}_{1},\dots,\mathbf{e}_{6},\mathbf{e}_{z}\}$ is shared to 3 neighboring convolutional layers with kernel dimensions of $k\times k \times 2c \times f$, $k\times 3k \times c \times f$ and $k\times k \times c \times f$, respectively. The kernel dimensions are selected for simple illustration. The kernel cells have the same dimensions of $k\times k \times c \times f$.
Note that the kernel cell $\mathbf{e}_{z}$ doesn't really exist and it keeps as a zero matrix constantly. It is only used for attentions normalization but not assembling kernels.
This kernel is mainly designed for attentions initialization when $b<1$ and not counted in the number of kernel cells $n$.
In the early training stage, we adopt a strategy to explicitly force every linear mixture to build relationship with one specified kernel cell according to our setting of $\beta_{ij}$.
As shown in Figure~\ref{fig:attention_initialization_1x}, we assign one of $\mathbf{e}_{1},\dots,\mathbf{e}_{6}$ in the warehouse to each of the 6 linear mixtures at the 3 convolutional layers without repetition.
So that in the beginning of the training process when temperature $\tau$ is 1, a ConvNet built with KW ($1\times$) can be roughly seen as a ConvNet with standard convolutions.
The results of Table~\ref{table:ablation_temperature} in the main manuscript validate the effectiveness of our proposed attentions initialization strategy. Here, we compare it with another alternative. In this alternative strategy, we force every linear mixture to build relationships with all the kernel cells equally by setting all the $\beta_{ij}$ to be 1. The results are shown in Table~\ref{table:ablation_initialization}.
The all-to-one strategy demonstrates similar performance with KernelWarehouse without using any attentions initialization strategy, while our proposed strategy outperforms it by 1.41\% top-1 gain.

\textbf{Attentions Initialization for KW ($2\times$).}
For KernelWarehouse with $b>1$, we adopt the same strategy for initializing attentions used in KW ($1\times$). Figure~\ref{fig:attention_initialization_2x}(a) provides a visualization example of attentions initialization strategy for KW ($2\times$). For building one-to-one relationships, we assign $\mathbf{e}_{1}$ to $\mathbf{w}_{1}$ and $\mathbf{e}_{2}$ to $\mathbf{w}_{2}$, respectively.
When $b>1$, another reasonable strategy is to assign multiple kernel cells to every linear mixture without repetition, which is shown in Figure~\ref{fig:attention_initialization_2x}(b). We use the ResNet18 backbone based on KW ($4\times$) to compare the two strategies. From the results in Table~\ref{table:ablation_initialization}, we can see that our one-to-one strategy performs better.

\textbf{Attentions Initialization for KW ($1/2\times$).}
For KernelWarehouse with $b<1$, the number of kernel cells is less than that of linear mixtures, meaning that we cannot adopt the same strategy used for $b\geq1$.
Therefore, we only assign one of the total $n$ kernel cells in the warehouse to $n$ linear mixtures respectively without repetition. And we assign $\mathbf{e}_{z}$ to all of the remaining linear mixtures. The visualization example for KW ($1/2\times$) is shown in Figure~\ref{fig:attention_initialization_1d2x}(a).
When temperature $\tau$ is 1, a ConvNet built with KW ($1/2\times$) can be roughly seen as a ConvNet with group convolutions (groups=2).
We also provide comparison results between our proposed strategy and another alternative strategy which assigns one of the $n$ kernel cells to every 2 linear mixtures without repetition.
As shown in Table~\ref{table:ablation_initialization}, our one-to-one strategy achieves better result again, showing that introducing an extra kernel $\mathbf{e}_{z}$ for $b<1$ can help the ConvNet learn more appropriate relationships between kernel cells and linear mixtures.
When assigning one kernel cell to multiple linear mixtures, a ConvNet could not balance the relationships between them well.

\subsection{Design Details of KernelWarehouse}

In this section, we describe the design details of our KernelWarehouse.
The corresponding values of $m$ and $n$ for each of our trained models are provided in the Table~\ref{table:m_and_n}.
Note that the values of $m$ and $n$ are naturally determined according to our setting of the dimensions of the kernel cells, the layers to share warehouses and $b$. Algorithm~\ref{algorithm} shows the implementation of KernelWarehouse, given a ConvNet backbone and the desired convolutional parameter budget $b$.

\begin{table}[ht]
\vskip 0.05 in
\caption{The values of $m$ and $n$ for the ResNet18, ResNet50, ConvNeXt-Tiny, MobileNetV2 (1.0$\times$) and MobileNetV2 (0.5$\times$) backbones based on KernelWarehouse.}
\vskip -0.05 in
\label{table:m_and_n}
\begin{center}
\begin{small}
\resizebox{0.98\linewidth}{!}{
\begin{tabular}{c|c|l|l}
\hline
Backbones & $b$ & m & n \\
\hline
\multirow{5}{*}{ResNet18} & 1/4 & 224, 188, 188, 108 & 56, 47, 47, 27 \\
& 1/2 & 224, 188, 188, 108 & 112, 94, 94, 54 \\
& 1 & 56, 47, 47, 27 & 56, 47, 47, 27 \\
& 2 & 56, 47, 47, 27 & 112, 94, 94, 54 \\
& 4 & 56, 47, 47, 27 & 224, 188, 188, 108 \\
\hline
\multirow{3}{*}{ResNet50} & 1/2 & 348, 416, 552, 188 & 174, 208, 276, 94 \\
& 1 & 87, 104, 138, 47 & 87, 104, 138, 47 \\
& 4 & 87, 104, 138, 47 & 348, 416, 552, 188 \\
\hline
\multirow{2}{*}{ConvNeXt-Tiny} & 1 & 16,4,4,4,147,24,147,24,147,24,147,24,147,24,147,24 & 16,4,4,4,147,24,147,24,147,24,147,24,147,24,147,24 \\
& 3/4 & 16,4,4,4,147,24,147,24,147,24,147,24,147,96,147,96 & 16,4,4,4,147,24,147,24,147,24,147,24,147,48,147,48 \\
\hline
\multirow{3}{*}{
\makecell[l]{MobileNetV2 (1.0$\times$) \\
MobileNetV2 (0.5$\times$)}
} & 1/2 & 9, 36, 18, 27, 36, 27, 12, 27, 80, 40 & 9, 36, 18, 27, 36, 27, 6, 27, 40, 20 \\
& 1 & 9, 36, 34, 78, 18, 42, 27, 102, 36, 120, 27, 58, 27 & 9, 36, 34, 78, 18, 42, 27, 102, 36, 120, 27, 58, 27 \\
& 4 & 9, 36, 11, 1, 2, 18, 7, 3, 27, 4, 4, 36, 9, 3, 27, 11, 3, 27, 20 &36, 144, 44, 4, 8, 72, 28, 12, 108, 16, 16, 144, 36, 12, 108, 44, 12, 108, 80 \\

\hline
\end{tabular}
}
\end{small}
\end{center}
\vskip -0.1 in
\end{table}

\begin{table}[ht]
\vskip 0.0 in
\caption{The example of warehouse sharing for the ResNet18 backbone based on KW ($1\times$) according to the original stages and reassigned stages.}
\vskip -0.05 in
\label{table:stage}
\begin{center}
\begin{small}
\resizebox{0.8\linewidth}{!}{
\begin{tabular}{c|c|c|c|c}
\hline
Dimensions of Kernel Cells & Original Stages & Layers & Reassigned Stages & Dimensions of Kernel Cells \\
\hline
\multirow{4}{*}{1$\times$1$\times$64$\times$64} & \multirow{4}{*}{1} & 3$\times$3$\times$64$\times$64 & \multirow{5}{*}{1} & \multirow{5}{*}{1$\times$1$\times$64$\times$64} \\
\cline{3-3}
& & 3$\times$3$\times$64$\times$64 & & \\
\cline{3-3}
& & 3$\times$3$\times$64$\times$64 & & \\
\cline{3-3}
& & 3$\times$3$\times$64$\times$64 & & \\
\cline{3-3}
\cline{1-2}
\multirow{4}{*}{1$\times$1$\times$64$\times$128} & \multirow{4}{*}{2} & 3$\times$3$\times$64$\times$128 &  & \\
\cline{3-3}
\cline{4-5}
& & 3$\times$3$\times$128$\times$128 &  \multirow{4}{*}{2} & \multirow{4}{*}{1$\times$1$\times$128$\times$128}\\
\cline{3-3}
& & 3$\times$3$\times$128$\times$128 & & \\
\cline{3-3}
& & 3$\times$3$\times$128$\times$128 & & \\
\cline{1-2}
\cline{3-3}
\multirow{4}{*}{1$\times$1$\times$128$\times$256} & \multirow{4}{*}{3} & 3$\times$3$\times$128$\times$256 & & \\
\cline{4-5}
\cline{3-3}
& & 3$\times$3$\times$256$\times$256 & \multirow{4}{*}{3} & \multirow{4}{*}{1$\times$1$\times$256$\times$256} \\
\cline{3-3}
& & 3$\times$3$\times$256$\times$256 & & \\
\cline{3-3}
& & 3$\times$3$\times$256$\times$256 & &  \\
\cline{1-2}
\cline{3-3}
\multirow{4}{*}{1$\times$1$\times$256$\times$512} & \multirow{4}{*}{4} & 3$\times$3$\times$256$\times$512 & & \\
\cline{4-5}
\cline{3-3}
& & 3$\times$3$\times$512$\times$512 & \multirow{3}{*}{4} & \multirow{3}{*}{1$\times$1$\times$512$\times$512}\\
\cline{3-3}
& & 3$\times$3$\times$512$\times$512 & &  \\
\cline{3-3}
& & 3$\times$3$\times$512$\times$512 & & \\
\hline
\end{tabular}
}
\end{small}
\end{center}
\vskip -0.1 in
\end{table}

\textbf{Design details of Attention Module of KernelWarehouse.} Following existing dynamic convolution methods
, KernelWarehouse also adopts a compact SE-typed structure as the attention module $\phi(x)$ (illustrated in Figure~\ref{fig:architecture}) to generate attentions for weighting kernel cells in a warehouse. For any convolutional layer with a static kernel $\mathbf{W}$, it starts with a channel-wise global average pooling (GAP) operation that maps the input $\mathbf{x}$ into a feature vector, followed by a fully connected (FC) layer, a rectified linear unit (ReLU), another FC layer, and a contrasting-driven attention function (CAF). The first FC layer reduces the length of the feature vector by 16, and the second FC layer generates $m$ sets of $n$ feature logits in parallel which are finally normalized by our CAF set by set.

\textbf{Design details of KernelWarehouse on ResNet18.}
Recall that in KernelWarehouse, a warehouse is shared to all same-stage convolutional layers.
While the layers are originally divided into different stages according to the resolutions of their input feature maps, the layers are divided into different stages according to their kernel dimensions in our KernelWarehouse. In our implementation, we usually reassign the first layer (or the first two layers) in each stage to the previous stage. An example for the ResNet18 backbone based on KW ($1\times$) is given in Table~\ref{table:stage}.
By reassigning the layers, we can avoid the condition that all the other layers have to be partitioned according to a single layer because of the greatest common dimension divisors.
For the ResNet18 backbone, we apply KernelWarehouse to all the convolutional layers except the first one. In each stage, the corresponding warehouse is shared to all of its convolutional layers. For KW ($1\times$), KW ($2\times$) and KW ($4\times$), we use the greatest common dimension divisors for static kernels as the uniform kernel cell dimensions for kernel partition. For KW ($1/2\times$) and KW ($1/4\times$), we use half of the greatest common dimension divisors.

\textbf{Design details of KernelWarehouse on ResNet50.}
For the ResNet50 backbone, we apply KernelWarehouse to all the convolutional layers except the first two layers. In each stage, the corresponding warehouse is shared to all of its convolutional layers. For KW ($1\times$) and KW ($4\times$), we use the greatest common dimension divisors for static kernels as the uniform kernel cell dimensions for kernel partition. For KW ($1/2\times$), we use half of the greatest common dimension divisors.

\begin{table}[ht]
\vskip 0.05 in
\caption{Comparison of memory requirements of DY-Conv, ODConv and KernelWarehouse for training and inference. For ResNet50, we set batch size to 128$|$100 for each gpu during training$|$inference; for MobileNetV2($1.0\times$), we set batch size to 32$|$100 for each gpu during training$|$inference.}
\label{table:memory}
\vskip -0.05 in
\begin{center}
\begin{small}
\begin{minipage}[t]{0.48\linewidth}
\resizebox{1.0\linewidth}{!}{
\begin{tabular}{l|c|c|c}
\hline
\multirow{2}*{\makecell[l]{Models}} & \multirow{2}*{\makecell[l]{Params}} & Training Memory & Inference Memory \\
& & (batch size=128) & (batch size=100) \\
\hline
ResNet50 & 25.56M & 11,084 MB & 1,249 MB \\
+ DY-Conv ($4\times$) & 100.88M & 24,552 MB & 2,062 MB \\
+ ODConv ($4\times$) & 90.67M & 31,892 MB & 5,405 MB \\
+ KW ($1/2\times$) & 17.64M & 23,323 MB & 2,121 MB \\
+ KW ($1\times$) & 28.05M & 23,231 MB & 2,200 MB \\
+ KW ($4\times$) & 102.02M & 24,905 MB & 2,762 MB \\
\hline
\end{tabular}
}
\end{minipage}
\hfill
\begin{minipage}[t]{0.50\linewidth}
\resizebox{1.0\linewidth}{!}{
\begin{tabular}{l|c|c|c}
\hline
\multirow{2}*{\makecell[l]{Models}} & \multirow{2}*{\makecell[l]{Params}} & Training Memory & Inference Memory \\
& & (batch size=32) & (batch size=100) \\
\hline
MobileNetV2 ($1.0\times$) & 3.50M & 2,486 MB & 1,083 MB \\
+ DY-Conv ($4\times$) & 12.40M & 2,924 MB & 1,151 MB \\
+ ODConv ($4\times$) & 11.52M & 4,212 MB & 1,323 MB \\
+ KW ($1/2\times$) & 2.65M & 3,002 MB & 1,076 MB \\
+ KW ($1\times$) & 5.17M & 2,823 MB & 1,096 MB \\
+ KW ($4\times$) & 11.38M & 2,916 MB & 1,144 MB \\
\hline
\end{tabular}
}
\end{minipage}
\end{small}
\end{center}
\vskip -0.1 in
\end{table}

\textbf{Design details of KernelWarehouse on ConvNeXt-Tiny.}
For the ConvNeXt backbone, we apply KernelWarehouse to all the convolutional layers. We partition the 9 blocks in the third stage of the ConvNeXt-Tiny backbone into three stages with the equal number of blocks. In each stage, the corresponding three warehouses are shared to the point-wise convolutional layers, the depth-wise convolutional layers and the downsampling layer, respectively. For KW (1$\times$), we use the greatest common dimension divisors for static kernels as the uniform kernel cell dimensions for kernel partition.  For KW ($3/4\times$), we apply KW ($1/2\times$) to the point-wise convolutional layers in the last two stages of ConvNeXt backbone using half of the greatest common dimension divisors. And we apply KW ($1\times$) to the other layers using the greatest common dimension divisors.

\textbf{Design details of KernelWarehouse on MobileNetV2.}
For the MobileNetV2 (1.0$\times$) and MobileNetV2 (0.5$\times$) backbones based on KW ($1\times$) and KW ($4\times$), we apply KernelWarehouse to all the convolutional layers. For MobileNetV2 (1.0$\times$, 0.5$\times$) based on KW ($1\times$), the corresponding two warehouses are shared to the point-wise convolutional layers and the depth-wise convolutional layers in each stage, respectively. For MobileNetV2 (1.0$\times$, 0.5$\times$) based on KW ($4\times$), the corresponding three warehouses are shared to the depth-wise convolutional layers, the point-wise convolutional layers for channel expansion and the point-wise convolutional layers for channel reduction in each stage, respectively. We use the greatest common dimension divisors for static kernels as the uniform kernel cell dimensions for kernel partition.
For the MobileNetV2 (1.0$\times$) and MobileNetV2 (0.5$\times$) backbones based on KW ($1/2\times$), we take the parameters in the attention modules and classifier layer into consideration in order to reduce the total number of parameters. We apply KernelWarehouse to all the depth-wise convolutional layers, the point-wise convolutional layers in the last two stages and the classifier layer. We set $b=1$ for the point-wise convolutional layers and $b=1/2$ for the other layers. For the depth-wise convolutional layers, we use the greatest common dimension divisors for static kernels as the uniform kernel cell dimensions for kernel partition. For the point-wise convolutional layers, we use half of the greatest common dimension divisors. For the classifier layer, we use the kernel cell dimensions of 1000$\times$32.

\subsection{More Experiments for Studying Other Potentials of KernelWarehouse}

In this section, we provide a lot of extra experiments conducted for studying other potentials of KernelWarehouse.

\textbf{Comparison of Memory Requirements.}
From the table~\ref{table:memory}, we can observe that, for both training and inference, the memory requirements of our method are very similar to those of DY-Conv, and are much smaller than those for ODConv (that generates attention weights along all four dimensions including the input channel number, the output channel number, the spatial kernel size and the kernel number, rather than one single dimension as DY-Conv and KernelWarehouse), showing that our method does not have a potential limitation on memory requirements compared to existing top-performing dynamic convolution methods. The reason is: although KernelWarehouse introduces dense attentive mixturing and assembling operations at the same-stage convolutional layers having a shared warehouse, the memory requirement for these operations is significantly smaller than that for convolutional feature maps and the memory requirement for attention weights are also significantly smaller than that for convolutional weights, under the same convolutional parameter budget $b$.

\begin{table}[ht]
\vskip 0.0 in
\caption{Effect of combining KernelWarehouse and ODConv, where KW* denotes attention function which combines KernelWarehouse and ODConv.}
\label{table:odconv_kernelwarehouse}
\vskip -0.05 in
\begin{center}
\begin{small}
\resizebox{0.5\linewidth}{!}{
\begin{tabular}{l|c|c|c}
\hline
Models & Params & Top-1 Acc (\%) & Top-5 Acc (\%) \\
\hline
MobileNetV2 ($1.0\times$) & 3.50M & 72.02 & 90.43 \\
\hline
+ ODConv ($4\times$) & 11.52M & 75.42 ($\uparrow$3.40) & 92.18 ($\uparrow$1.75) \\
+ KW ($4\times$) & 11.38M & 75.92 ($\uparrow$3.90) & 92.22 ($\uparrow$1.79) \\
+ KW*($4\times$) & 12.51M & \textbf{76.54} ($\uparrow$\textbf{4.52}) & \textbf{92.35} ($\uparrow$\textbf{1.92}) \\
\hline
\end{tabular}
}
\end{small}
\end{center}
\vskip -0.1 in
\end{table}

\textbf{Combining KernelWarehouse with ODConv.}
The improvement of KernelWarehouse to ODConv could be further boosted by a simple combination of  KernelWarehouse and ODConv to compute attention weights for KernelWarehouse along the aforementioned four dimensions instead of one single dimension. We add experiments to explore this potential, and the results are summarized in the Table~\ref{table:odconv_kernelwarehouse}. We can see that, on the ImageNet dataset with MobileNetV2 ($1.0\times$) backbone, combining ODConv with KernelWarehouse ($4\times$) further brings 1.12\% absolute top-1 improvement to ODConv ($4\times$) while retaining the similar model size.

\subsection{More Visualization Results for Learnt Attentions of KernelWarehouse}
In the main manuscript, we provide visualization results of learnt attention values $\alpha_{ij}$ for the ResNet18 backbone based on KW ($1\times$) (see Figure~\ref{figure:attention_resnet18_heatmap} in the main manuscript). For a better understanding of KernelWarehouse, we provide more visualization results in this section, covering different alternative attention functions, alternative initialization strategies and values of $b$.
For all the results, the statistical mean values of learnt attention $\alpha_{ij}$ are obtained using all of the 50,000 images on the ImageNet validation dataset.

\begin{figure}[ht]
\vskip 0.05 in
    \begin{minipage}[t]{1.0\linewidth}
        \begin{minipage}[t]{0.24\linewidth}
            \begin{center}
                \centerline{\includegraphics[width=\textwidth]{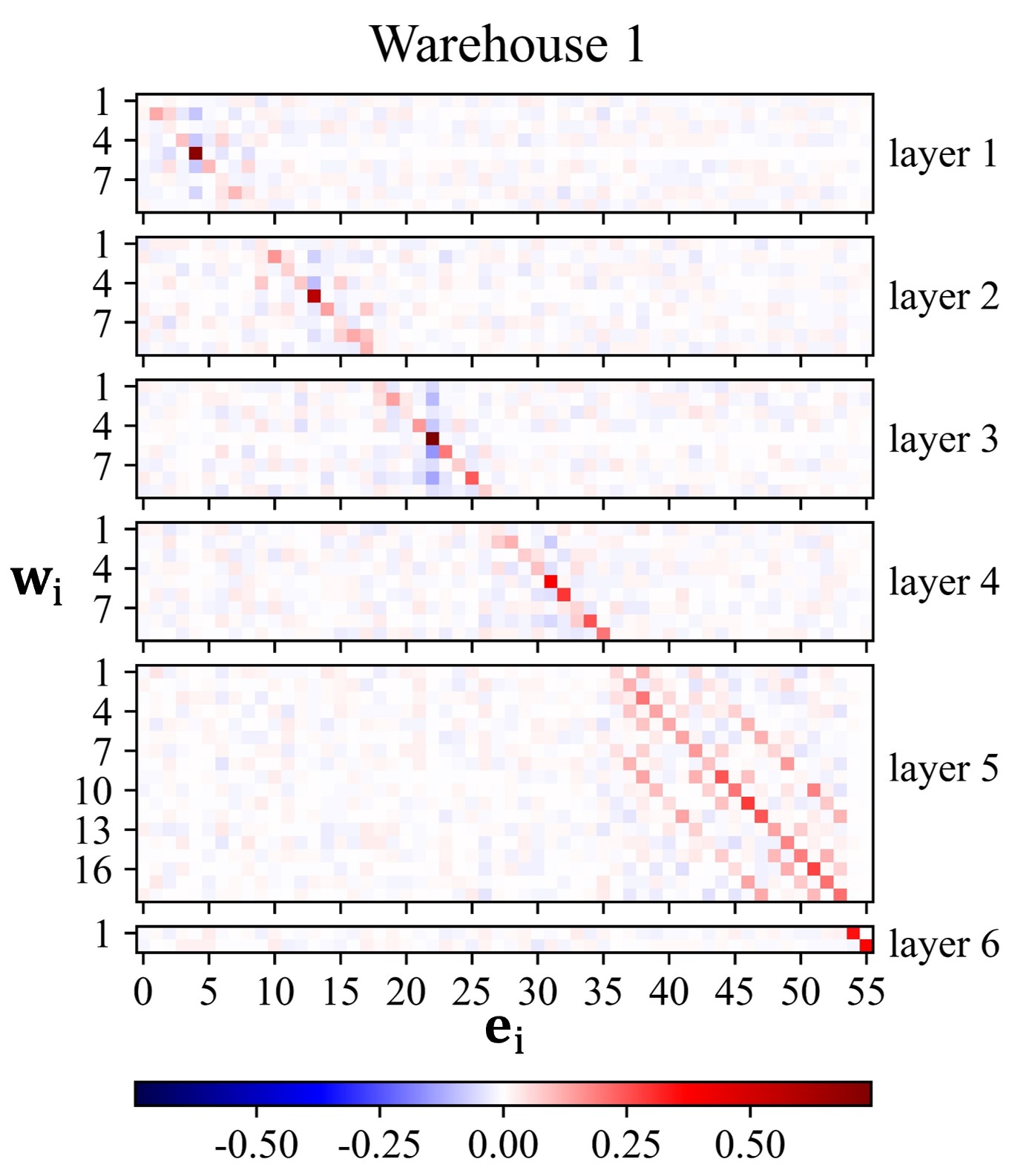}}
            \end{center}
        \end{minipage}
        \hfill
        \begin{minipage}[t]{0.24\linewidth}
            \begin{center}
                \centerline{\includegraphics[width=\textwidth]{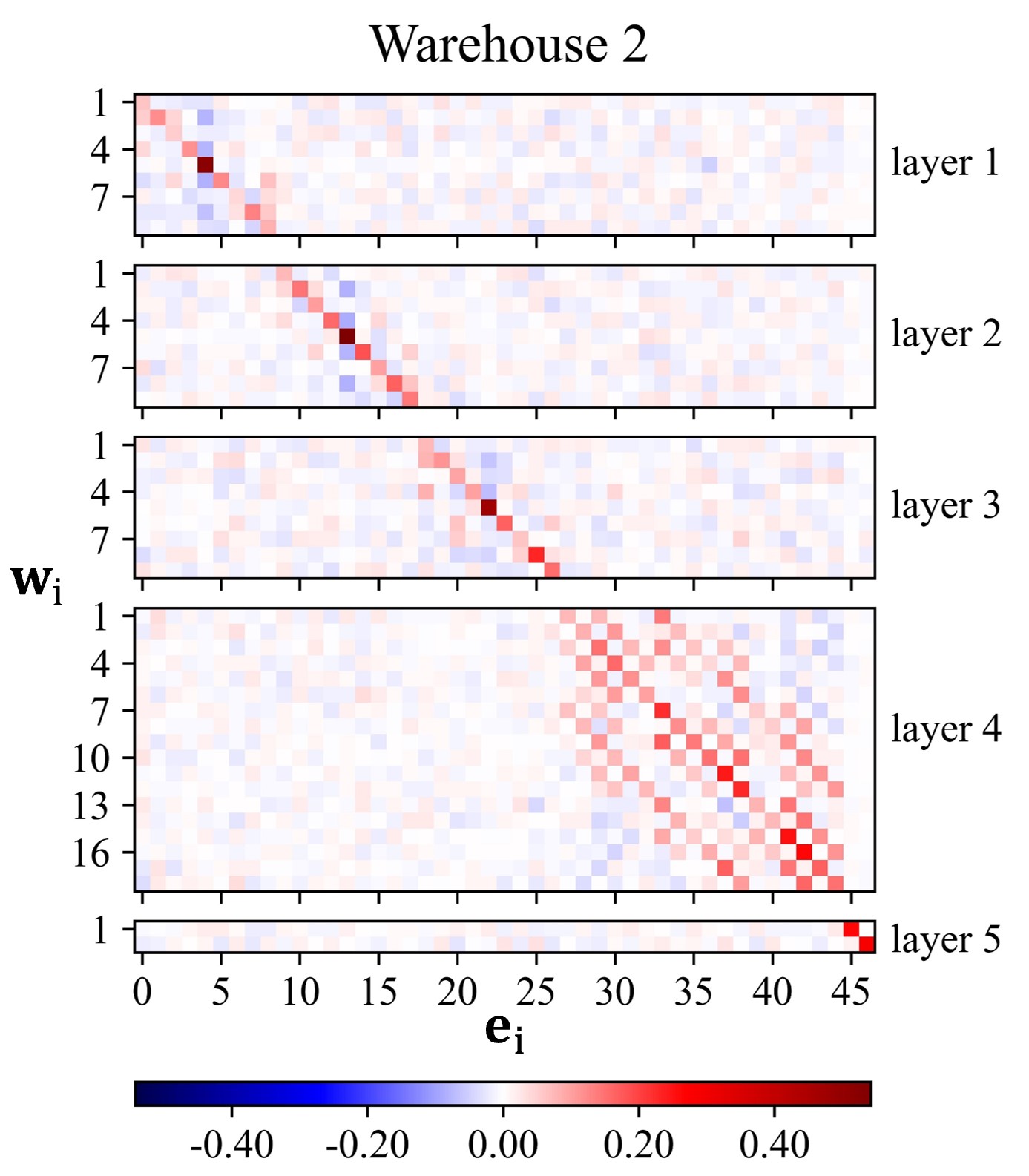}}
            \end{center}
        \end{minipage}
        \hfill
        \begin{minipage}[t]{0.24\linewidth}
            \begin{center}
                \centerline{\includegraphics[width=\textwidth]{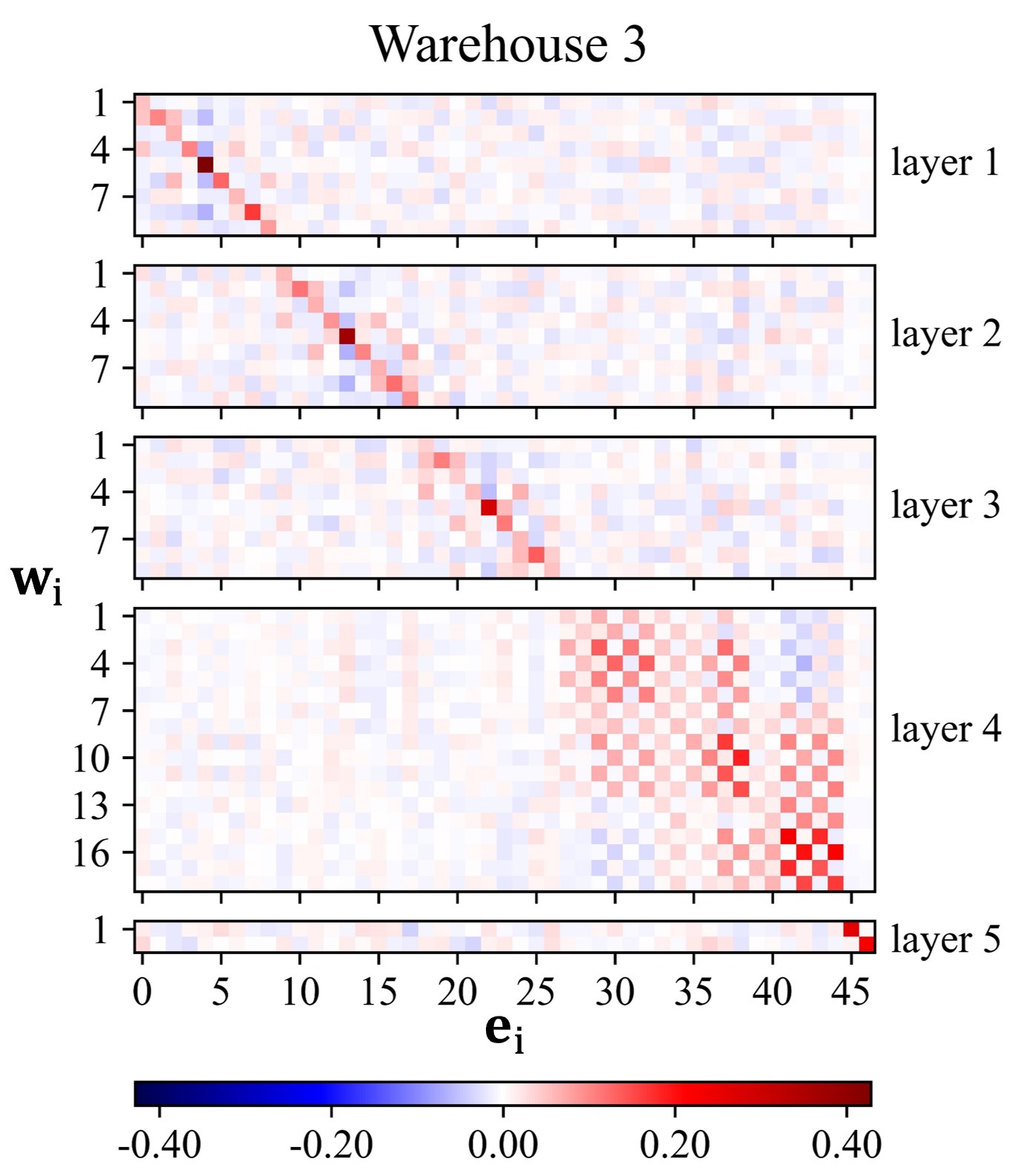}}
            \end{center}
        \end{minipage}
            \hfill
        \begin{minipage}[t]{0.24\linewidth}
            \begin{center}
                \centerline{\includegraphics[width=\textwidth]{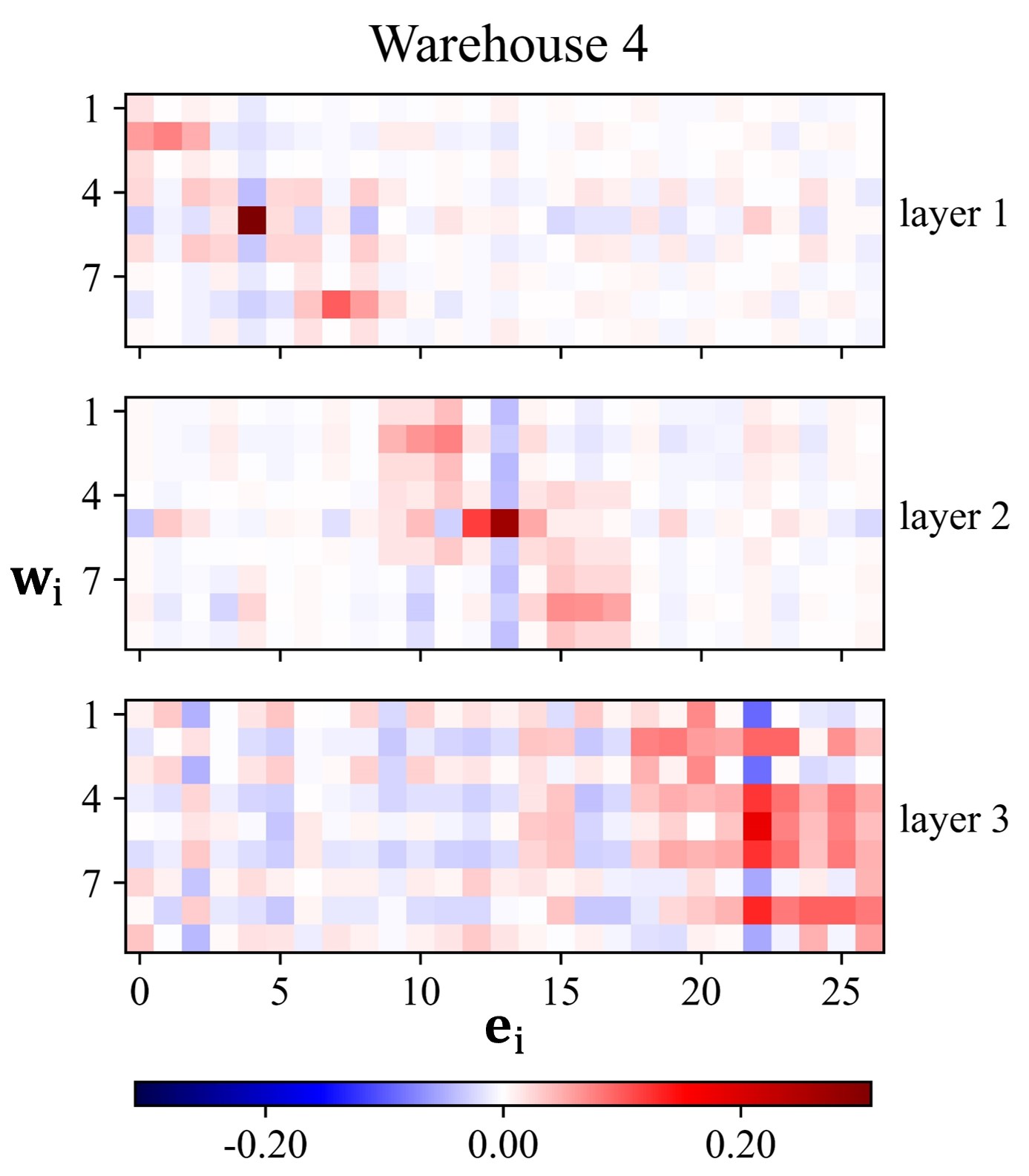}}
            \end{center}
        \end{minipage}
    \vskip -0.2 in
    \subcaption{}
    \end{minipage}
    \begin{minipage}[t]{1.0\linewidth}
        \begin{minipage}[t]{0.24\linewidth}
            \begin{center}
                \centerline{\includegraphics[width=\textwidth]{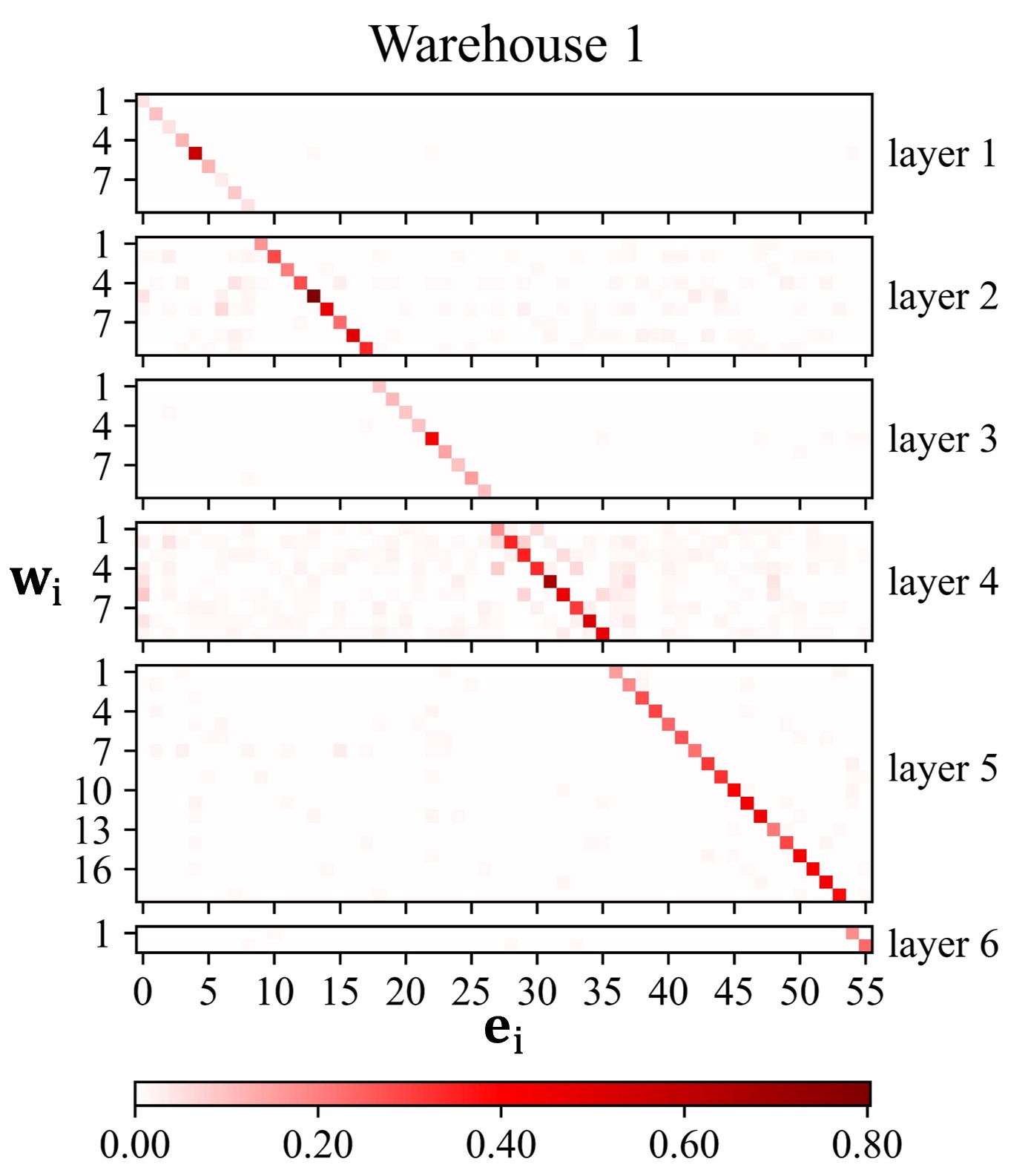}}
            \end{center}
        \end{minipage}
        \hfill
        \begin{minipage}[t]{0.24\linewidth}
            \begin{center}
                \centerline{\includegraphics[width=\textwidth]{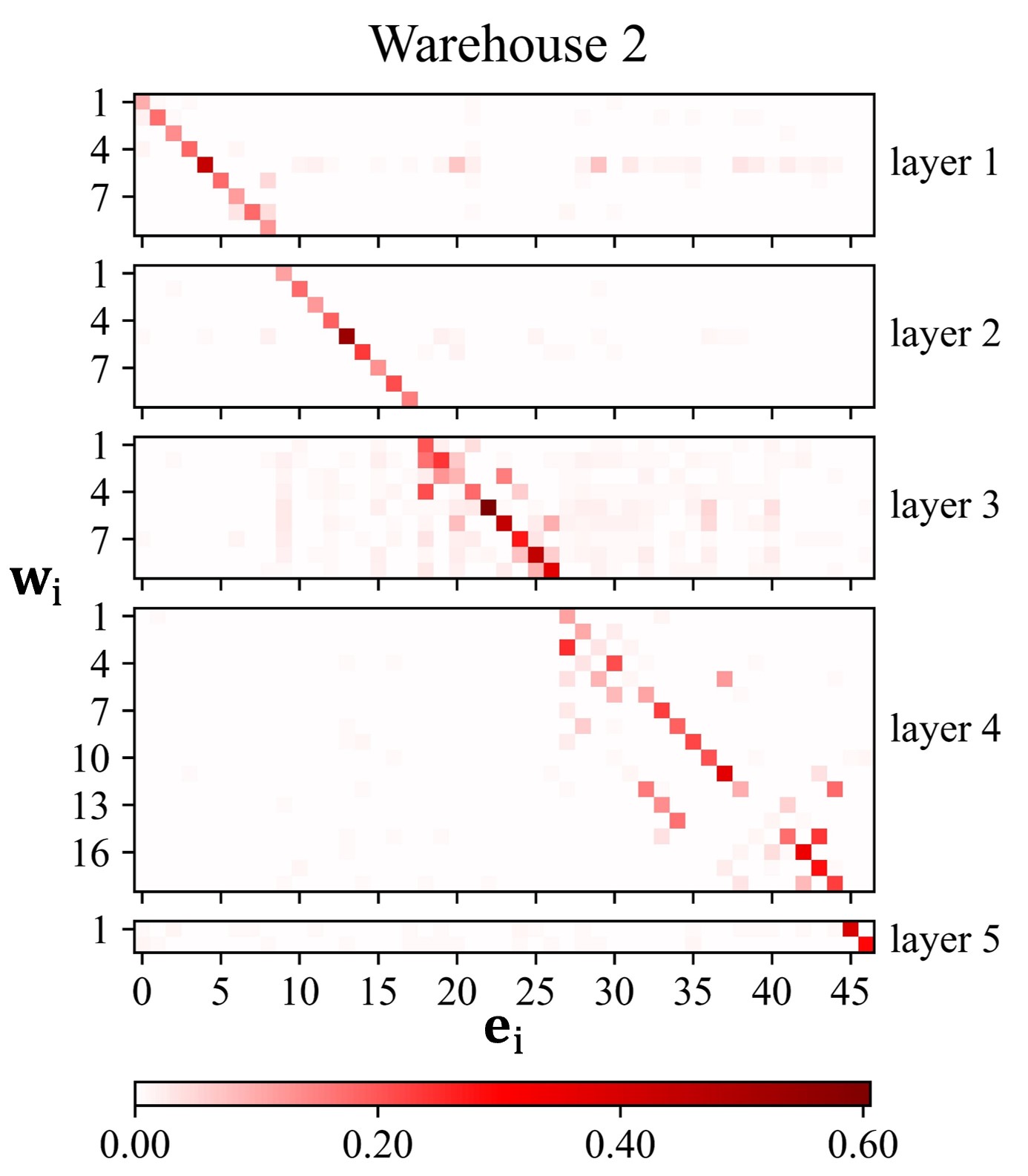}}
            \end{center}
        \end{minipage}
        \hfill
        \begin{minipage}[t]{0.24\linewidth}
            \begin{center}
                \centerline{\includegraphics[width=\textwidth]{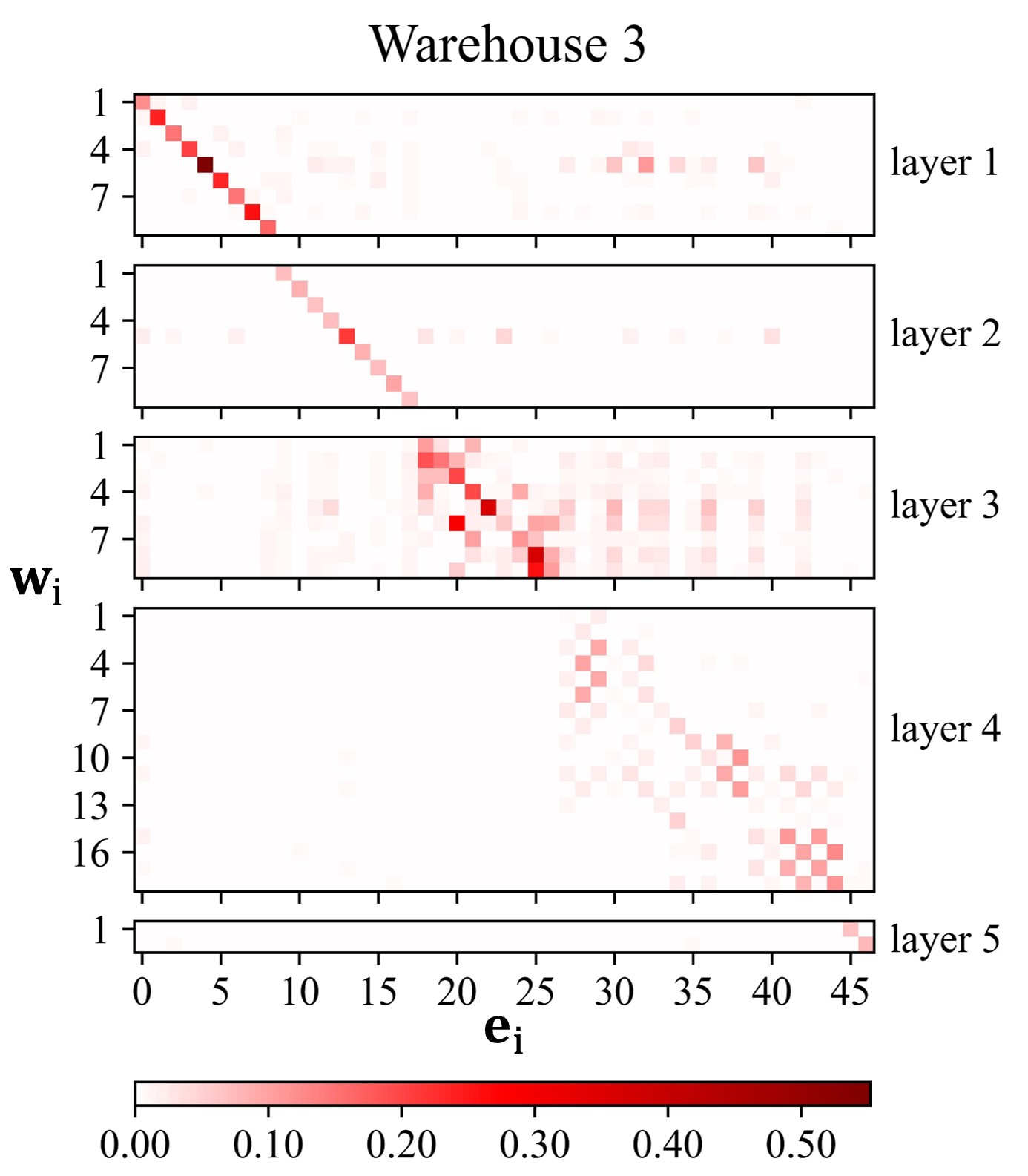}}
            \end{center}
        \end{minipage}
            \hfill
        \begin{minipage}[t]{0.24\linewidth}
            \begin{center}
                \centerline{\includegraphics[width=\textwidth]{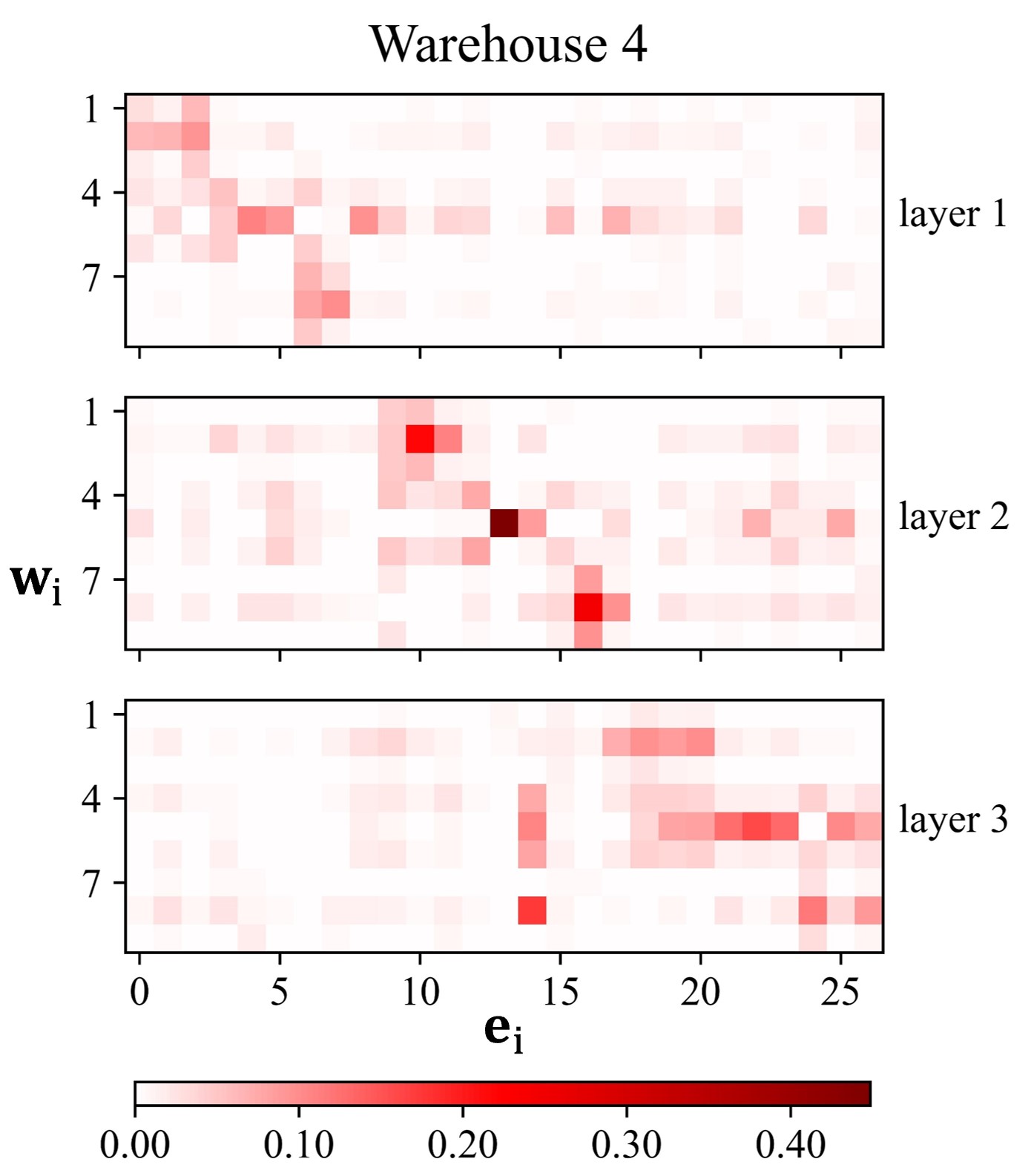}}
            \end{center}
        \end{minipage}
    \vskip -0.2 in
    \subcaption{}
    \end{minipage}
    \begin{minipage}[t]{1.0\linewidth}
        \begin{minipage}[t]{0.24\linewidth}
            \begin{center}
                \centerline{\includegraphics[width=\textwidth]{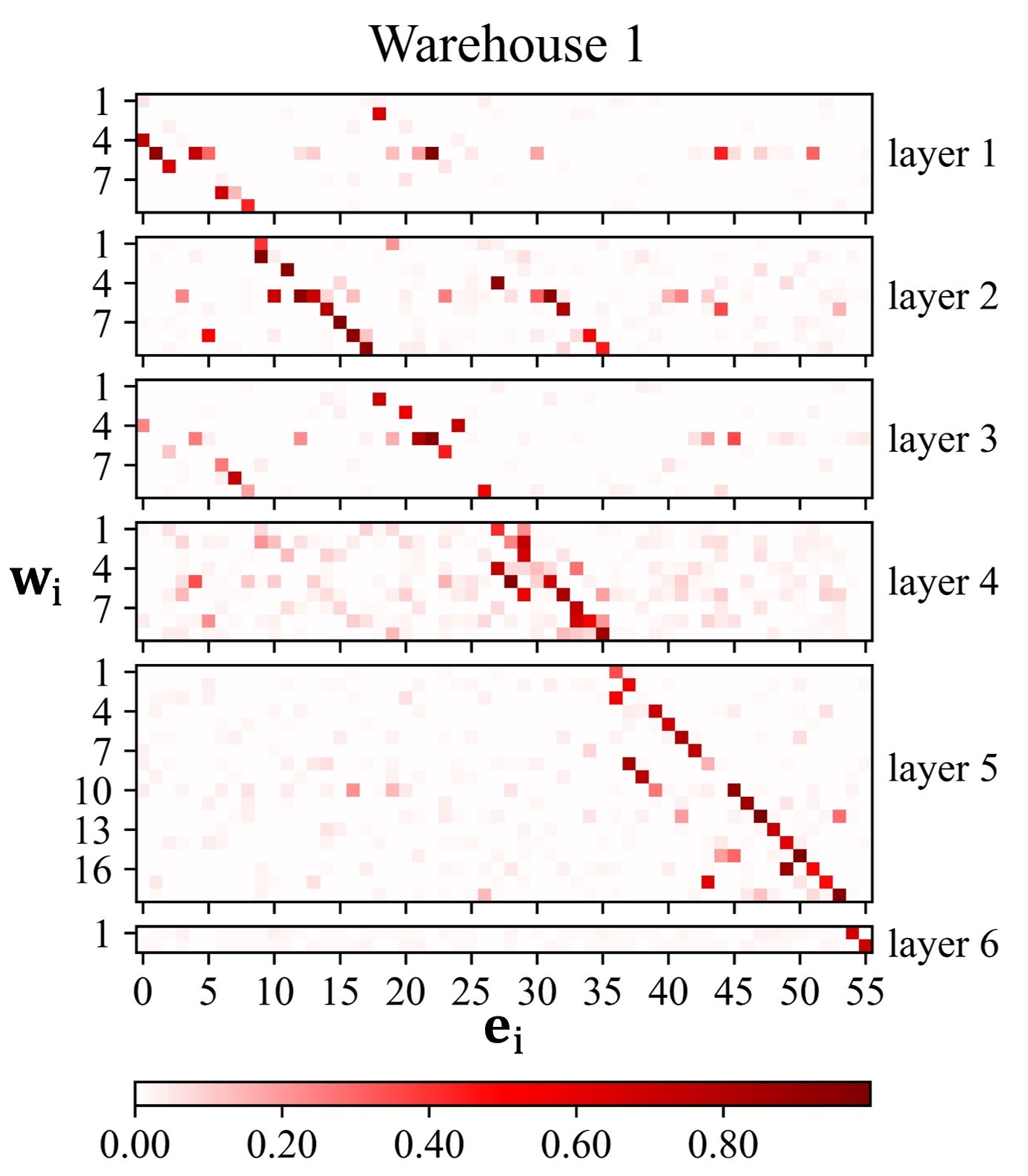}}
            \end{center}
        \end{minipage}
        \hfill
        \begin{minipage}[t]{0.24\linewidth}
            \begin{center}
                \centerline{\includegraphics[width=\textwidth]{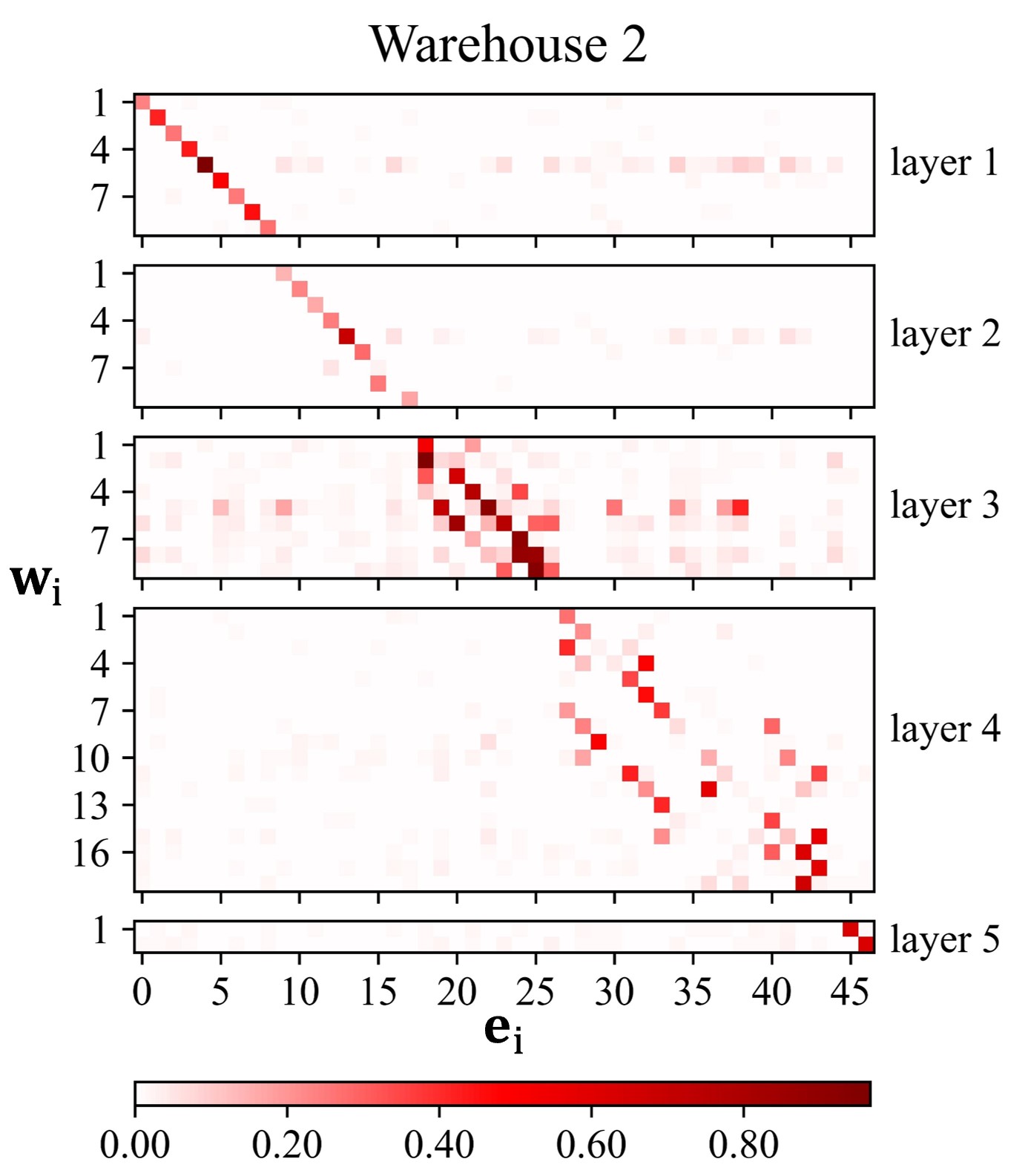}}
            \end{center}
        \end{minipage}
        \hfill
        \begin{minipage}[t]{0.24\linewidth}
            \begin{center}
                \centerline{\includegraphics[width=\textwidth]{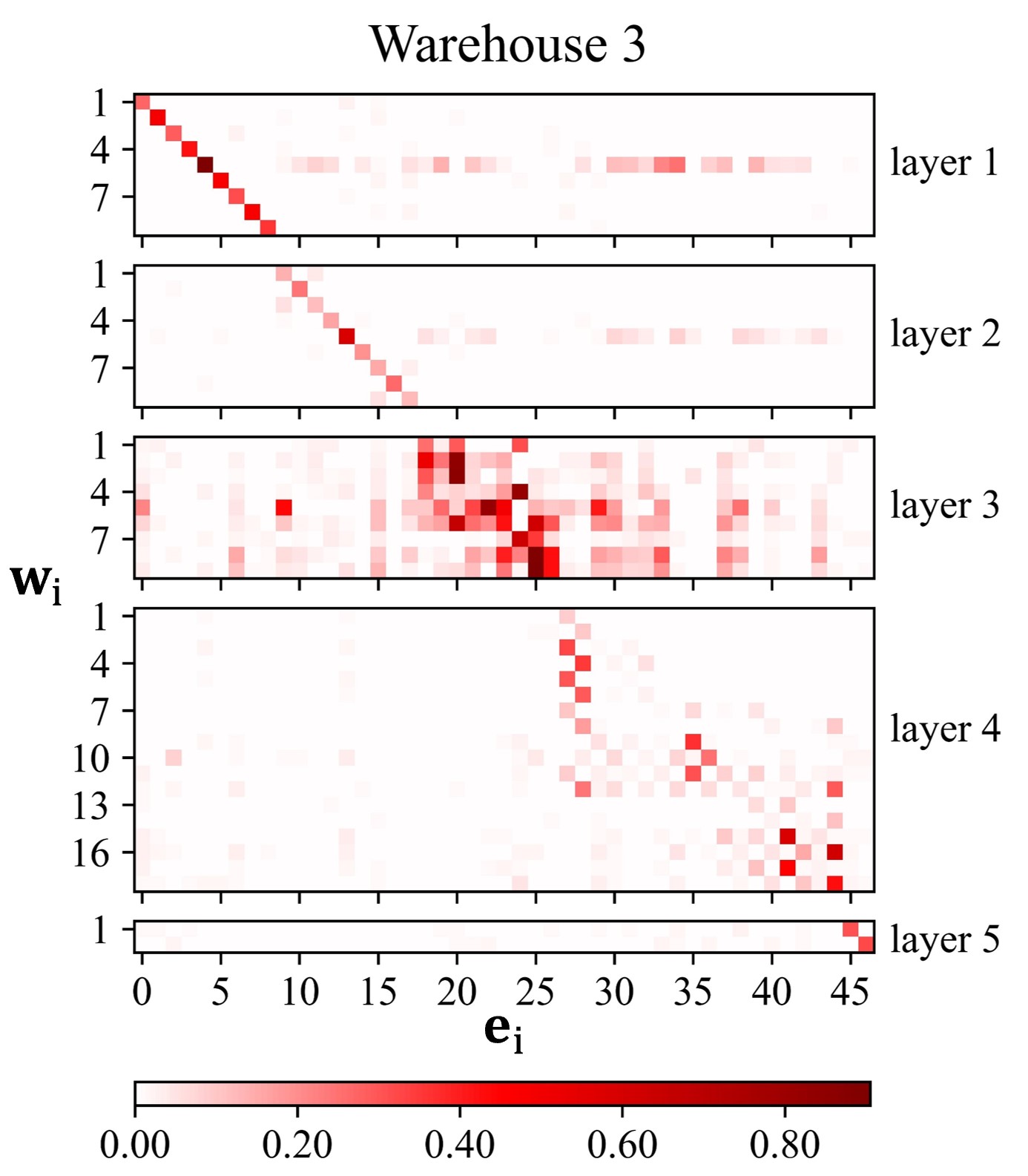}}
            \end{center}
        \end{minipage}
            \hfill
        \begin{minipage}[t]{0.24\linewidth}
            \begin{center}
                \centerline{\includegraphics[width=\textwidth]{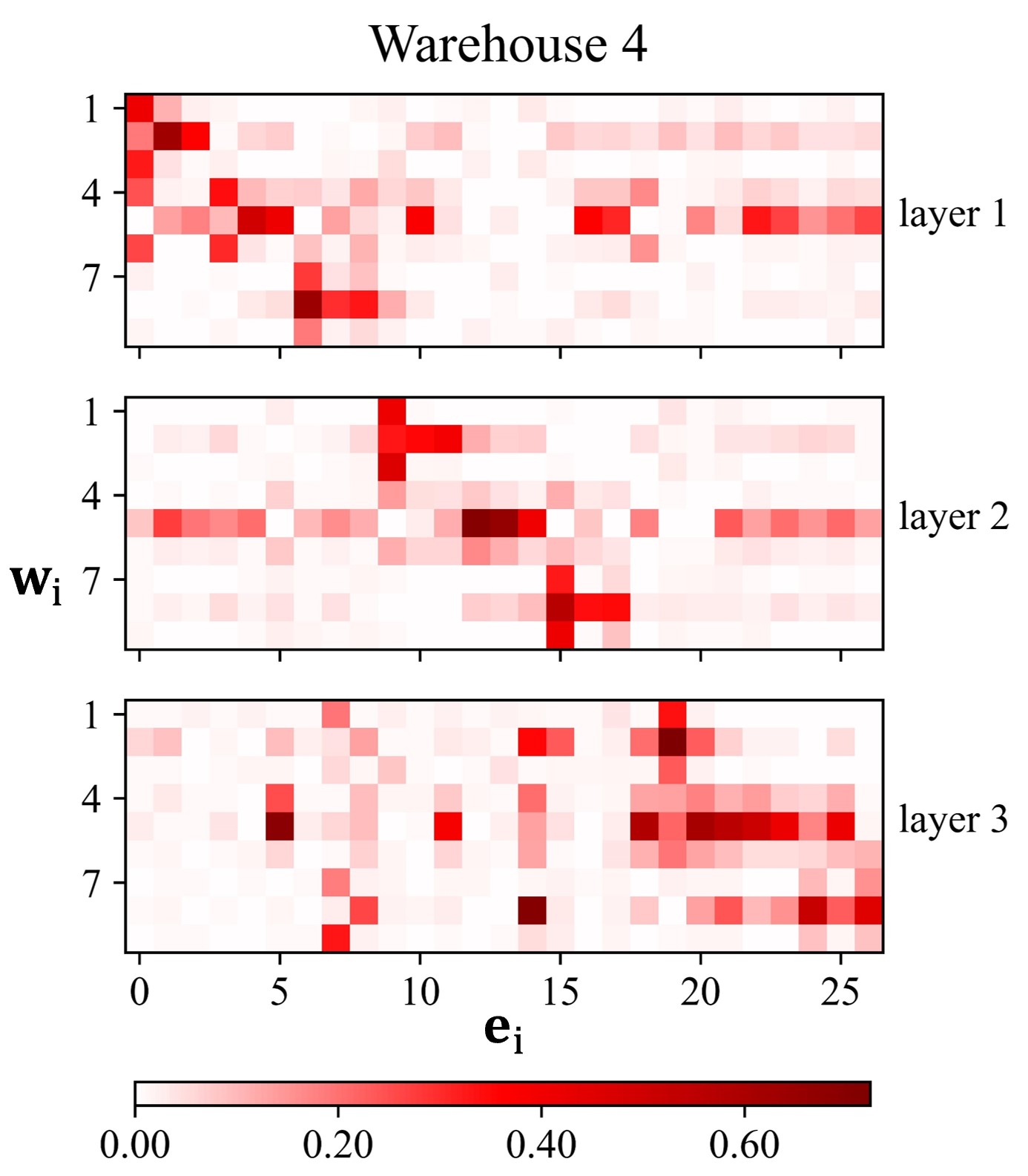}}
            \end{center}
        \end{minipage}
    \vskip -0.2 in
    \subcaption{}
    \end{minipage}
    \begin{minipage}[t]{1.0\linewidth}
        \begin{minipage}[t]{0.24\linewidth}
            \begin{center}
                \centerline{\includegraphics[width=\textwidth]{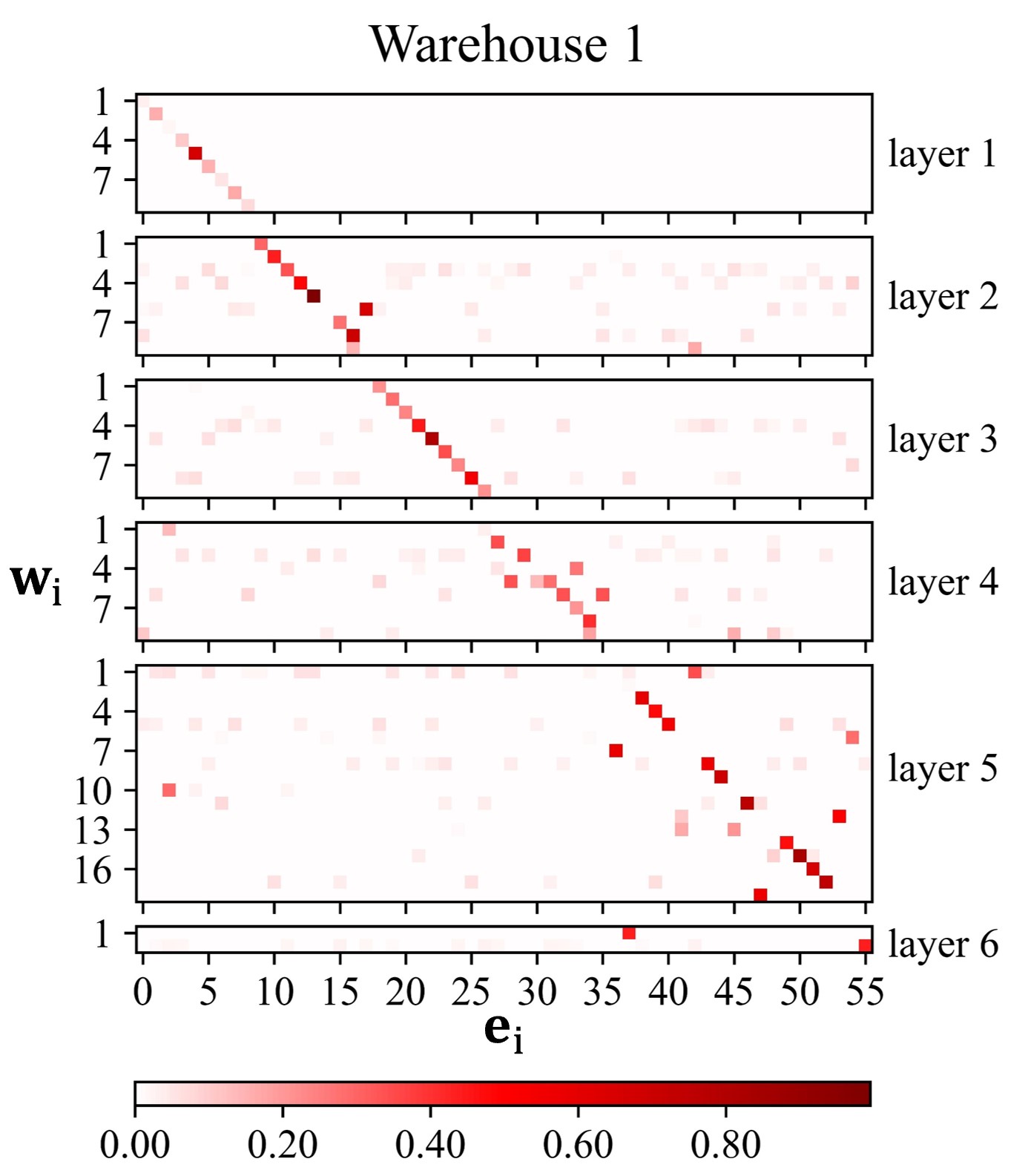}}
            \end{center}
        \end{minipage}
        \hfill
        \begin{minipage}[t]{0.24\linewidth}
            \begin{center}
                \centerline{\includegraphics[width=\textwidth]{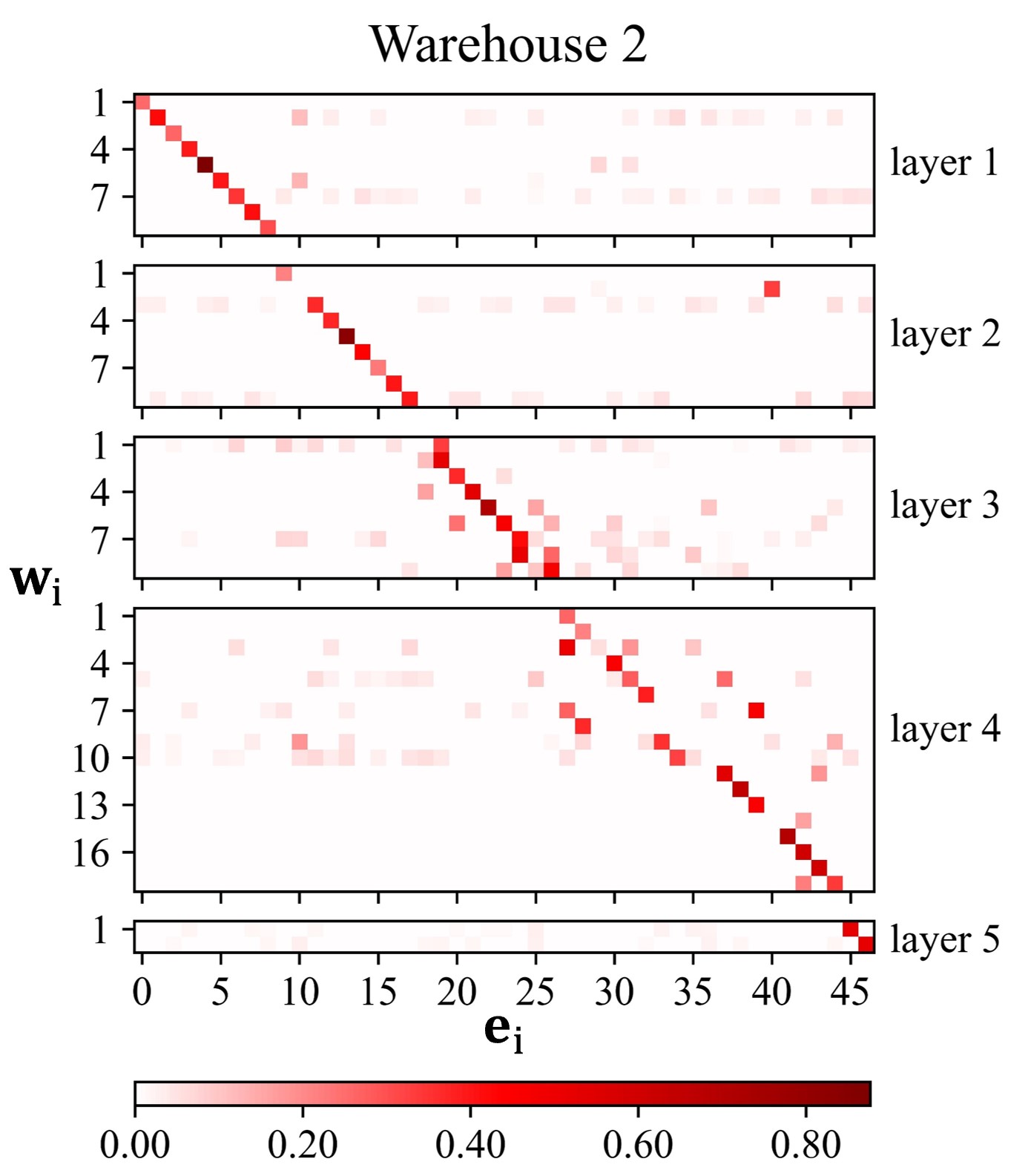}}
            \end{center}
        \end{minipage}
        \hfill
        \begin{minipage}[t]{0.24\linewidth}
            \begin{center}
                \centerline{\includegraphics[width=\textwidth]{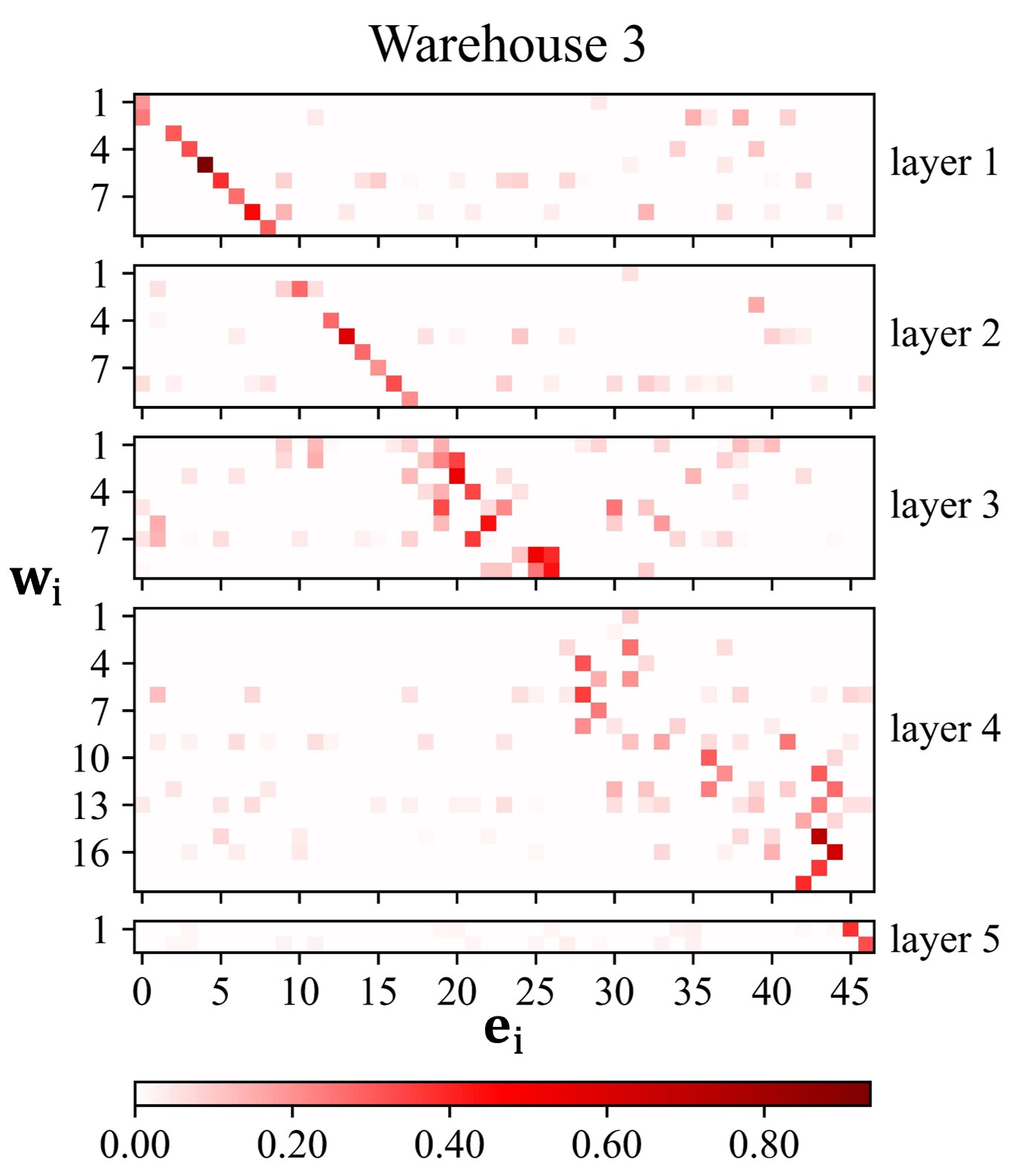}}
            \end{center}
        \end{minipage}
            \hfill
        \begin{minipage}[t]{0.24\linewidth}
            \begin{center}
                \centerline{\includegraphics[width=\textwidth]{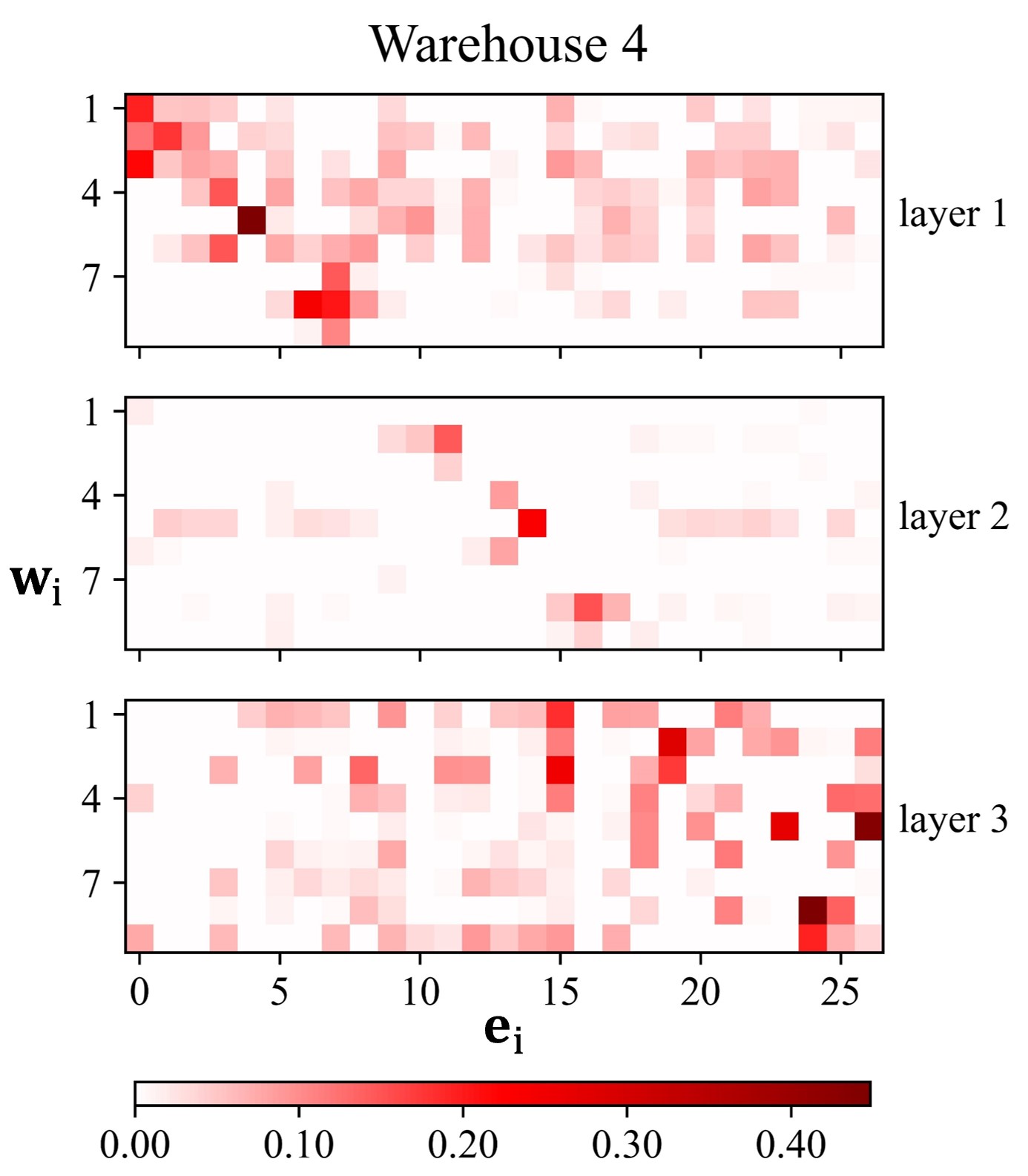}}
            \end{center}
        \end{minipage}
    \vskip -0.2 in
    \subcaption{}
    \end{minipage}
    \vskip -0.05 in
    \caption{Visualization of statistical mean values of learnt attention $\alpha_{ij}$ in each warehouse for KernelWarehouse with different attention functions. The results are obtained from the pre-trained ResNet18 backbone with KW ($1\times$) for all of the 50,000 images on the ImageNet validation set. Best viewed with zoom-in.
    The attention functions for the groups of visualization results are as follows:
    (a) $z_{ij}/\sum^{n}_{p=1}|z_{ip}|$ (our design); (b) softmax; (c) sigmoid; (d) $max(z_{ij},0)/\sum^{n}_{p=1}|z_{ip}|$.}
    \label{fig:visualization_attention_function}
\vskip 0.2 in
\end{figure}

\textbf{Visualization Results for KernelWarehouse with Different Attention Functions.}
The visualization results for KernelWarehouse with different attention functions are shown in Figure~\ref{fig:visualization_attention_function}, which are corresponding to the comparison results of Table~\ref{table:ablation_function} in the main manuscript. From which we can observe that: (1) for all of the attention functions, the maximum value of $\alpha_{ij}$ in each row mostly appears in the diagonal line throughout the whole warehouse. It indicates that our proposed attentions initialization strategy also works for the other three attention functions, which helps our KernelWarehouse to build one-to-one relationships between kernel cells and linear mixtures; (2) with different attention functions, the scalar attentions learnt by KernelWarehouse are obviously different, showing that the attention function plays an important role in our design; (3) compared to the other three functions, the maximum value of $\alpha_{ij}$ in each row tends to be relatively lower for our design (shown in Figure~\ref{fig:visualization_attention_function}(a)). It indicates that the introduction of negative values for scalar attentions can help the ConvNet to enhance warehouse sharing, where each linear mixture not only focuses on the kernel cell assigned to it.

\begin{figure}[ht]
\vskip 0.05 in
    \begin{minipage}[t]{1.0\linewidth}
        \begin{minipage}[t]{0.24\linewidth}
            \begin{center}
                \centerline{\includegraphics[width=\textwidth]{Figures/Attentions/resnet18_1x/Fig_Warehouse1.jpg}}
            \end{center}
        \end{minipage}
        \hfill
        \begin{minipage}[t]{0.24\linewidth}
            \begin{center}
                \centerline{\includegraphics[width=\textwidth]{Figures/Attentions/resnet18_1x/Fig_Warehouse2.jpg}}
            \end{center}
        \end{minipage}
        \hfill
        \begin{minipage}[t]{0.24\linewidth}
            \begin{center}
               \centerline{\includegraphics[width=\textwidth]{Figures/Attentions/resnet18_1x/Fig_Warehouse3.jpg}}
            \end{center}
        \end{minipage}
            \hfill
        \begin{minipage}[t]{0.24\linewidth}
            \begin{center}
                \centerline{\includegraphics[width=\textwidth]{Figures/Attentions/resnet18_1x/Fig_Warehouse4.jpg}}
            \end{center}
        \end{minipage}
    \vskip -0.2 in
    \subcaption{}
    \end{minipage}
    \begin{minipage}[t]{1.0\linewidth}
        \begin{minipage}[t]{0.24\linewidth}
            \begin{center}
                \includegraphics[width=\textwidth]{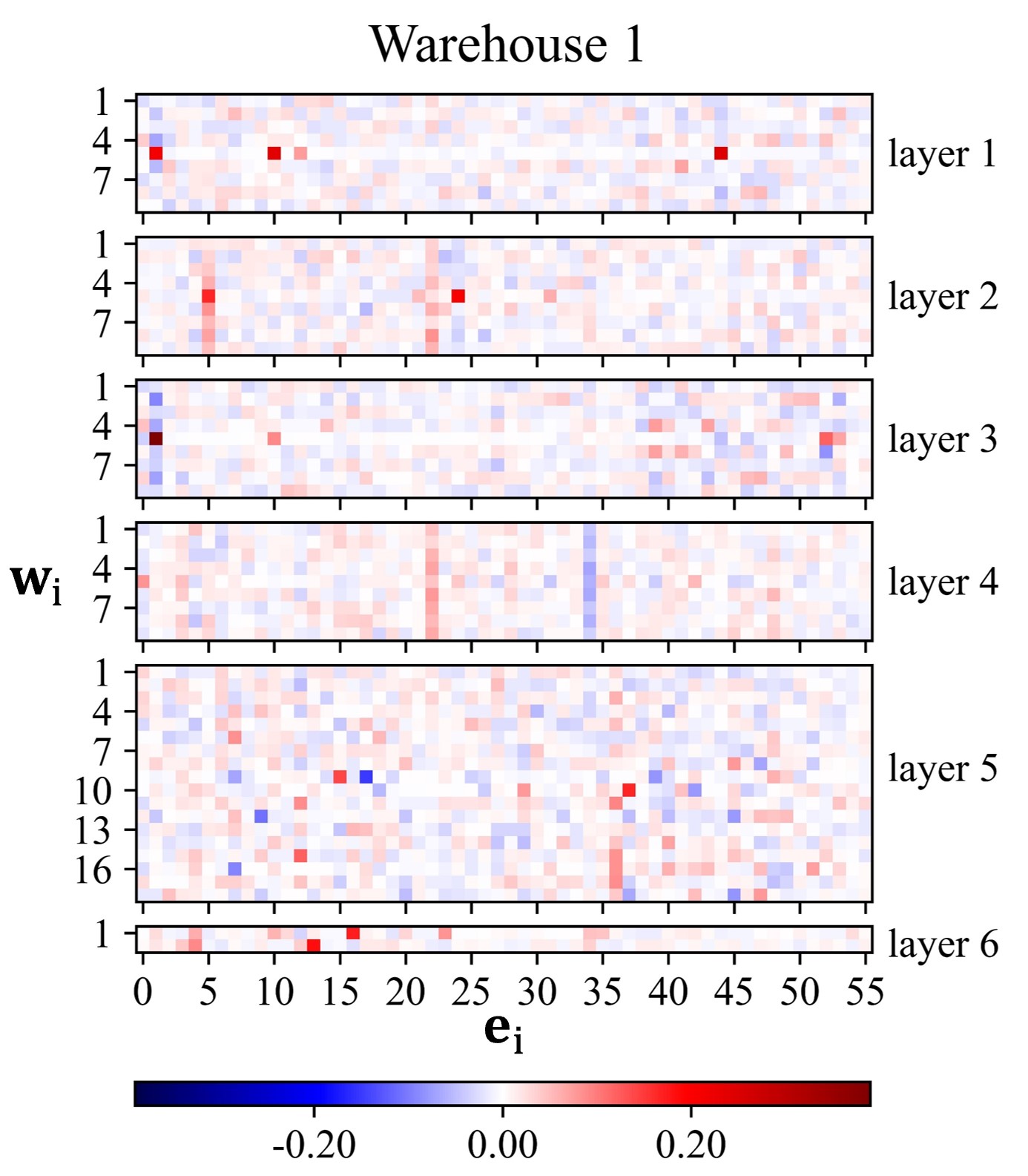}
            \end{center}
        \end{minipage}
        \hfill
        \begin{minipage}[t]{0.24\linewidth}
            \begin{center}
                \centerline{\includegraphics[width=\textwidth]{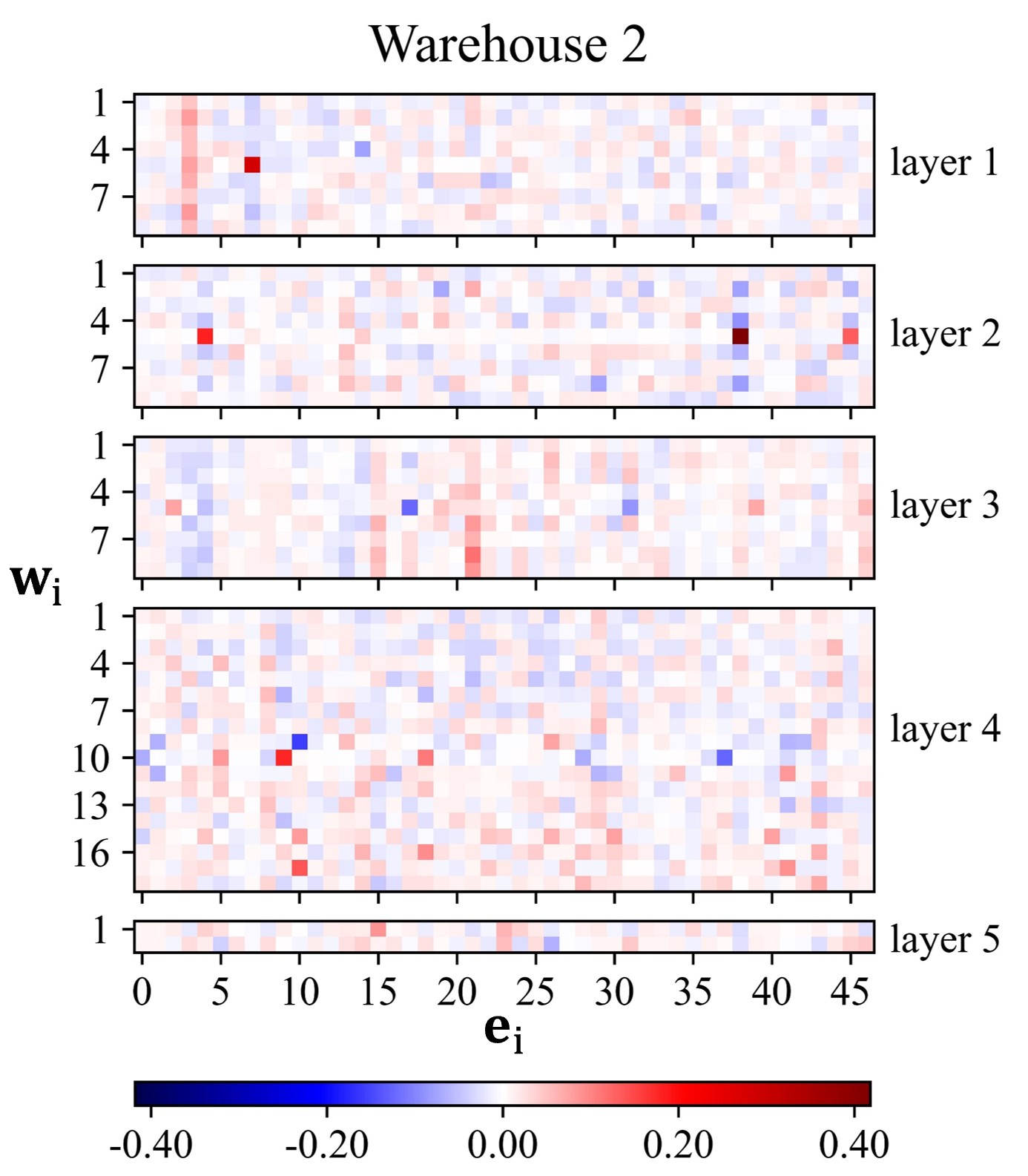}}
            \end{center}
        \end{minipage}
        \hfill
        \begin{minipage}[t]{0.24\linewidth}
            \begin{center}
                \centerline{\includegraphics[width=\textwidth]{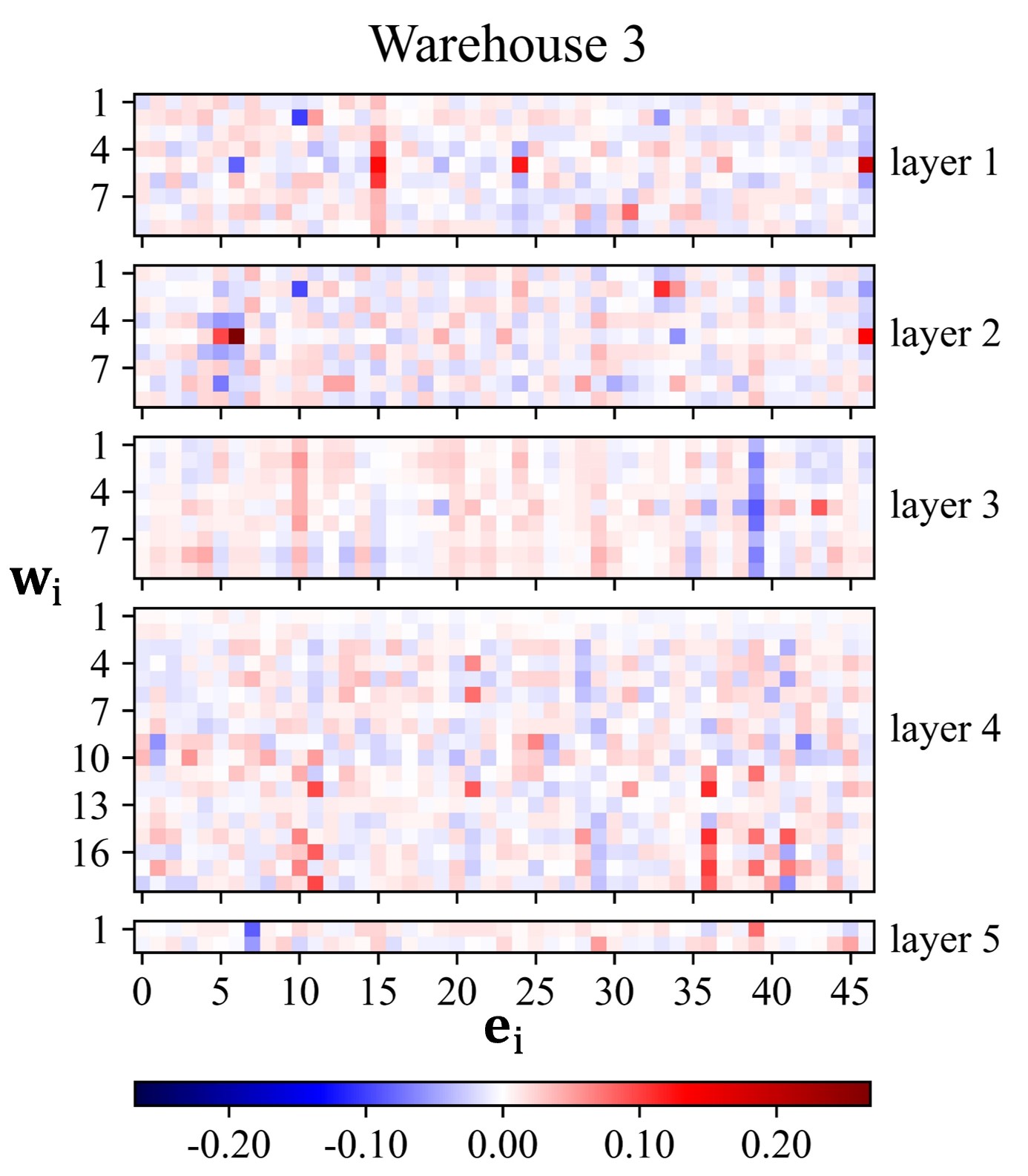}}
            \end{center}
        \end{minipage}
            \hfill
        \begin{minipage}[t]{0.24\linewidth}
            \begin{center}
                \centerline{\includegraphics[width=\textwidth]{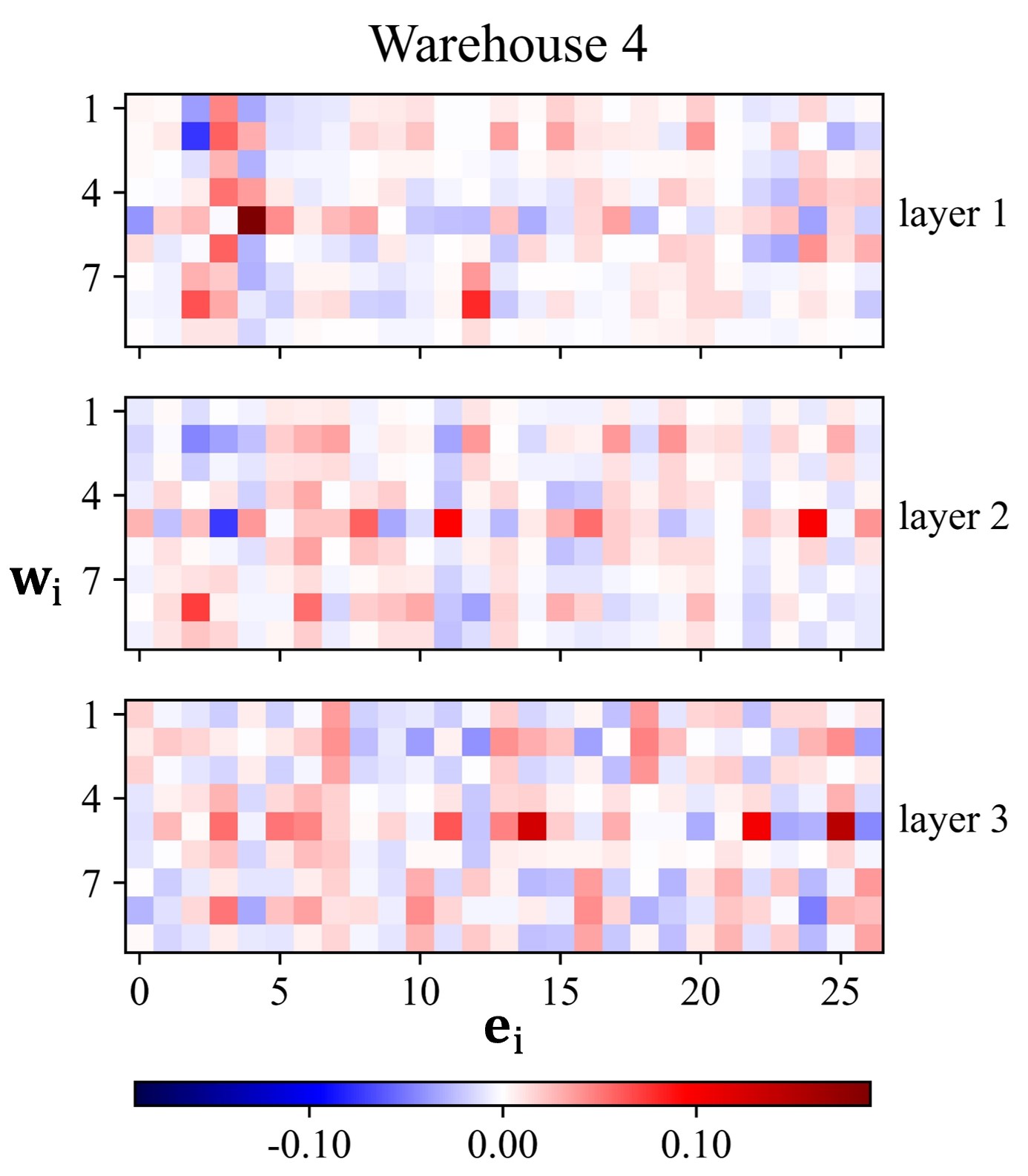}}
            \end{center}
        \end{minipage}
    \vskip -0.2 in
    \subcaption{}
    \end{minipage}
    \begin{minipage}[t]{1.0\linewidth}
        \begin{minipage}[t]{0.24\linewidth}
            \begin{center}
                \centerline{\includegraphics[width=\textwidth]{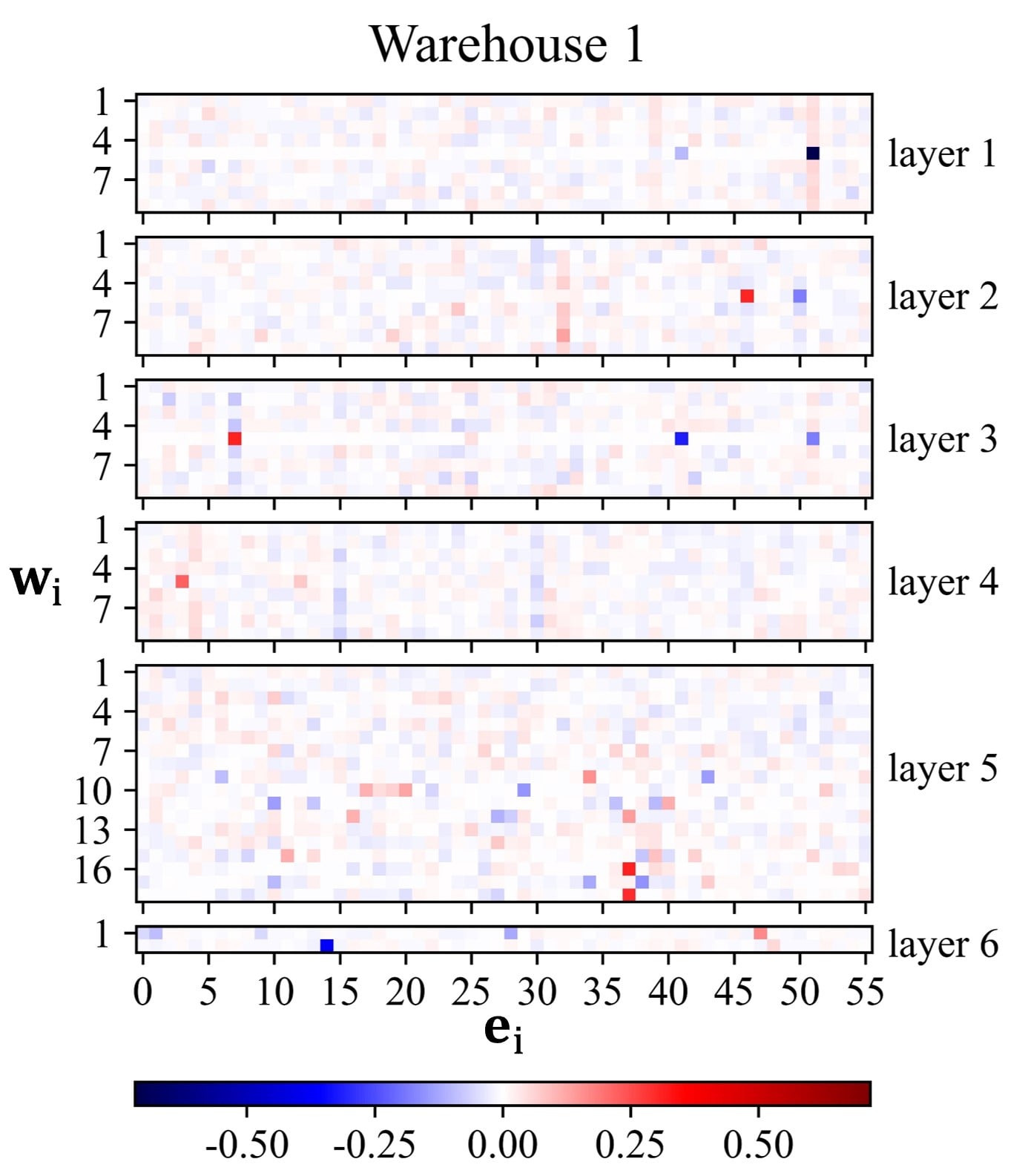}}
            \end{center}
        \end{minipage}
        \hfill
        \begin{minipage}[t]{0.24\linewidth}
            \begin{center}
                \centerline{\includegraphics[width=\textwidth]{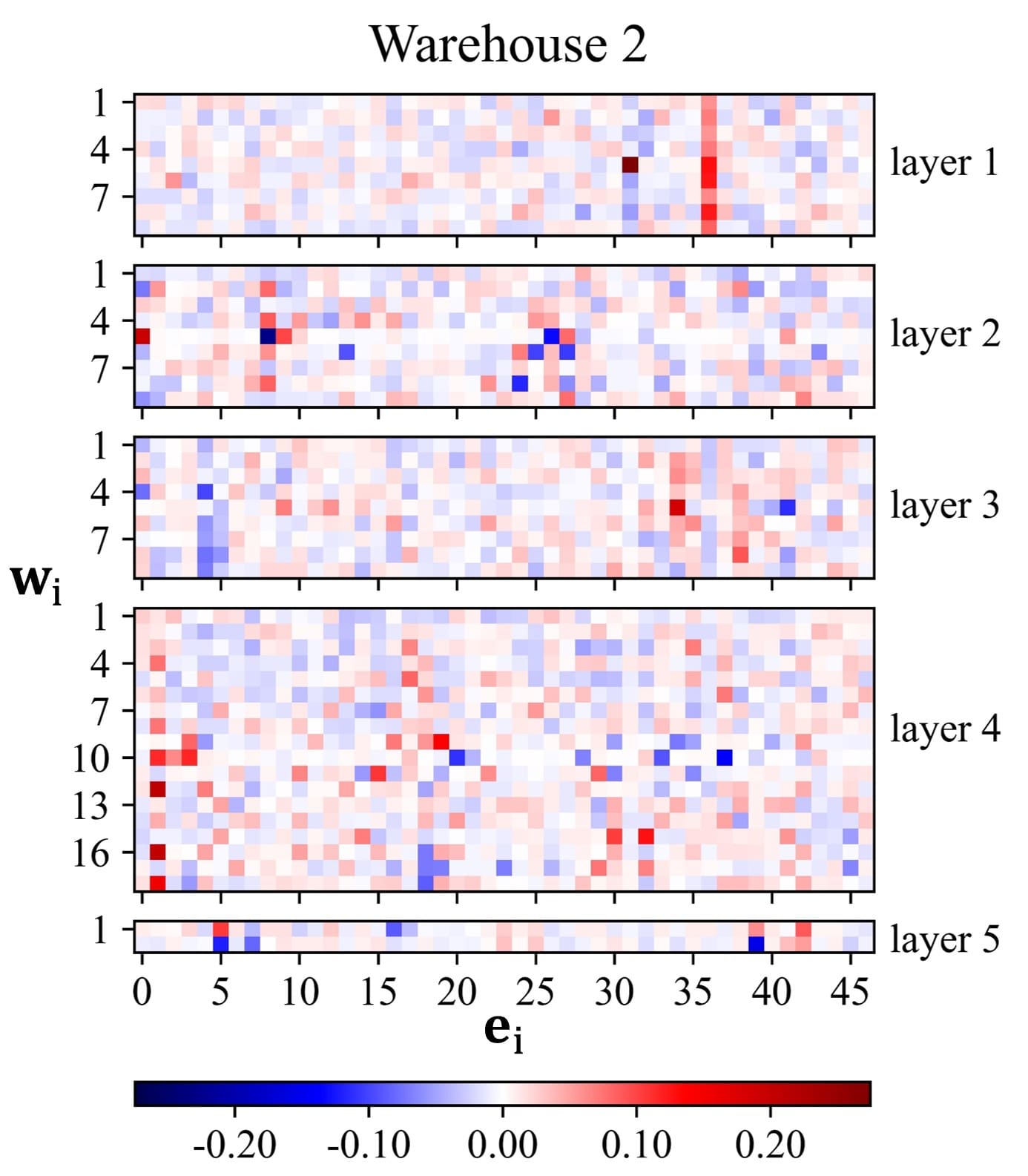}}
            \end{center}
        \end{minipage}
        \hfill
        \begin{minipage}[t]{0.24\linewidth}
            \begin{center}
                \centerline{\includegraphics[width=\textwidth]{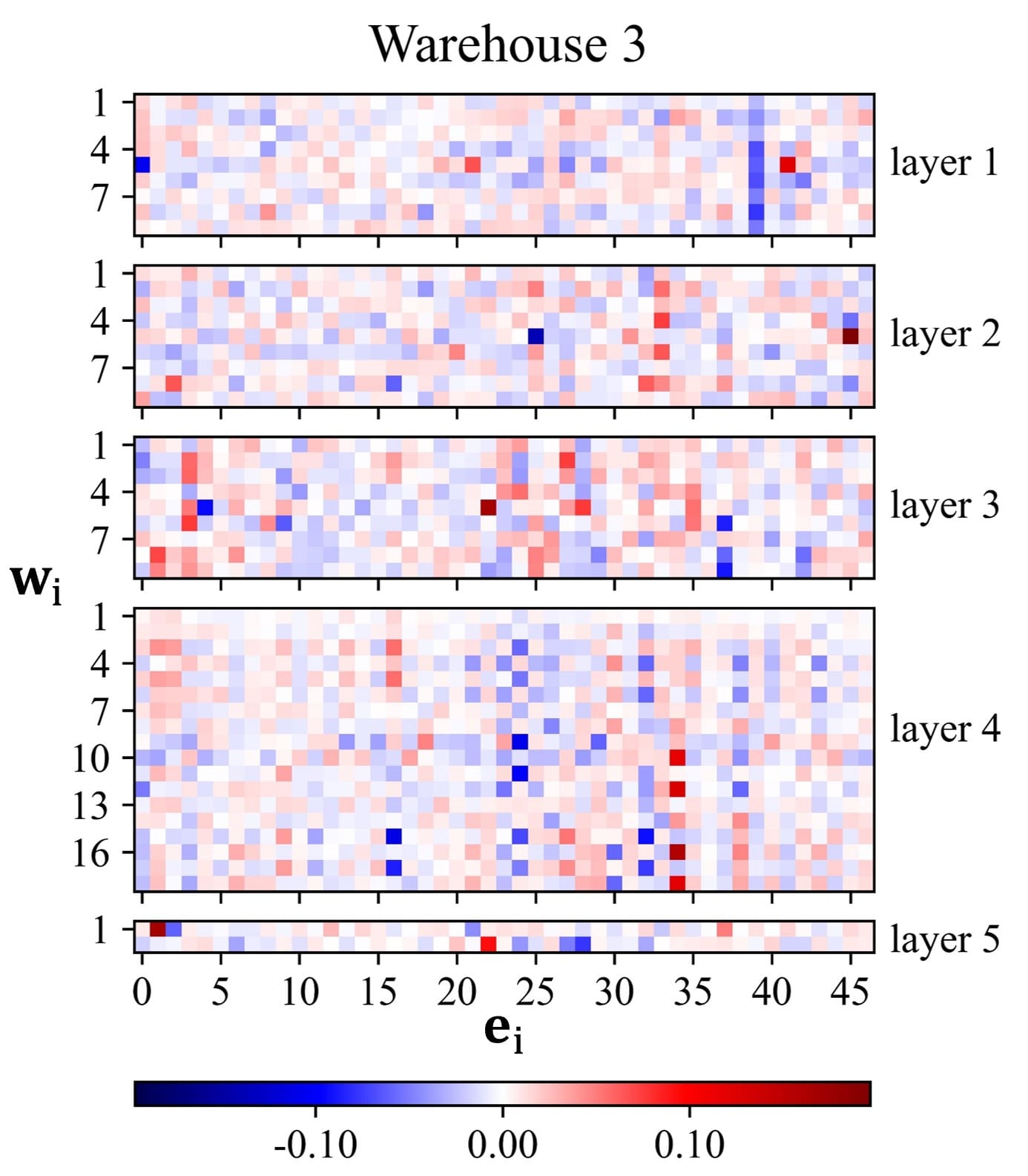}}
            \end{center}
        \end{minipage}
            \hfill
        \begin{minipage}[t]{0.24\linewidth}
            \begin{center}
                \centerline{\includegraphics[width=\textwidth]{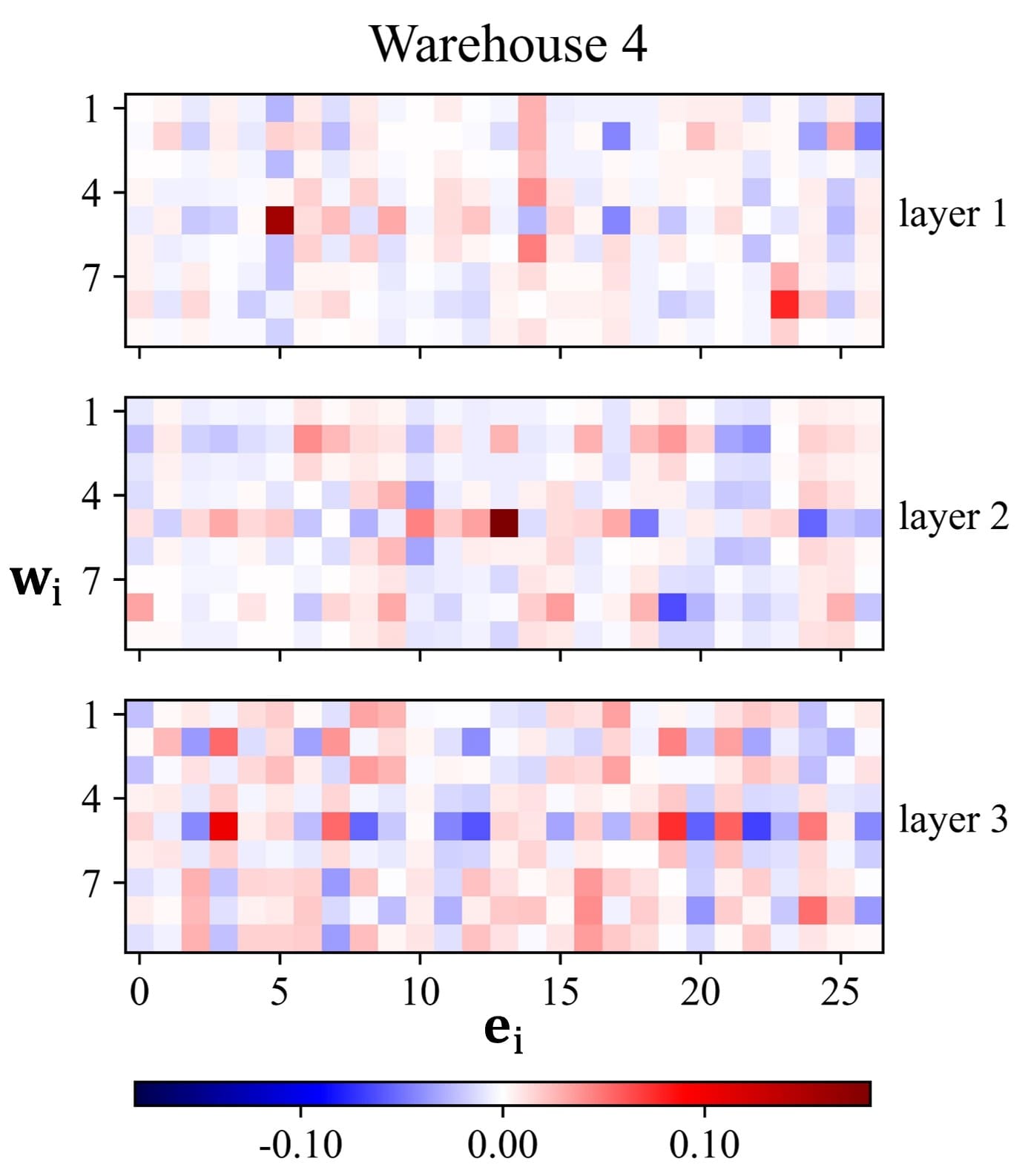}}
            \end{center}
        \end{minipage}
    \vskip -0.2 in
    \subcaption{}
    \end{minipage}
    \vskip -0.05 in
    \caption{Visualization of statistical mean values of learnt attention $\alpha_{ij}$ in each warehouse for KernelWarehouse with different attentions initialization strategies. The results are obtained from the pre-trained ResNet18 backbone with KW ($1\times$) for all of the 50,000 images on the ImageNet validation set. Best viewed with zoom-in.
    The attentions initialization strategies for the groups of visualization results are as follows:
    (a) building one-to-one relationships between kernel cells and linear mixtures; (b) building all-to-one relationships between kernel cells and linear mixtures; (c) without initialization.}
    \label{fig:visualization_initialization_strategy_1x}
\vskip 0.2 in
\end{figure}

\begin{figure}[ht]
\vskip 0.05 in
    \begin{minipage}[t]{1.0\linewidth}
        \begin{minipage}[t]{0.49\linewidth}
            \begin{center}
                \centerline{\includegraphics[width=\textwidth]{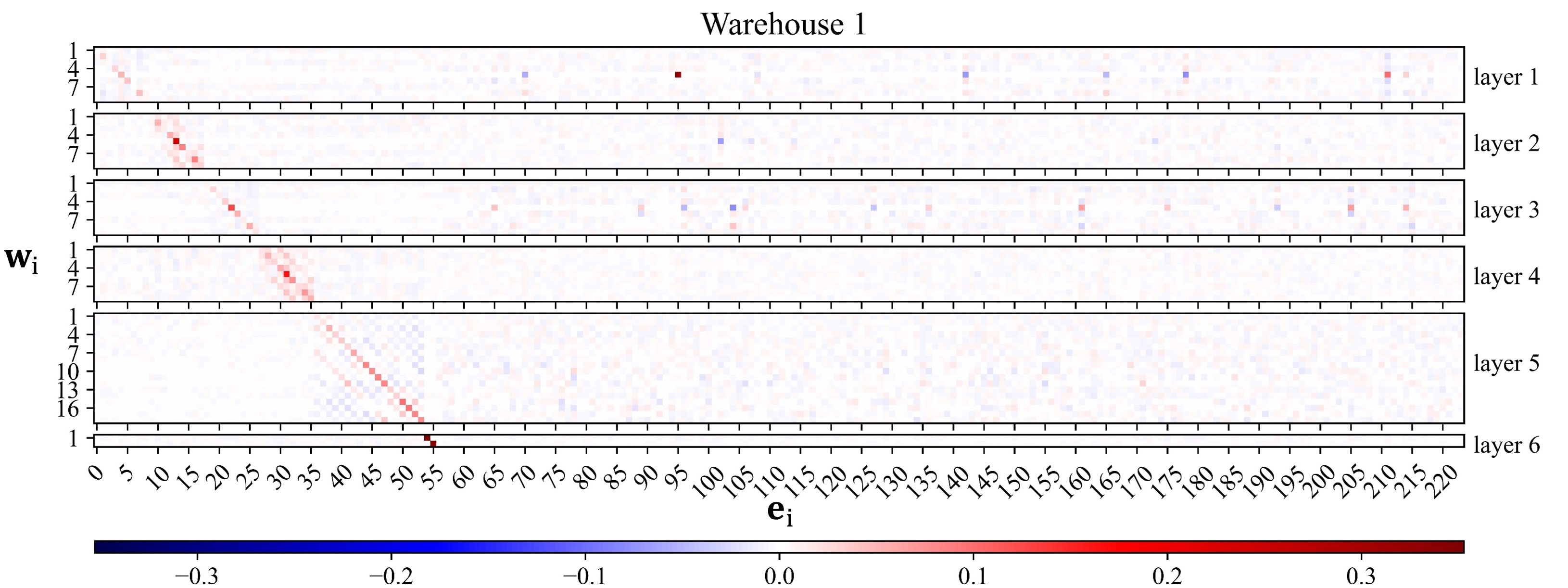}}
            \end{center}
        \end{minipage}
        \hfill
        \begin{minipage}[t]{0.49\linewidth}
            \begin{center}
                \centerline{\includegraphics[width=\textwidth]{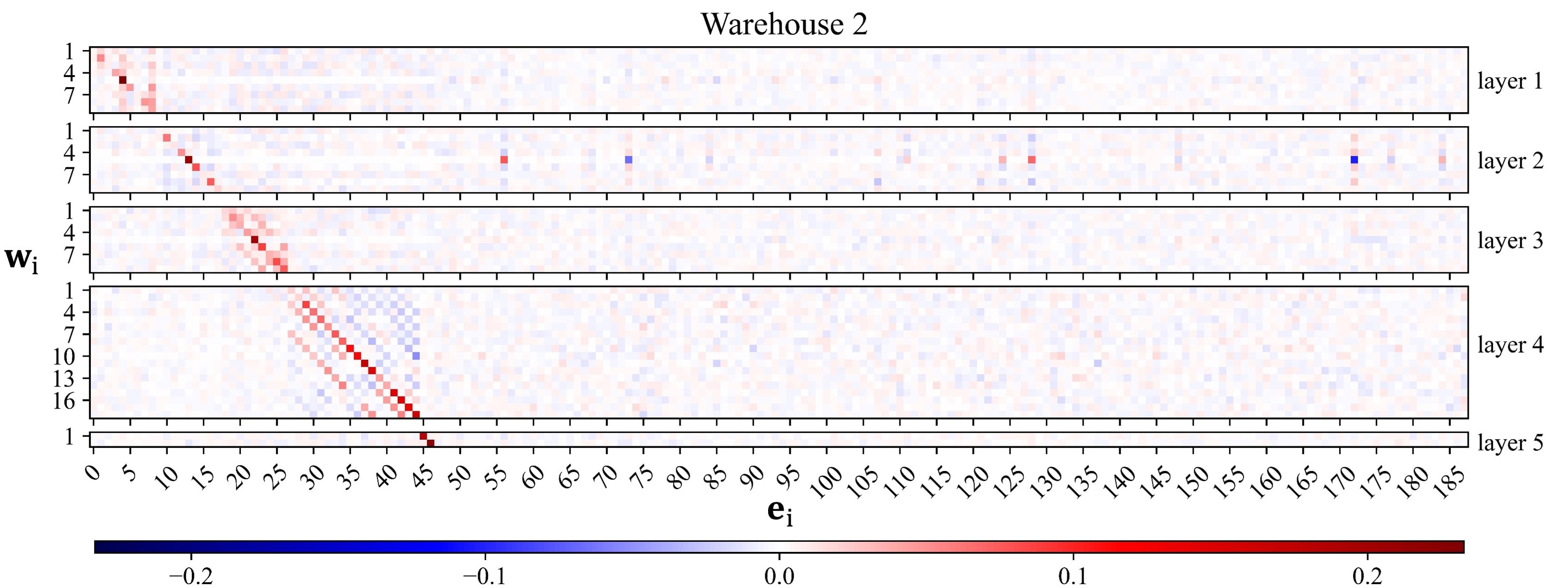}}
            \end{center}
        \end{minipage}
        \hfill
        \begin{minipage}[t]{0.49\linewidth}
            \begin{center}
                \centerline{\includegraphics[width=\textwidth]{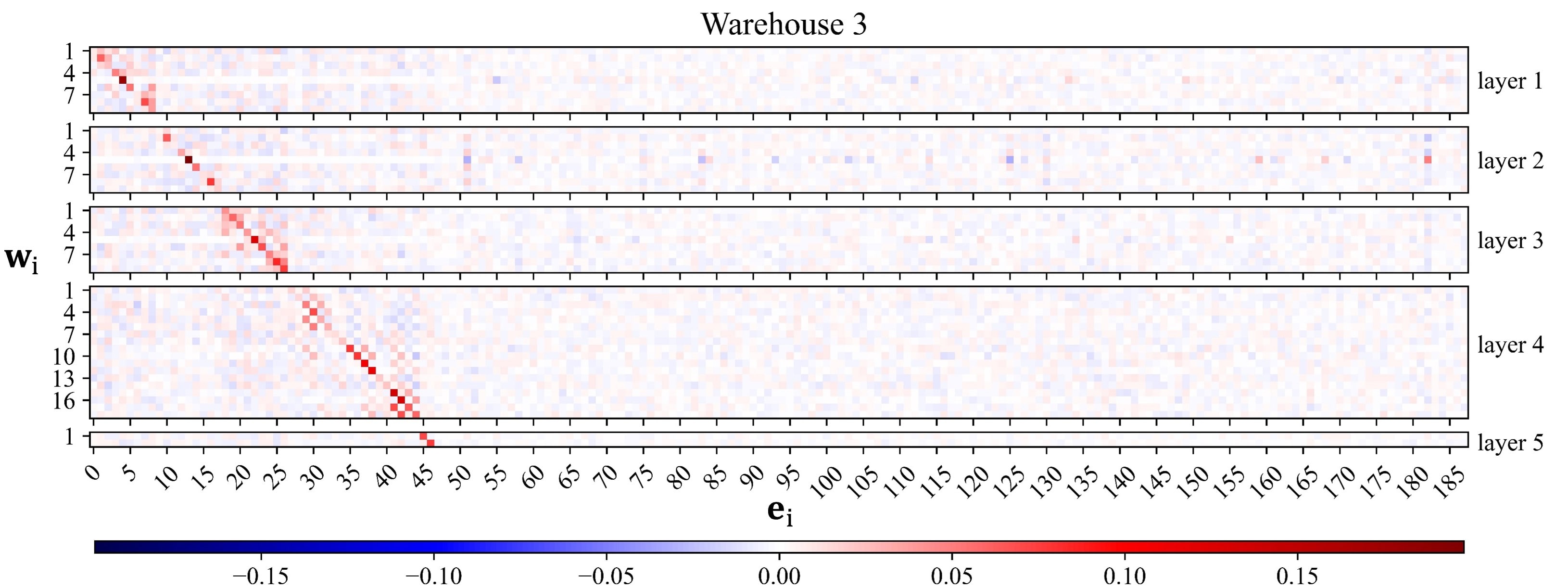}}
            \end{center}
        \end{minipage}
            \hfill
        \begin{minipage}[t]{0.49\linewidth}
            \begin{center}
                \centerline{\includegraphics[width=\textwidth]{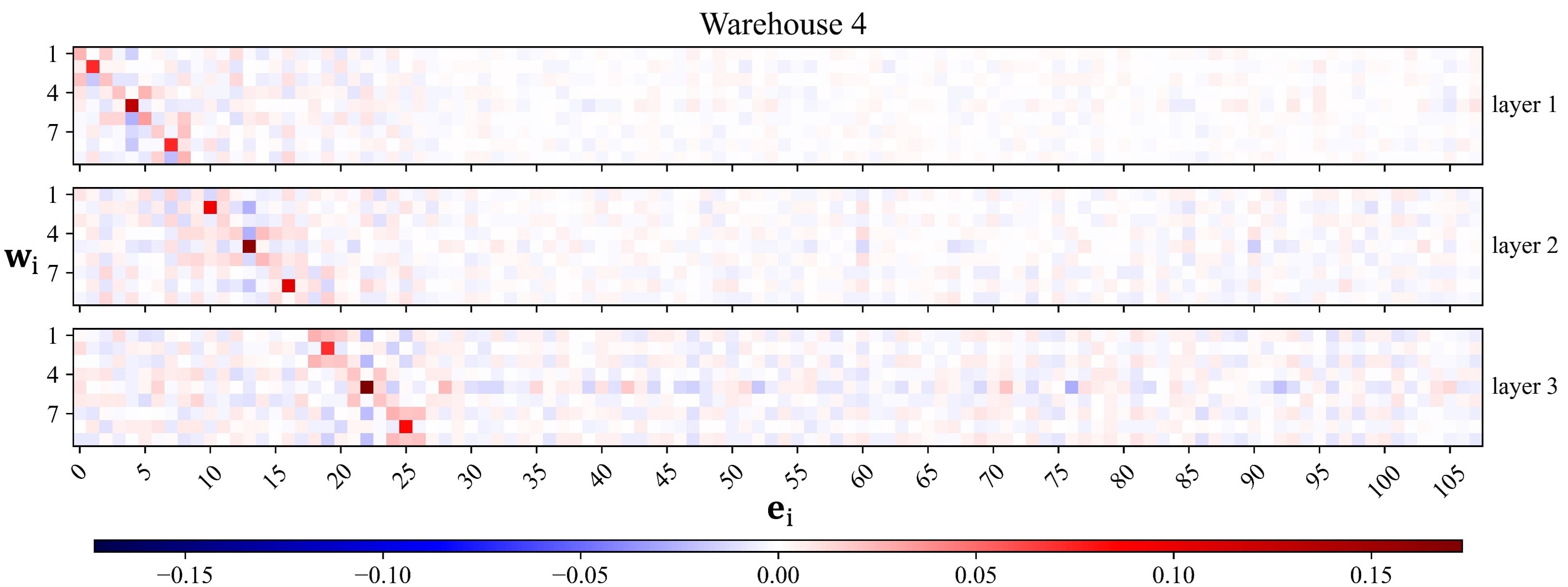}}
            \end{center}
        \end{minipage}
    \vskip -0.2 in
    \subcaption{}
    \end{minipage}

    \begin{minipage}[t]{1.0\linewidth}
        \begin{minipage}[t]{0.49\linewidth}
            \begin{center}
                \centerline{\includegraphics[width=\textwidth]{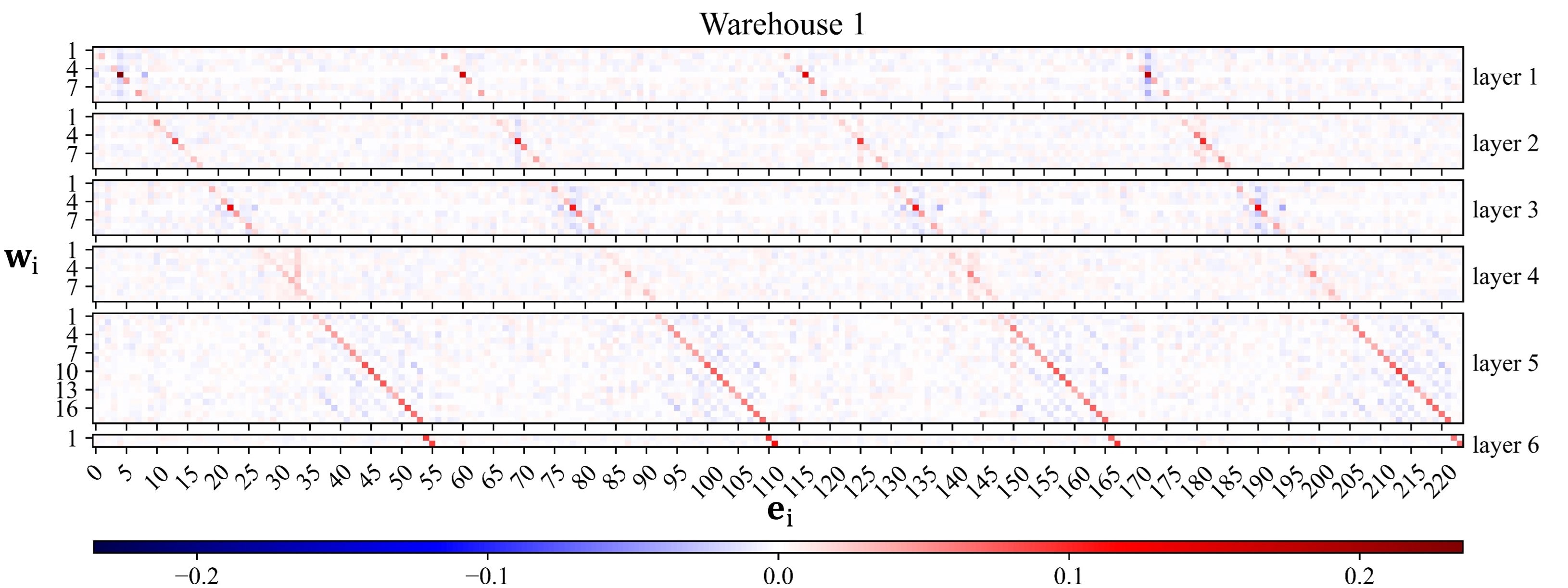}}
            \end{center}
        \end{minipage}
        \hfill
        \begin{minipage}[t]{0.49\linewidth}
            \begin{center}
                \centerline{\includegraphics[width=\textwidth]{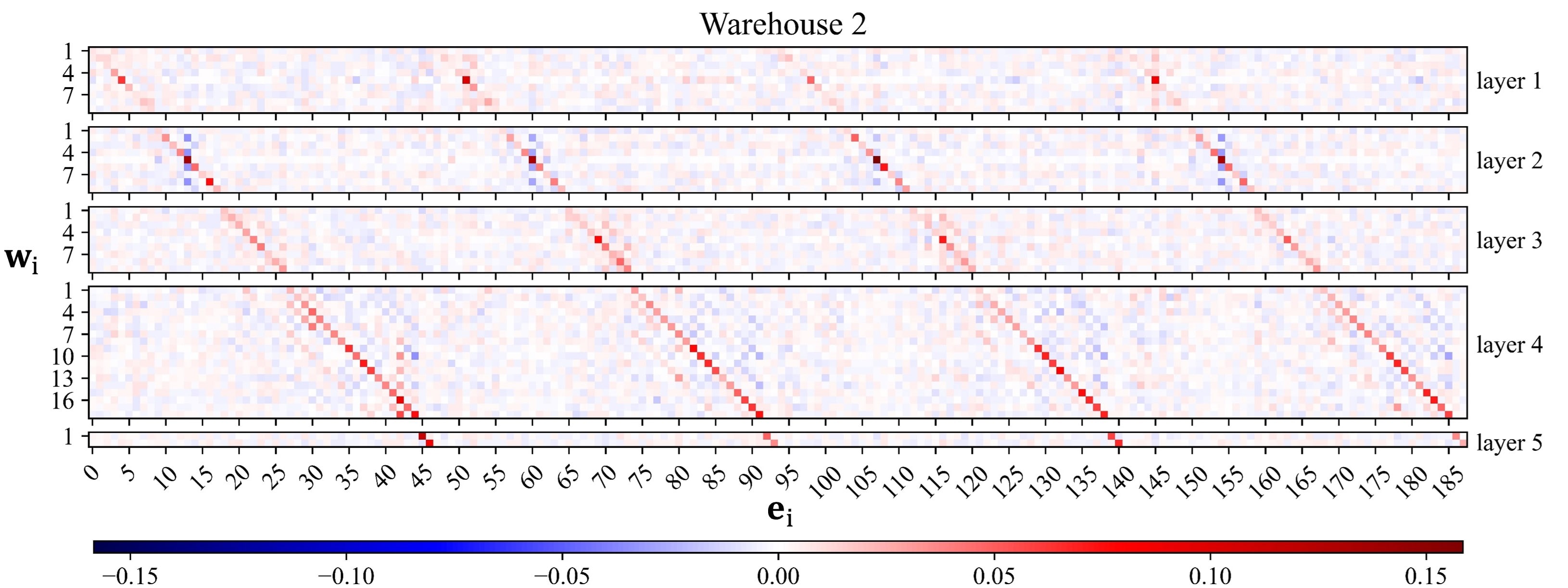}}
            \end{center}
        \end{minipage}
        \hfill
        \begin{minipage}[t]{0.49\linewidth}
            \begin{center}
                \centerline{\includegraphics[width=\textwidth]{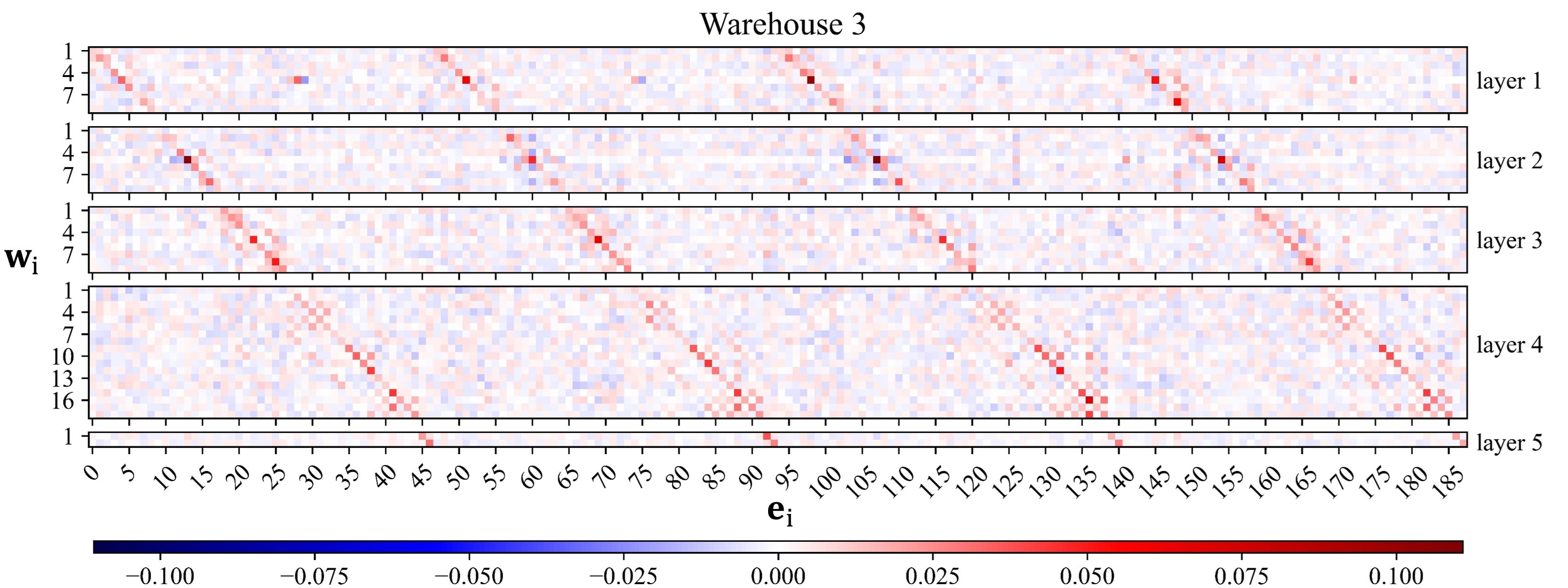}}
            \end{center}
        \end{minipage}
            \hfill
        \begin{minipage}[t]{0.49\linewidth}
            \begin{center}
                \centerline{\includegraphics[width=\textwidth]{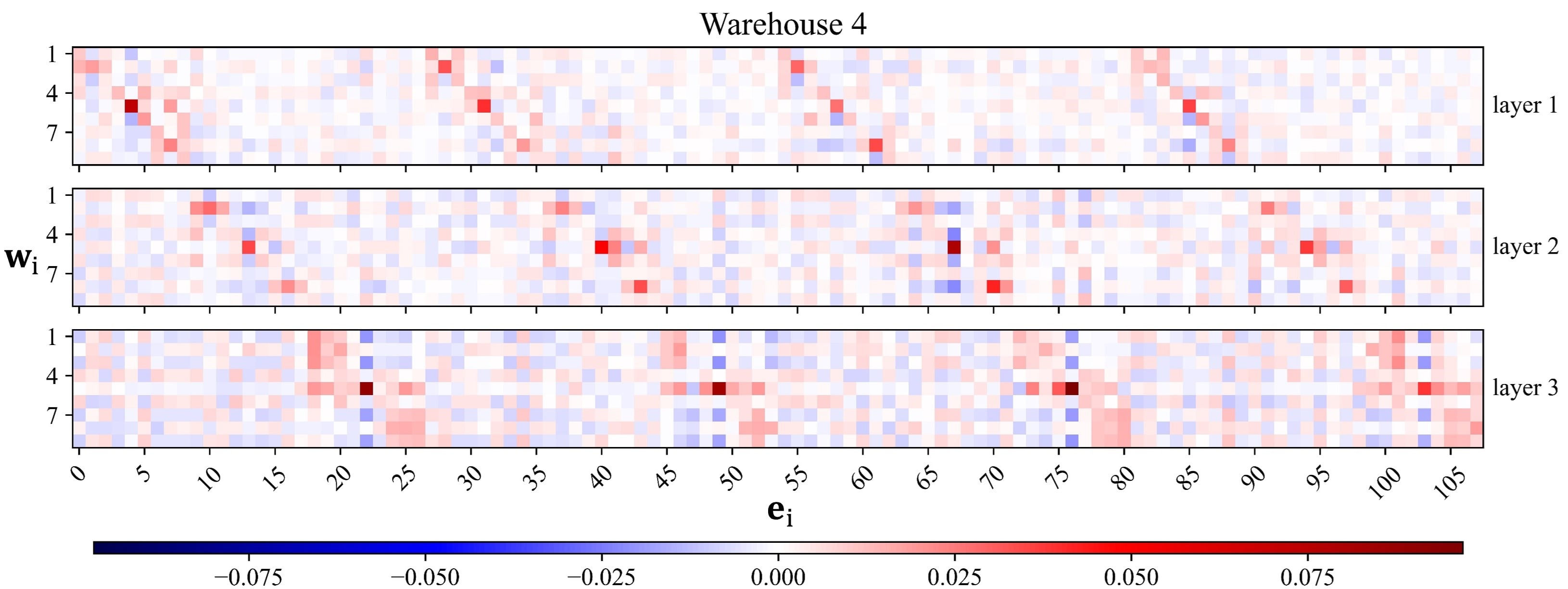}}
            \end{center}
        \end{minipage}
    \vskip -0.2 in
    \subcaption{}
    \end{minipage}
    \vskip -0.05 in
    \caption{Visualization of statistical mean values of learnt attention $\alpha_{ij}$ in each warehouse for KernelWarehouse with different attentions initialization strategies. The results are obtained from the pre-trained ResNet18 backbone with KW ($4\times$) for all of the 50,000 images on the ImageNet validation set. Best viewed with zoom-in.
    The attentions initialization strategies for the groups of visualization results are as follows:
    (a) building one-to-one relationships between kernel cells and linear mixtures; (b) building four-to-one relationships between kernel cells and linear mixtures.}
    \label{fig:visualization_initialization_strategy_4x}
\vskip -0.2 in
\end{figure}

\begin{figure}[ht]
\vskip 0.05 in
    \begin{minipage}[t]{1.0\linewidth}
        \begin{minipage}[t]{0.24\linewidth}
            \begin{center}
                \centerline{\includegraphics[width=\textwidth]{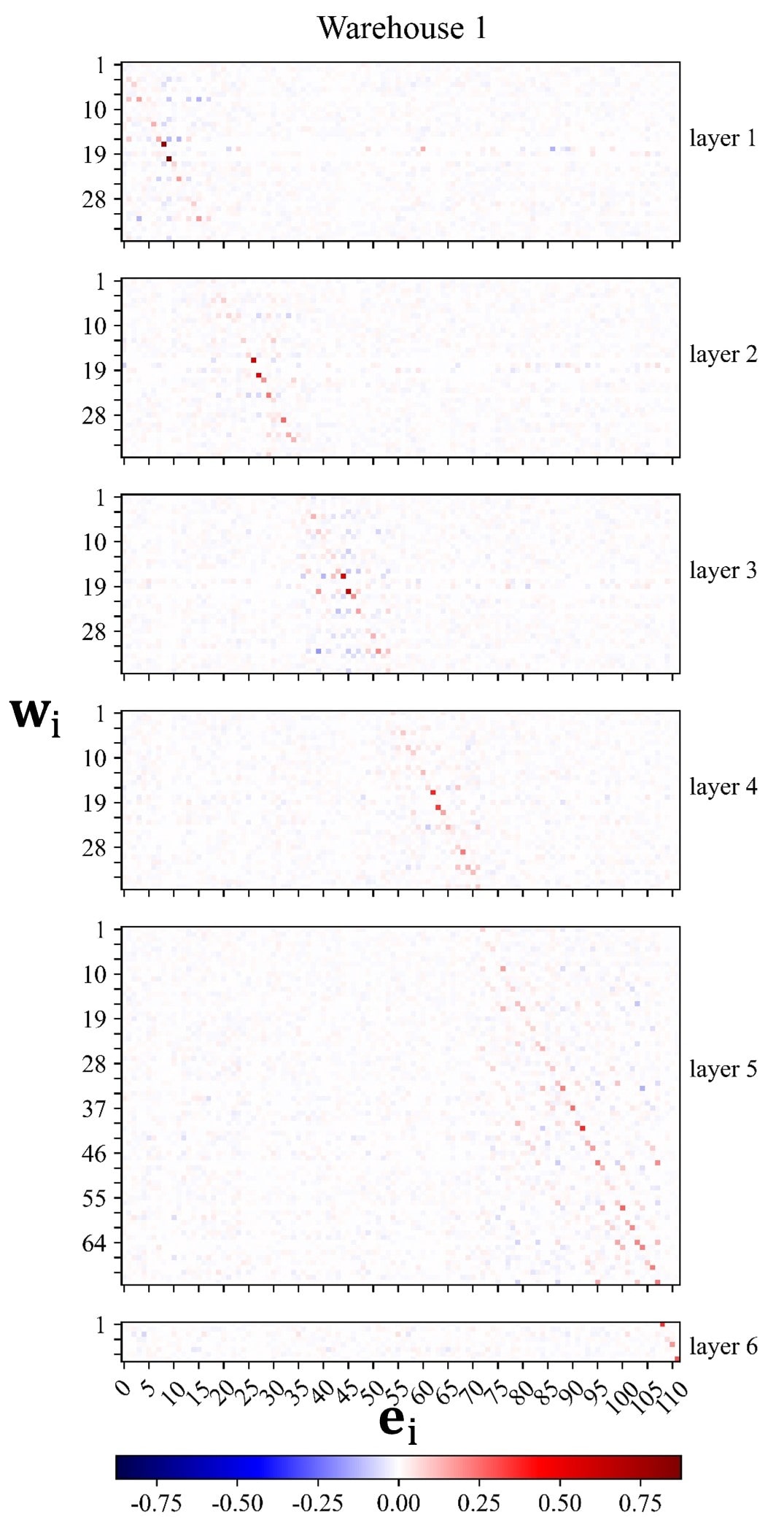}}
            \end{center}
        \end{minipage}
        \hfill
        \begin{minipage}[t]{0.24\linewidth}
            \begin{center}
                \centerline{\includegraphics[width=\textwidth]{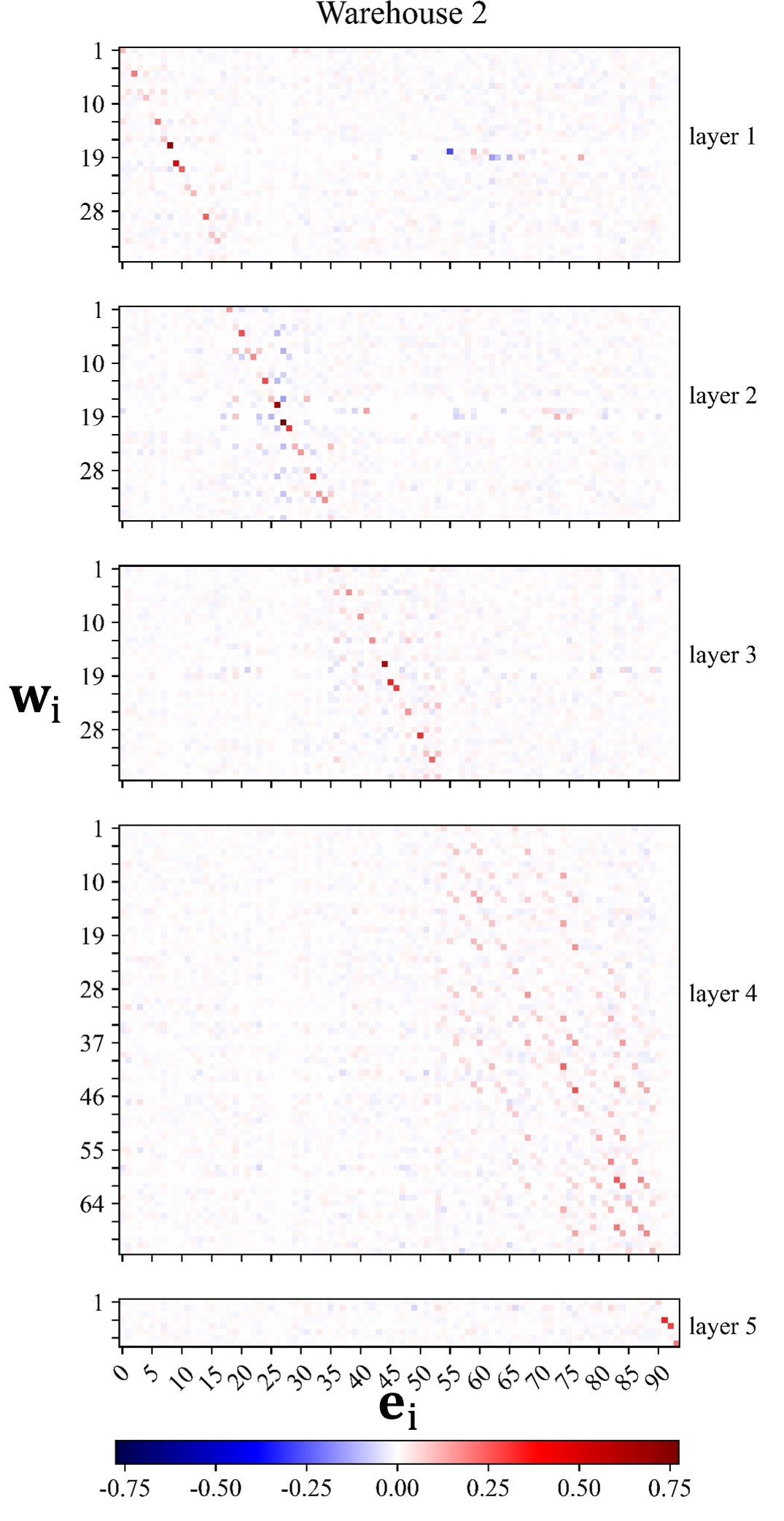}}
            \end{center}
        \end{minipage}
        \hfill
        \begin{minipage}[t]{0.24\linewidth}
            \begin{center}
                \centerline{\includegraphics[width=\textwidth]{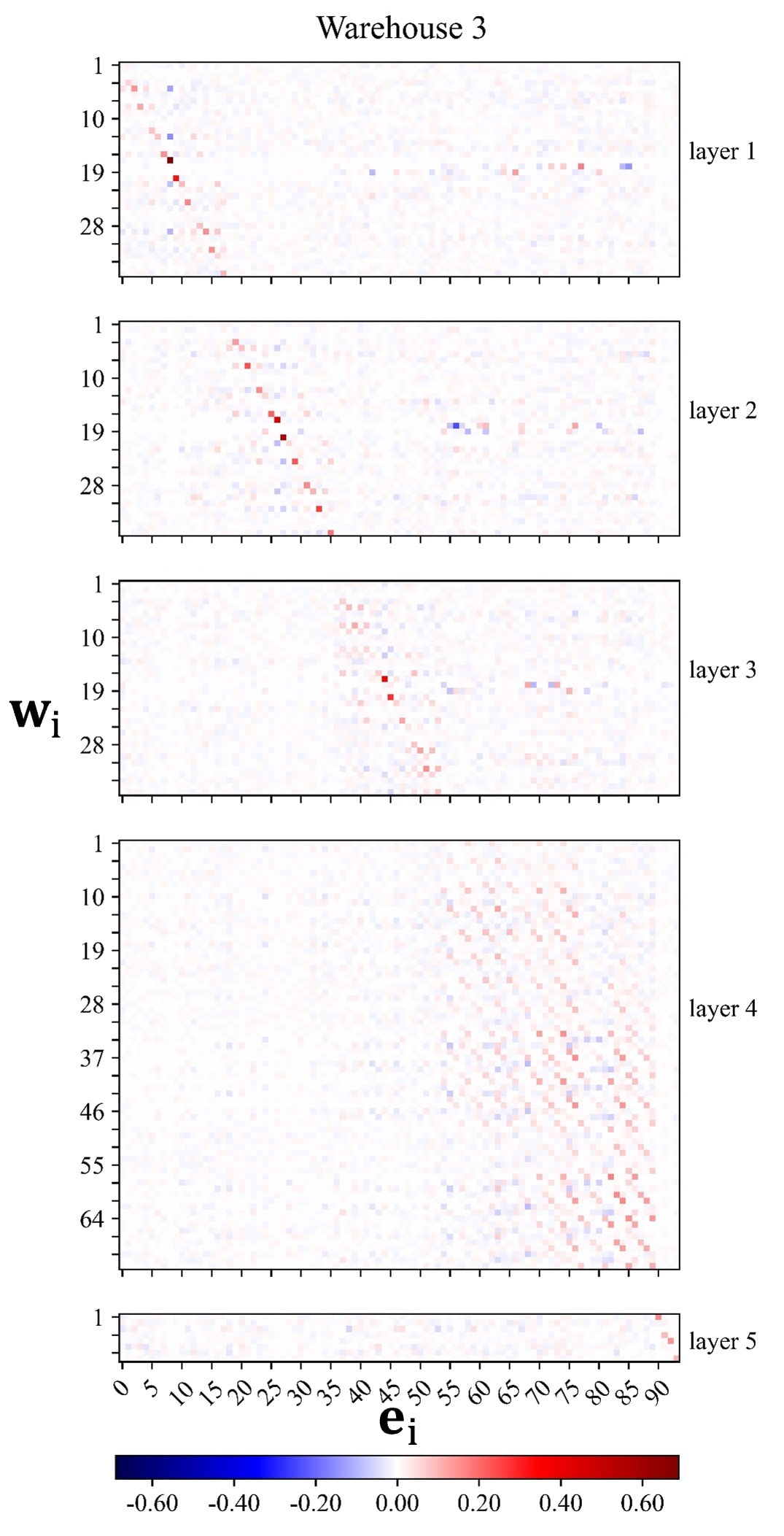}}
            \end{center}
        \end{minipage}
            \hfill
        \begin{minipage}[t]{0.24\linewidth}
            \begin{center}
                \centerline{\includegraphics[width=\textwidth]{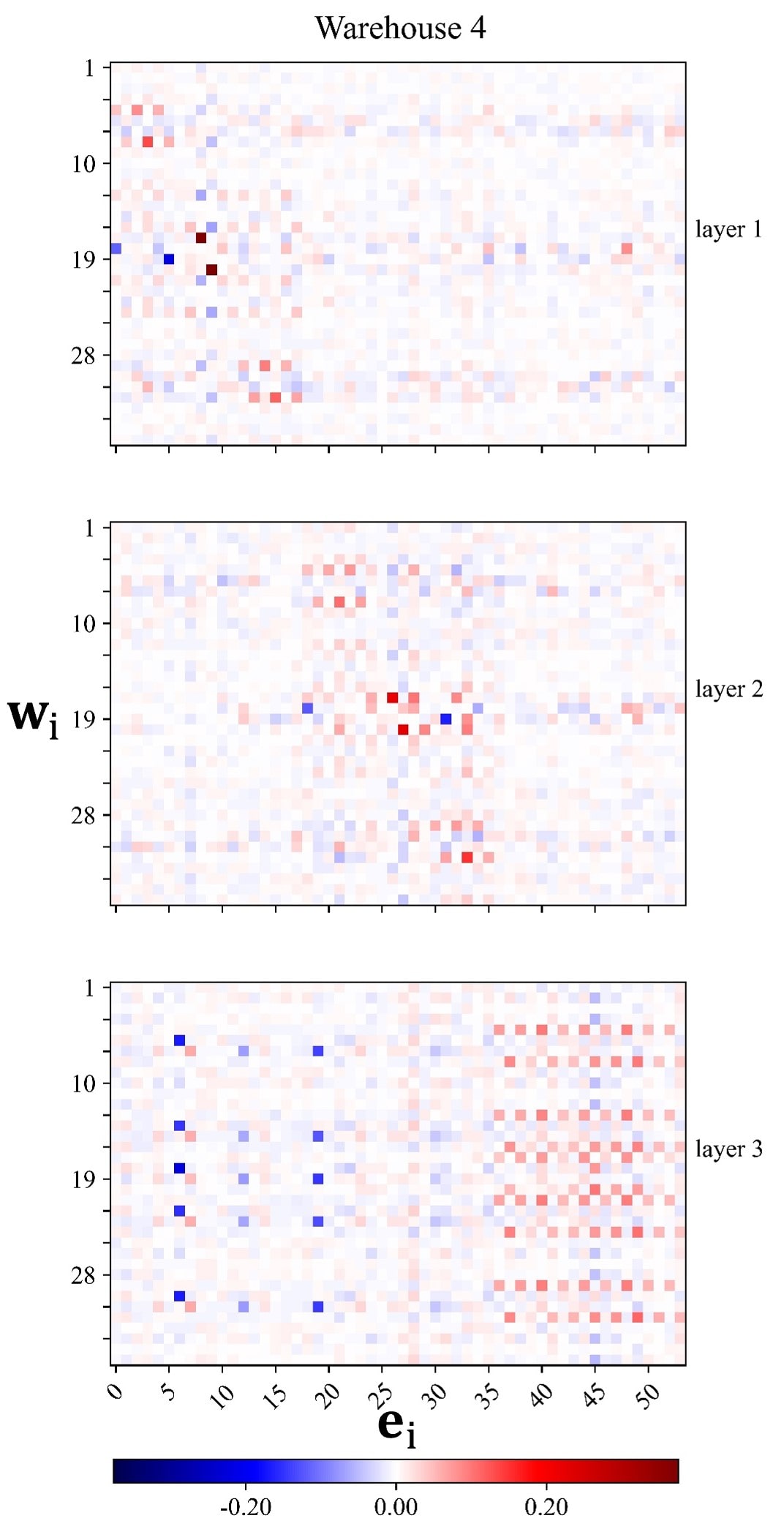}}
            \end{center}
        \end{minipage}
    \vskip -0.2 in
    \subcaption{}
    \end{minipage}

    \begin{minipage}[t]{1.0\linewidth}
        \begin{minipage}[t]{0.24\linewidth}
            \begin{center}
                \centerline{\includegraphics[width=\textwidth]{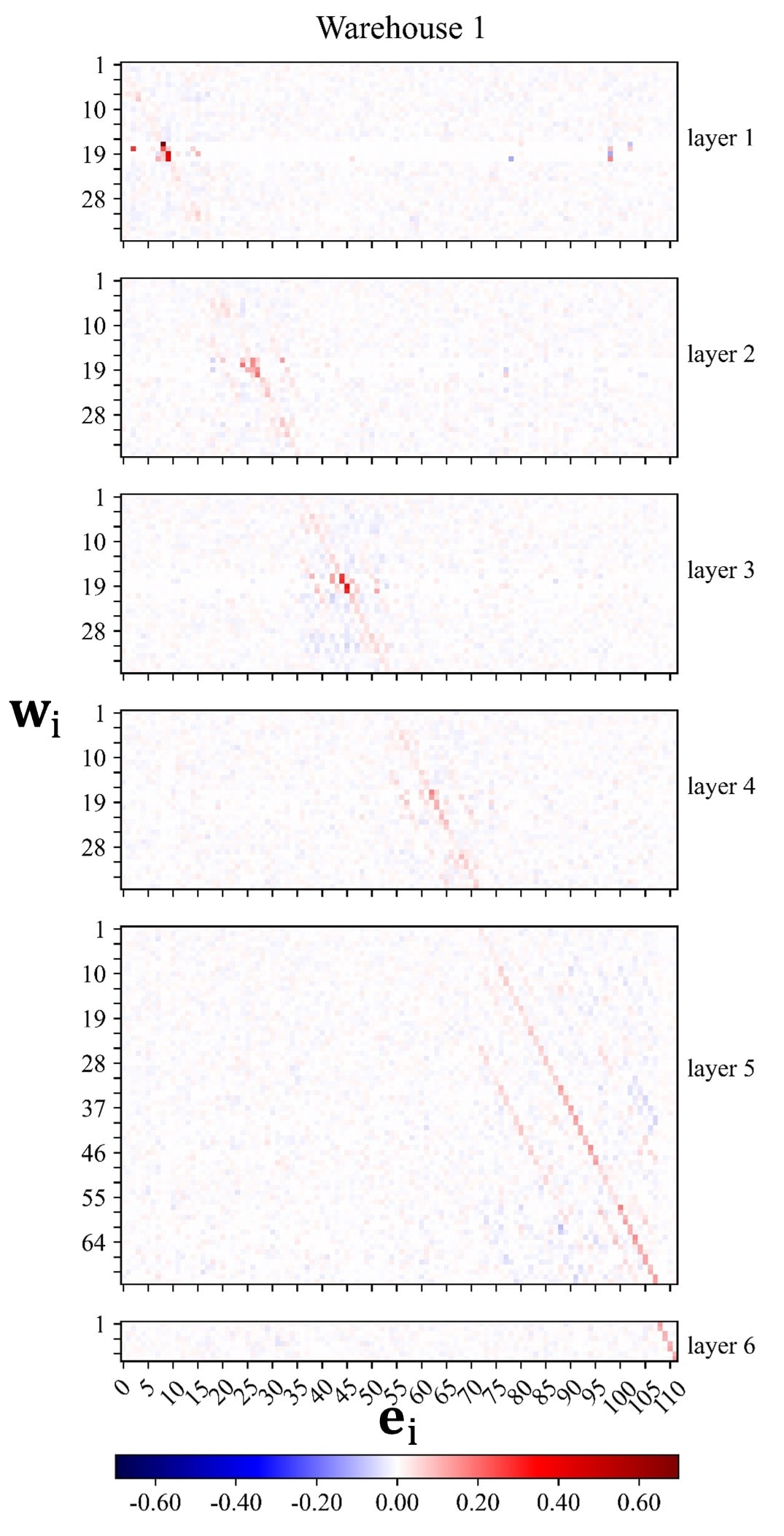}}
            \end{center}
        \end{minipage}
        \hfill
        \begin{minipage}[t]{0.24\linewidth}
            \begin{center}
                \centerline{\includegraphics[width=\textwidth]{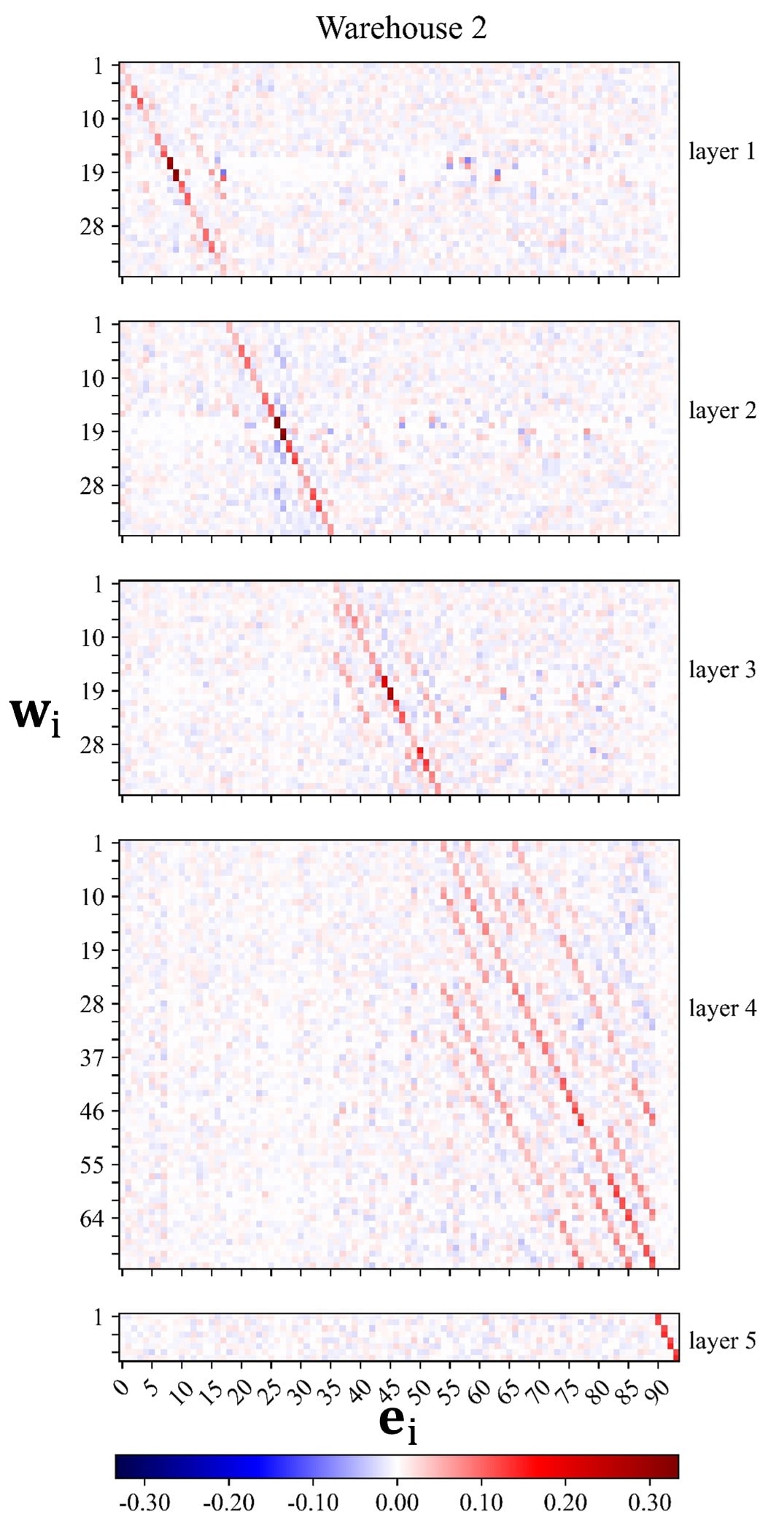}}
            \end{center}
        \end{minipage}
        \hfill
        \begin{minipage}[t]{0.24\linewidth}
            \begin{center}
                \centerline{\includegraphics[width=\textwidth]{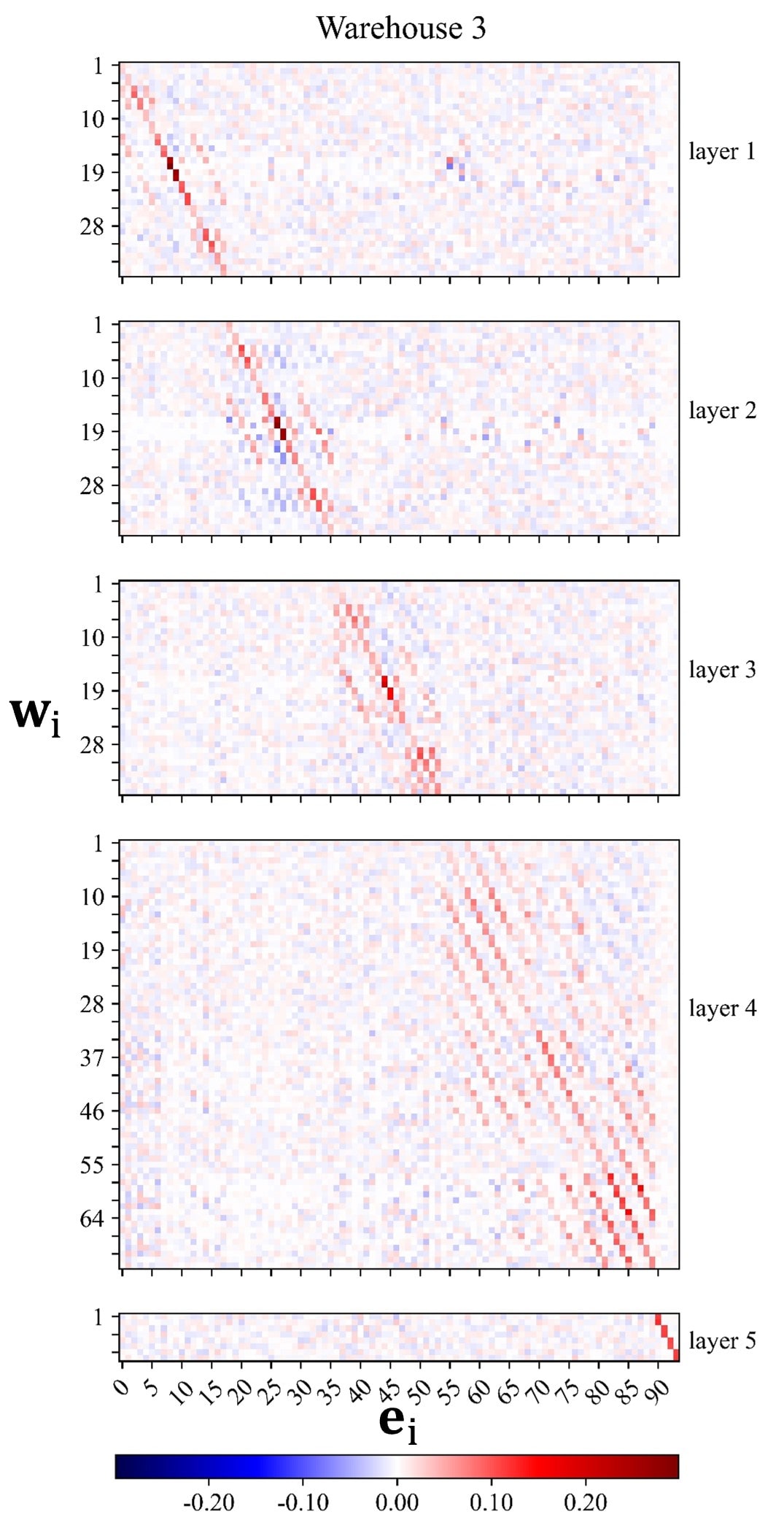}}
            \end{center}
        \end{minipage}
            \hfill
        \begin{minipage}[t]{0.24\linewidth}
            \begin{center}
                \centerline{\includegraphics[width=\textwidth]{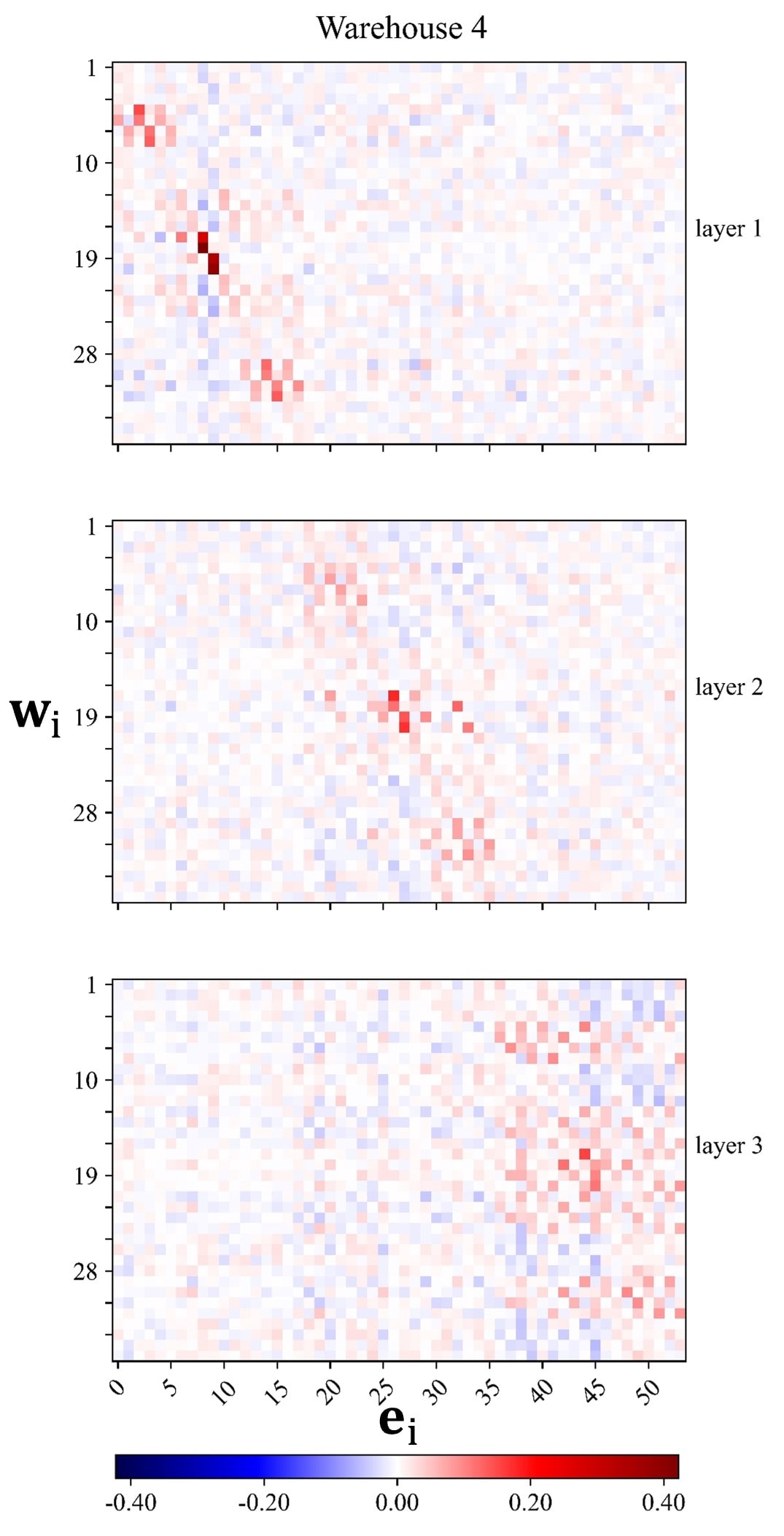}}
            \end{center}
        \end{minipage}
    \vskip -0.2 in
    \subcaption{}
    \end{minipage}
    \vskip -0.05 in
    \caption{Visualization of statistical mean values of learnt attention $\alpha_{ij}$ in each warehouse for KernelWarehouse with different attentions initialization strategies. The results are obtained from the pre-trained ResNet18 backbone with KW ($1/2\times$) for all of the 50,000 images on the ImageNet validation set. Best viewed with zoom-in.
    The attentions initialization strategies for the groups of visualization results are as follows:
    (a) building one-to-one relationships between kernel cells and linear mixtures; (b) building one-to-two relationships between kernel cells and linear mixtures.}
    \label{fig:visualization_initialization_strategy_1d2x}
\vskip 0.2 in
\end{figure}

\textbf{Visualization Results for KernelWarehouse with Attentions Initialization Strategies.}
The visualization results for KernelWarehouse with different attentions initialization strategies are shown in Figure~\ref{fig:visualization_initialization_strategy_1x}, Figure~\ref{fig:visualization_initialization_strategy_4x} and Figure~\ref{fig:visualization_initialization_strategy_1d2x}, which are corresponding to the comparison results of Table~\ref{table:ablation_initialization}. From which we can observe that: (1) with all-to-one strategy or without initialization strategy, the distribution of scalar attentions learnt by KernelWarehouse seems to be disordered, while our proposed strategy can help the ConvNet learn more appropriate relationships between kernel cells and linear mixtures;
(2) for KW ($4\times$) and KW ($1/2\times$), it's hard to directly determine which strategy is better only according to the visualization results. While the results demonstrate that the learnt attentions of KernelWarehouse are highly related to our setting of $\alpha_{ij}$;
(3) for KW ($1\times$), KW ($4\times$) and KW ($1/2\times$) with our proposed initialization strategy, some similar patterns of the value distributions can be found.
For example, the maximum value of $\alpha_{ij}$ in each row mostly appears in the diagonal line throughout the whole warehouse. It indicates that our proposed strategy can help the ConvNet learn stable relationships between kernel cells and linear mixtures.


\end{document}